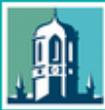

# ARAN - Access to Research at NUI Galway

| Title | Improved cardiac arrhythmia prediction based on heart rate variability analysis |
| --- | --- |
| Author(s) | Parsi, Ashkan |
| Publication Date | 2021-05-10 |
| Publisher | NUI Galway |
| Item record | http://hdl.handle.net/10379/17007 |

# Improved Cardiac Arrhythmia Prediction Based on Heart Rate Variability Analysis

Presented by:
**Ashkan Parsi**

To:
Electrical and Electronic Engineering,
School of Engineering,
College of Science and Engineering,
National University of Ireland Galway

In fulfilment of the requirements for the degree of Doctor of Philosophy.

Supervised by:
Prof. Edward Jones

Co-supervised by:
Prof. Martin Glavin

May 2021

# Contents

# Declaration of Originality

I the Candidate **Ashkan Parsi**, certify that this thesis entitled "**Improved Cardiac Arrhythmia Prediction Based on Heart Rate Variability Analysis**":

- is all my own work;
- has not been previously submitted for any degree or qualification at this University or any other institution;
- and where any work in this thesis was conducted in collaboration, appropriate reference to published work by my collaborators has been made and the nature and extent of my contribution has been clearly stated.

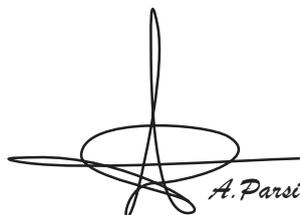

Name:

Ashkan Parsi

# Abstract

Many types of ventricular and atrial cardiac arrhythmias have been discovered in clinical practice in the past 100 years, and these arrhythmias are a major contributor to sudden cardiac death. Ventricular tachycardia, ventricular fibrillation, and paroxysmal atrial fibrillation are the most commonly-occurring and dangerous arrhythmias, therefore early detection is crucial to prevent any further complications and reduce fatalities. Implantable devices such as pacemakers are commonly used in patients at high risk of sudden cardiac death. While great advances have been made in medical technology, there remain significant challenges in effective management of common arrhythmias. This thesis proposes novel arrhythmia detection and prediction methods to differentiate cardiac arrhythmias from non-life-threatening cardiac events, to increase the likelihood of detecting events that may lead to mortality, as well as reduce the incidence of unnecessary therapeutic intervention. The methods are based on detailed analysis of Heart Rate Variability (HRV) information.

A range of features based on HRV analysis are investigated, including features from time, frequency, bispectrum and nonlinear analysis, and a range of classification techniques is used for prediction based on these features. Firstly, a thorough review of existing methods for ventricular arrhythmia prediction is conducted, including a detailed comparative experimental study. Following this, feature extraction for ventricular arrhythmia prediction is investigated using feature ranking methods based on mutual information. Using classification techniques such as support vector machine, k-nearest neighbour and random forests, the proposed approaches compare well with related work in the literature using different signal analysis durations. Furthermore, for paroxysmal atrial fibrillation prediction, seven novel features are proposed and investigated using a number of standard classification techniques. Using only the seven newly-proposed features, classification performance outperforms those of the classical state-of-the-art feature set, and the results further improve when the features are combined with several of the classical features.

The results of the work show good performance of the proposed methods and support the potential for their deployment in resource-constrained devices for ventricular and atrial arrhythmia prediction, such as implantable pacemakers and defibrillators.

# Acknowledgments

I would like to thank the following people, without whom I would not have been able to complete this research. First and foremost, I would like to thank my primary supervisor, Prof. Edward Jones, and co-supervisor Prof. Martin Glavin for all their help, direction, and guidance throughout my PhD. Your continued interest in my work and professional development is greatly appreciated and your unwavering enthusiasm was invaluable to me. Thank you for believing in me.

Without the support of Irish Research Council (IRC), this project would not have been possible, so I thank them for backing up my work (Grant number GOIPG/2016/1604). I would also like to thank all the members of my Graduate Research Committee at NUI Galway, Prof. Peter Corcoran, Dr Maeve Duffy, and Dr Ted Vaughan. To all the staff of Electrical and Electronic Engineering Mary Costello, Myles Meehan, Martin Burke, Liam Kilmartin, Dr Adnan Elahi, Prof. Martin O'Halloran, and Prof. Gearóid Ó Laighin, thank you for always supporting me.

I would like to express my gratitude to my wife, SaBa, for listening to my boring daily challenges, providing guidance, and helping to step one at a time every day. Without her tremendous understanding and encouragement in the past few years, it would not be impossible for me to complete my study.

Special thanks to Dr Dallan Byrne and Dr Declan O'Loughlin for fruitful collaboration and support in the conduct of this research. Thanks also to Dr Darragh Mullins and Dr Bárbara Luz Oliveira for asking the right questions, for challenging my thought through my research in the best possible way, and their help along the way.

I would like to thank Dr Hossein Javidnia and Dr Shabab Bazrafkan for all the support, guidance, and laughter specially during the start of my PhD journey. You guys made it much more enjoyable for me, I finally did it!

I have been lucky to be surrounded by so many great friends who made brunch, lunch, and tea such a pleasure, not least. I am particularly grateful for it and would like to thank Dara, Dave, Eoghan, Esteban, Hao, Jibran, Jordan, Joseph, Kevin, Luke, Richie, Shane, Stephen, Tim, Tomás, Toto and many past and current researchers in the Alice Perry Engineering Building. Many thanks to everyone in Galway Climbing Co-op and moreover, to all my friends from throughout my time in Galway including Adib, Afrooz, Alessio, Amin, Amir, Amirali, Anna, Alireza, Behrouz, Charlie, Christopher, Colm, Cynthia, David, Davood, Dennis, Hoda, Jajo, Jakub, Julia, Kambiz, Laura, Luison, Mahmoud, Mahshid, Malcolm, Mariusz, Mona, Neda, Sahar, Santi, Sara, Shadan, Shideh, Shirin, Padhraic, Pat, Patrick, Ronan, Tarlan, Théo, Tohid. You have made my life better, and I am lucky to have you guys here!

Finally, to my mother, Nabegheh, my father, Sadegh, my sister, Niusha, my brother Kasra and extended family, thank you for being there for me and for supporting me in every possible way during this time. I truly cannot thank you enough for your endless support, encouragement, and love.

Ashkan Parsi
Spring of 2021.

# List of Figures

# List of Tables

# List of Acronyms

| | |
|---|---|
| **AED** | Automated External Defibrillator |
| **AF** | Atrial Fibrillation |
| **AFPDB** | Atrial Fibrillation Prediction Database |
| **ANS** | Autonomic Nervous System |
| **ATP** | Antitachycardia Pacing |
| **AV** | Atrioventricular |
| **CNN** | Convolutional Neural Network |
| **CON** | Normal/Controlled |
| **CRM** | Cardiac Rhythm Management |
| **ECG** | Electrocardiogram |
| **EGM** | Electrogram |
| **HOS** | Higher Order Spectral |
| **HR** | Heart Rate |
| **HRV** | Heart Rate Variability |
| **ICD** | Implantable Cardioverter Defibrillator |
| **IHR** | Instantaneous Heart Rate |
| **ILFS** | Infinite Latent Feature Selection |
| **KDE** | Kernel Density Estimation |
| **k-NN** | K-Nearest Neighbours |
| **ME** | Mixture of Experts |
| **MLP** | Multilayer Perceptron |
| **mRMR** | Minimal Redundancy-Maximal Relevance |
| **Na$^+$** | Sodium Ions |
| **N-N** | Normal-to-Normal |
| **PAC** | Premature Atrial Complex |
| **PAF** | Paroxysmal Atrial Fibrillation |
| **PSVT** | Paroxysmal Supraventricular Tachycardia |
| **RF** | Random Forest |
| **ROC** | Receiver Operator Characteristic |
| **SVM** | Support Vector Machine |
| **SVT** | Supraventricular Tachycardia |
| **VF** | Ventricle Fibrillation |
| **VT** | Ventricular Tachycardia |
| **WPW** | Wolff-Parkinson-White |

# CHAPTER 1
# Introduction

Cardiovascular diseases, especially those which can result in sudden cardiac death, remain a major cause of mortality globally [1]. Up to 80% of sudden cardiac deaths are caused by ventricular arrhythmias [2], and the accurate and reliable prediction of ventricular arrhythmias has the potential to tangibly impact the survival rate in patients with cardiovascular disease; in particular, the successful prediction and management of both ventricular tachycardia (VT) and ventricle fibrillation (VF) will increase the survival rate in patients with cardiac issues. Their distinctive characteristics motivate the use of machine learning to distinguish these rhythms from normal cardiac activity.

Heart rate variability (HRV) calculated from cardiac electrical activities is a key indicator of an individual's cardiovascular condition. Assessment of HRV has been shown to aid diagnosis and treatment strategies and is widely used in clinical practice. However, the variety of HRV estimation and analysis methods in cardiac arrhythmia prediction and detection indicate that there is a need for a more rigorous investigation of HRV features. This thesis investigates the development of appropriate HRV signal processing techniques in the context of ventricular and atrial arrhythmias prediction, with a particular focus on techniques that may be appropriate for use in implantable defibrillators.

## 1.1 Motivation

Currently, automated external defibrillators (AEDs) and implantable cardiac defibrillators (ICDs) are the most effective therapeutic approaches for those at high risk of sudden cardiac death due to cardiac arrhythmias [3]–[5]. An AED is a portable device which includes a control algorithm that checks the surface electrocardiogram (ECG) and can send an electrical shock in order to restore normal cardiac rhythm if either VT or VF are detected by its algorithm [6]. ICDs are battery-operated devices placed under the skin in the chest that constantly monitor cardiac activity through thin leads directly lodged into the right ventricle, which produce an electrogram (EGM). ICDs are the more usual treatment for patients with substantial daily risk of severe arrhythmia and are highly effective in decreasing mortality due to cardiac arrhythmias in high-risk patients [7], [8]. By processing intra-cardiac electrical activities, ICDs deliver treatments such as electrical shock, if necessary, to restore normal heart rhythm. ICDs have been proven to reduce the mortality of patients with life-threatening ventricular arrhythmias [9] and have become the therapy of choice for many patients, however, delivery of inappropriate shocks caused by misclassification of events is problematic. For example, mistaking rapidly conducted supraventricular tachycardia (SVT) affects 8% to 40% of patients [10]–[12]. Inappropriate shocks can lead to pain, anxiety, depression, impaired quality of life, proarrhythmia[1], and poor tolerance of life-saving ICD therapy [13].

ICDs may apply one of two treatment mechanisms in a situation of arrhythmia event recognition:

1. Direct current defibrillation (electrical shock)
2. Antitachycardia pacing (ATP).

Because of the somewhat traumatic nature of electric shock therapy needed to address arrhythmias; it is important to ensure that such therapy is only applied when needed. Appropriate therapy delivery is dependent upon accurately classifying a patient's rhythm in ICDs, however, as noted above, misclassification of supraventricular arrhythmias like SVT have been shown to lead to inappropriate therapy delivery in ICD recipients. For example, studies described in [14], [15] indicated that "inappropriate" detection of SVT can be as high as 34.5% of all detection in both types of ICDs, single and dual-chamber, of which approximately 53% were atrial fibrillation (AF) or more specifically paroxysmal atrial fibrillation (PAF) episodes. The MADIT II study in [16] shows inappropriate electric shocks were applied on 11.5% of 719 patients, constituting 31.2% of all electrical shocks applied in therapy overall [16]. While there is no specific guidance on what the minimum required detection accuracy for clinical practice is, it is clear that current ICD usage results in unacceptably high rates of false positives, leading to inappropriate therapy, therefore, a significant increase in detection accuracy should result in improved outcomes. A robust method would be one that combines accurate prediction with detection, and where appropriate, this may be used to trigger ATP which is a less aggressive therapy that controls heartbeats. In particular, this thesis considers ventricular arrhythmias which could be prevented by nonaggressive treatment like ATP, with more invasive electrotherapy reserved only for cases that absolutely require it. For example, the PainFree I and II trials showed the effectiveness and safety of the application of ATPs (2 sequences of 8-pulse burst pacing train at 88% of the VT cycle length) before shock delivery even on fast VT with a heart rate of 188-250 beats per minute [17]–[19].

---

[1] Proarrhythmia is a new or more frequent occurrence of pre-existing arrhythmias, paradoxically precipitated by antiarrhythmic therapy, which means it is a side effect associated with the administration of some existing antiarrhythmic drugs, as well as drugs for other indications.

On the other hand, a major drawback of therapy with an implantable defibrillator is the low specificity of detection [20]. In principle, adding atrial sensing information to a decision algorithm could improve specificity of detection especially with current advances in dial chamber ICDs which can record atrial and ventricular activities separately. As shown in Figure 1.1, about 50% of misclassification is due to AF. AF occurs when the heart beats in a disorganized and irregular way, which can lead to a range of symptoms and potential complications, and this irregular beating could be misclassified as VT-VF arrhythmias by ICDs.

Inappropriate Detection Share

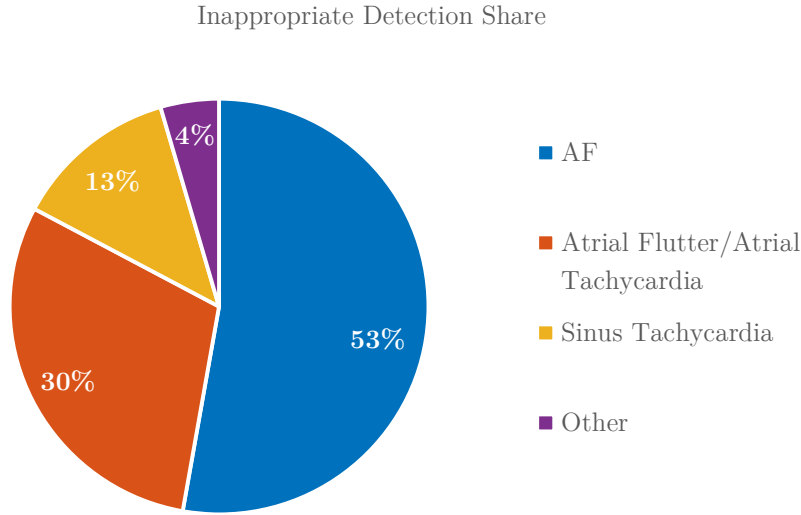

**Figure 1.1.** Rate of inappropriate detection of SVT (totally 34.5% of all detection in both types of ICDs, single and dual-chamber), with break-down by arrhythmia subtype (AF, atrial flutter/atrial tachycardia, sinus tachycardia, and the other. "Other" arrhythmias include junctional tachycardia, AV nodal re-entrant tachycardia, and AV re-entrant tachycardia. Subtype classification was based on blinded episode reviewer [15].

The most desirable method of limiting inappropriate therapies is appropriate rhythm prediction and discrimination by the ICD [15]. Many cardiac event detector algorithms have been presented to distinguish VT-VF from AF, SVT, and normal sinus rhythm. Generally, these algorithms are sub-divided into two main categories: algorithms based on the EGM morphology, and algorithms based on heart rate interval statistics. Algorithms based on EGM morphology use either direct/figurative models of heartbeat (QRS complex) processed using wavelet transform [21], [22], covariance-based model [23], spatial projection of tachycardia [24], or an indirect/implicit model of heartbeat such as similarity measures, which are extracted from information theory concepts [25]. Unlike morphological algorithms, which are generally more intensive computationally, the second category of algorithms utilize computationally-lighter features such as rate, onset, stability and entropy from HRV signal [4], [26]–[28]. All proposed algorithms have strengths and weaknesses however, to date, no consensus exists concerning the detection zones (different levels of heart rate) of VT-VF arrhythmias to decreasing inappropriate shocks due to misclassification of SVT and VT-VF arrhythmias. The second category of algorithms, based on heart rate, is the primary focus of this thesis.

Heart rate (HR) represents the number of contractions per minute that occurs in the heart. Changes in heart rate, which is called heart rate variability (HRV), occurs because of the rising and falling of heart rate caused by various factors such as human activity or cardiovascular related diseases. The irregularity of the time interval between human heartbeats was first noted in the early 1600s, however, its physiological importance was not fully appreciated until 1965 when Hon and Lee [29] found that the changes in pattern of foetal HR were an indicator of distress. In the late 1980s, the clinical importance of HRV became apparent when it was confirmed that the HRV was a strong and independent predictor of mortality following an acute myocardial infarction [30]. Today, there is much research using HRV as a biological marker to diagnosis various cardiovascular related diseases such as arrhythmia detection or prediction [31], [32], diabetes [33], [34], and heart failure [35]. Such interest arises because researchers have shown that HRV is one of the most promising markers to assess autonomic nervous system (ANS) activity, which can be directly correlated to cardiovascular disease [36]. At the same time, with massive improvements and advances in consumer and wearable technology, HRV data can be calculated using smart watches or small portable chest straps, which makes it more accessible for diagnosis [37], [38]. Existing works on HRV analysis and their limitations will be addressed in Chapter 3.

In the context of the field outlined above, this thesis explores techniques for signal processing of HRV from motivation to application, in an attempt to develop robust methods for HRV analysis that may be applied in implanted devices. A particular focus of the work is on minimising the number of features and HRV parameters to predict mainly VT, VF, and AF arrhythmias.

## 1.2    Thesis Contributions

The thesis aims to contribute first to an improved understanding of different types of ventricular and atrial arrhythmias, in both ischemic and nonischaemic heart disease and secondly to develop an HRV-based analysis method to predict and differentiate them from normal cardiac activities. This will contribute to the development of novel, more effective therapies for implanted devices such as ICDs. A further benefit of automatic prediction and detection of cardiac arrhythmias is the ability to develop time- and cost-effective clinical management procedures during a patient's short-time visit to clinics. To achieve these targets, optimized HRV-based prediction methods have been proposed to overcome weaknesses of existing arrhythmia prediction studies. Before listing contributions of this thesis, it is important to point out the scope of this work:

1. The main cardiac arrhythmias which have been included in this thesis are onset of VT, VF and AF.
2. The data that have used in this work are coming from standard publicly-available database where a fair comparison with other work is possible.
3. HRV is the main input of all proposed prediction and detection methods.
4. All algorithms have been implemented in MATLAB 2018 using additional toolboxes and libraries.

The primary contributions of the work are as follows:

1. A comprehensive comparative study on the performance of HRV features in time domain, frequency domain, bispectrum, and nonlinear analysis in VT-VF prediction. As many as fifty HRV features and their performance in VT-VF prediction system are analysed, and a prediction study based on mixtures of these feature is carried out, achieving competitive results in comparison with the literature.
2. The application of feature selection techniques to HRV signals to predict VT and VF with a reduced set of features, with performance comparable to using an exhaustive set of features. This work proposes low-complexity pre-processing using a mix of wavelet transform and median filter. The features proposed in this study achieved a more balanced distribution of false positive and false negative results than related methods in the literature, using as few as six features which is the lowest published within this field.
3. Seven novel HRV features for onset of PAF detection are proposed, and evaluated in comparison with the state of the art. The method proposed in this work provides an improvement of 3% in accuracy over the leading published results in literature using the same classifier.

## 1.3     Thesis Structure

The structure of the rest of the thesis is as follows. Chapter 2 presents human heart anatomy, electrical functionality, and cardiac arrhythmia as heart abnormal behaviour. Chapter 3 summarises HRV and classical features in time domain, frequency domain, bispectrum, and nonlinear analysis.

Chapter 4 presents a full review of existing VT-VF prediction methods based on HRV signals from ICDs, highlights current issues, and discusses some directions and motivations for the research presented later in this thesis. It describes current research approaches, features and databases used to address the problem. A comparative study on the prediction performance of a range of features, using a common database, is also presented.

Chapter 5 proposes a feature selection technique to reduce the computational overhead of prediction. Signal buffer lengths of 1-minute and 5-minute duration are examined, feature extraction and selection are applied, and evaluation is carried out. Learning and evaluation results using cross-validation are compared with other studies in the literature.

Detection of PAF, a cardiac arrhythmia that can eventually lead to heart failure, is presented in Chapter 6. Early detection of PAF is crucial to prevent any further fatal cardiac arrhythmia such as VT or VF. Seven novel HRV-based features are presented in this chapter and analysed using feature ranking methods. To evaluate these features, they are applied to four standard classification techniques and their performance is compared to the existing state of the art from the literature. Finally, in Chapter 7 the summary, conclusion, and future work of this thesis are presented.

## 1.4    Publications

### 1.4.1    Peer-reviewed Journal Publications

### 1.4.2    Peer-reviewed Conference Publications

In this chapter, the fundamental basis of human cardiac signals relevant to this thesis is discussed. Firstly, an overview of the anatomy and electrophysiology of the human heart is presented. This includes consideration of the ECG and HRV, which are widely used in clinical practice for cardiac health monitoring [39], [40]. Following this, cardiac arrhythmias which have been studied in this research are covered. Finally, in the last part of the chapter, two commonly-used cardiac rhythm management systems, automated external defibrillator, and implantable cardioverter defibrillator, are discussed. Platforms like this are the main intended target application of the prediction algorithms developed in this thesis.

## 2.1  Cardiac Anatomy and Electrophysiology

The human cardiovascular system is responsible for circulating blood throughout the human body. The heart is located in the mediastinum, a region that extends from the sternum (breast-bone) to the vertebral column in the dorsoventral aspect, from the first rib to the diaphragm in the anterior-posterior aspect, and between the lungs. Heart has an electrical conduction system which makes your heart pump blood around your body.

The cardiovascular system is a closed circulatory system mainly composed of the heart, blood vessels and blood. Its function is to transport nutrients, oxygen, carbon dioxide, hormones, and blood cells throughout the organism in order to provide nourishment, protect from diseases, regulate body temperature and pH, and maintain homeostasis.

### 2.1.1  Anatomy of Human Heart

The heart is composed of four chambers [41]:

- Two superior receiving chambers called atria (left and right), which receive blood coming back to the heart from all over the body.

- Two inferior pumping chambers called ventricles (left and right) that pump blood out of the heart for supply to all parts of the body.

Generally, the right chambers, the right atrium and right ventricle are responsible for collecting deoxygenated (oxygen-poor) blood from all parts of the body and supplying it to the lungs for oxygenation. The left chambers are responsible for collecting oxygenated (oxygen-rich) blood from the lungs through the pulmonary vein and supplying it all over the body through the aorta. The right atrium receives deoxygenated blood from the body through the superior vena cava and inferior vena cava; while oxygenated blood arrives to the left atrium coming back from the lungs, through the pulmonary veins.

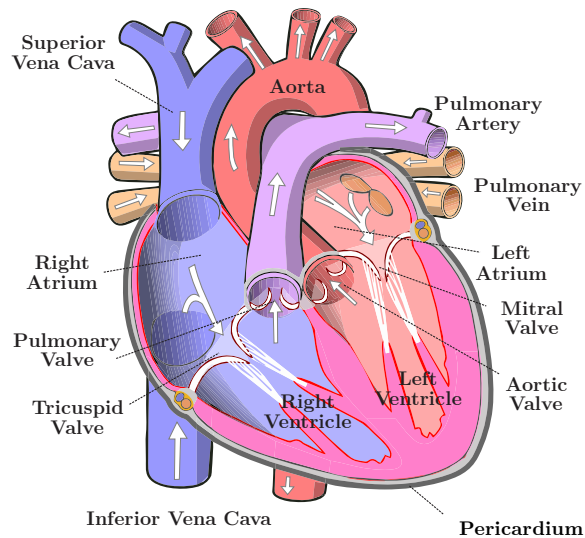

**Figure 2.1.** Anatomical structure of the heart [42].

Figure 2.1 displays the main anatomical structures of the heart. Atrioventricular valves connect atria and ventricles, ensuring unidirectional blood flows. The tricuspid valve separates the right atrium from the right ventricle, whereas the mitral valve separates the left atrium from the left ventricle. Two additional semilunar valves are located at the exit of each ventricle to prevent blood backflow, one on the aorta for the left ventricle and one on the pulmonary artery from the right ventricle. The flow of blood in and out of these chambers is controlled by the valves opening and closing in response to the pressure changes as the heart contracts and relaxes during a heartbeat. This contraction of the heart muscle is in turn modulated by the flow of electrical signals through the muscle [41].

### 2.1.2    Electrocardiogram Signal

The electrocardiogram signal captures the electrical activity of the heart over a period of time. The cardiac electrical system is based on the ability of cardiac cells to contract in response to the depolarization of the cells. At rest, the cardiac cell has less positive charge inside than outside so the net electric potential in the cell is negative. As the cell is stimulated, it goes into an excited state and in response it allows sodium ions ($Na^+$), among a series of other positively charged ions, to diffuse into the cells; this entry of positive ions creates a positive charge inside the cell. This permeation process is known as depolarization and as depolarization occurs, the cardiac cell contracts. There is selective permeability of ions back and forth through cardiac tissue which have special ion channels. To repeatedly depolarize, cardiac tissue must return to its resting state; this return is called the repolarization [43].

The continuous cycle of depolarization and repolarization is the essential part of the cardiac electrical system. This cycle is conducted from cell to cell, causing an electric current to propagate. Depolarization of the atria and ventricles creates the contraction of the atria and ventricles. This wave of depolarization goes through the heart in the following path (Figure 2.2):

1. Sinoatrial SA Node
2. Atrioventricular (AV) Node
3. Bundle of His
4. Purkinje Fibres

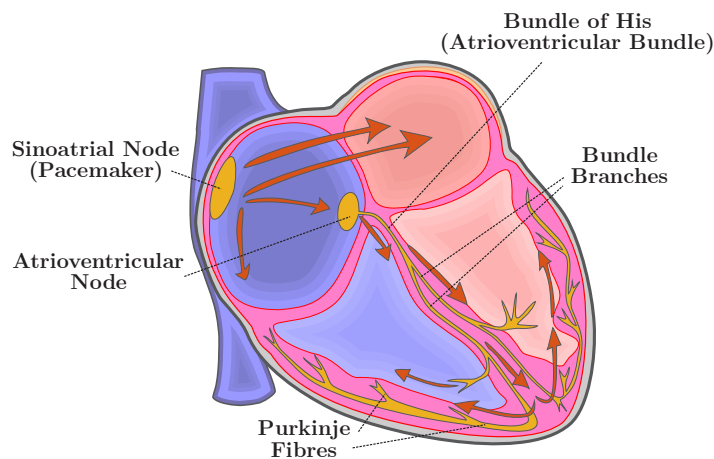

**Figure 2.2.** Heart conduction system.

On every heartbeat, the electrical conduction begins in the SA node in the right atrium which acts as a natural pacemaker. From here it is transferred to the AV node. From AV node the impulse propagates through the bundle of His (the only electric connection between the atria and the ventricles) and then into specialized Purkinje fibres, spreading throughout the ventricles muscle [44]. This conduction pathway is illustrated in Figure 2.2.

This cellular depolarization and subsequent repolarization or conduction are recorded as the ECG. To measure ECG, typically several electrodes are placed at several strategic positions over the heart. The electrodes detect the tiny electrical changes that occur because of the depolarization of the heart muscles during each heartbeat. The output from these electrodes is recorded over time and the output graph of electrical voltage (y-axis) versus time (x-axis) is the electrocardiogram (Figure 2.3).

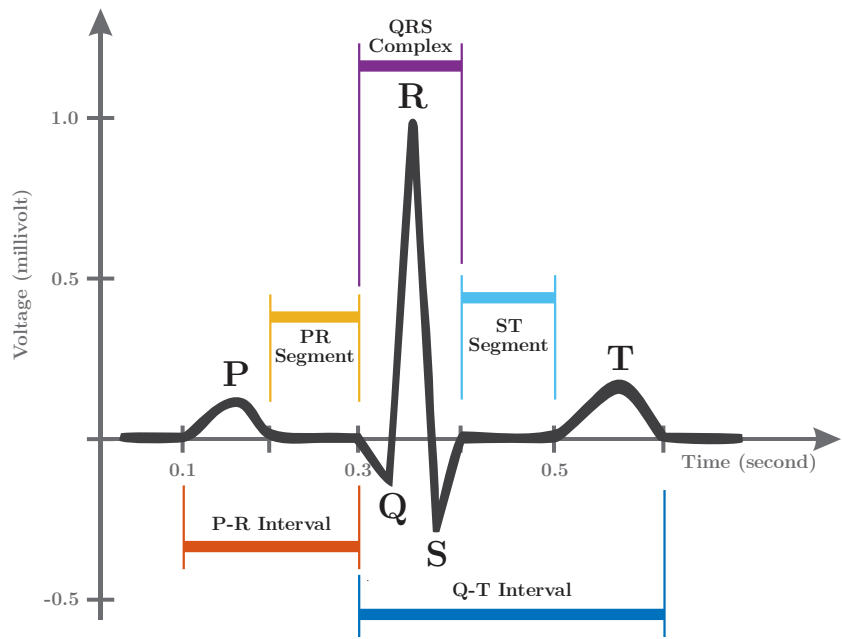

**Figure 2.3.** Main wave, interval and segments in normal ECG signal.

A typical ECG is depicted in Figure 2.3 and is composed of three main parts: P-wave (corresponding to atrial depolarization), the QRS complex (ventricular depolarization) and the T wave (ventricular repolarization). The ECG reflects the transfer of ions through the myocardium, which changes in each heartbeat. The isoelectric line is the baseline voltage of ECG which is traced following the T wave and preceding the next P-wave. The P-wave is a small upward deflection on the ECG. The P-wave represents atrial depolarization, which arises in the SA node and spreads through contractile fibres in both atria. After the P-wave begins, the atria contract. During the plateau period of steady depolarization, the ECG tracing is flat. The QRS complex as a second wave, represents rapid ventricular depolarization, as the action potential reaches the AV node and spreads through the bundle branches and Purkinje fibres to all the ventricular contractile fibres. This causes the ventricles to contract shortly after the QRS complex appears and continues in the ST segment. At the same time, atrial repolarization is occurring, but it is not usually evident in an ECG because the larger QRS complex masks it. The final wave, which is called the T wave, is produced in the resting state of the ventricles. It indicates ventricular repolarization and occurs just as the ventricles are starting to relax. The T wave is smaller and wider than the QRS complex because repolarization occurs more slowly than depolarization. The repetitive cycle of the electrical activity of heart is represented by the P-QRS-T sequences. The ECG is a key diagnostic tool used to assess the health conditions of the heart. In reading an ECG, the amplitude of the waves and time intervals between the occurrence of the waves can provide clues to abnormalities. A summary description and typical durations for the different components of the ECG waveform is presented in Table 2.1.

**Table 2.1.** Typical ECG durations for healthy person [43], [44].

| Feature | Description | Duration (Second) |
|---|---|---|
| P-wave | During normal atrial depolarization, the main electrical vector is directed from the SA node toward the AV node and spreads from the right atrium to the left atrium. This turns into the P-wave on the ECG. | 0.8 |
| P-R interval | The P-R interval is measured from the beginning of the P-wave to the beginning of the QRS complex. The P-R interval reflects the time the electrical impulse takes to travel from the SA node through the AV node and entering the ventricles. The P-R interval is, therefore, a good estimate of AV node function. | 0.12-0.2 |
| PR segment | The PR segment connects the P-wave and the QRS complex. The impulse vector is from the AV node to the bundle of His to the bundle branches and then to the Purkinje fibres. This electrical activity does not produce a contraction directly and is merely traveling down toward the ventricles, and this shows up flat on the ECG. The P-R interval is more clinically relevant. | 0.5-0.12 |
| QRS complex | The QRS complex reflects the rapid depolarization of the right and left ventricles. The ventricles have a large muscle mass compared to the atria, so the QRS complex usually has a much larger amplitude than the P-wave. | 0.8-0.12 |
| ST segment | The ST segment connects the QRS complex and the T wave. The ST segment represents the period when the ventricles are depolarized. It is isoelectric. | 0.32 |
| Q-T interval | The Q-T interval is measured from the beginning of the QRS complex to the end of the T wave. A prolonged Q-T interval is a risk factor for ventricular arrhythmias and sudden cardiac death. It varies with heart rate and needs a correction for clinical relevance. | Up to 0.42 seconds in heart rate of 60 bpm |
| T wave | The T wave represents the repolarization (or recovery) of the ventricles. The interval from the beginning of the QRS complex to the apex of the T wave is referred to as the absolute refractory period. The last half of the T wave is referred to as the relative refractory period (or vulnerable period). | 0.16 |

### 2.1.3 Heart Rate Variability

Heart rate variability (HRV) was first used clinically in 1965 when Hon and Lee [29] noted that foetal distress was accompanied by changes in beat-to-beat variation of the foetal heart, even before there was detectable change in the heart rate. Heart rate (HR) is the number of heartbeats per minute, while HRV captures the temporal variation between sequences of adjacent heartbeats. It measures the time interval between instantaneous heartbeats calculated from cardiac R-R intervals (Figure 2.4). On a standard ECG, the maximum upwards deflection of a normal QRS complex is at the peak of the R wave (Figure 2.3), and the duration between two adjacent R wave peaks is termed the R-R interval. R-R intervals may reflect normal or pathological behaviour. The ECG signal requires editing before HRV analysis can be performed, a process requiring the removal of all non-sinus-node-originating beats. The period between adjacent QRS complexes resulting from normal sinus node depolarizations is termed the normal-to-normal (N-N) interval. HRV is the measurement of the variability of the N-N intervals [45].

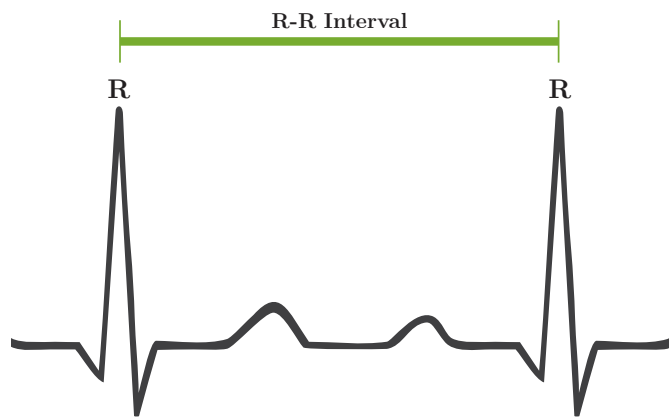

**Figure 2.4.** The interval between an R wave and the next R wave; normal resting heart rate is between 60 and 100 bpm. R-R Interval duration generally varies between 0.6-1.2 seconds [43].

HRV is a measure of the balance between sympathetic mediators of HR (i.e. the effect of epinephrine and norepinephrine, released from sympathetic nerve fibres, acting on the SA and AV nodes), which increase the rate of cardiac contraction and facilitate conduction at the AV node, and parasympathetic mediators of HR (i.e. the influence of acetylcholine, released by the parasympathetic nerve fibres, acting on the SA and AV nodes), leading to a decrease in the HR and a slowing of conduction at the AV node. HRV operate on different time scales to help us adapt to environmental and psychological challenges. It also reflects regulation of autonomic balance, blood pressure, gas exchange, gut, heart, and vascular tone, which refers to the diameter of the blood vessels that regulate blood pressure [46].

HRV analysis has been widely used in clinical practice for detecting or predicting life-threatening arrhythmias [47]–[50] for many years. In the 1970s, Ewing *et al.* [51] used short-term HRV measurements as a marker of diabetic autonomic neuropathy. In 1977, Wolf *et al.* [52] showed that patients with reduced HRV after a myocardial infarction had an increased mortality, and this was confirmed by other studies showing that HRV is an accurate predictor of mortality post-myocardial infarction [30], [53]. HRV reduces within 2 to 3 days after myocardial infarction, begins to recover within a few weeks, and is maximally but not fully recovered by 6 to 12 months [54].

Patients with persisting low HRV values have mortality almost three times greater than those with a normal HRV [55]. Higher HRV is not always better since pathological conditions can increase HRV. When cardiac conduction abnormalities elevate HRV measurements, this is strongly linked to increased risk of mortality. An optimal level of HRV is associated with health and self-regulatory capacity, and adaptability or resilience. Higher levels of resting HRV are linked to performance of executive functions like attention and emotional processing by the prefrontal cortex [56]. This signal as a main source of data for cardiac arrhythmias in the current research will be explained and investigated in the next chapter.

## 2.2   Cardiac Arrhythmia

Cardiac arrhythmia affects millions of people worldwide, causing almost half of deaths due to cardiovascular disease  [43], [57]. A cardiac arrhythmia is an abnormal or irregular heart rhythm. The normal heart rhythm is called normal sinus rhythm where the triggering impulses from the SA node propagate throughout the four chambers of the heart in a coordinated manner. Cardiac arrhythmia is the disruption of the normal electrical activity caused by an abnormality in that electrical conduction system. It can make the heartbeat too slowly, too quickly, or in an irregular way. Some cardiac arrhythmias are more serious than others which may lead to cardiac arrest and sudden cardiac death if not detected and left untreated [57].

There are many reasons why a person may develop a cardiac arrhythmia. For example, they are more common in older people, and in people with a heart condition such as coronary heart disease or heart valve disease. The symptoms of an arrhythmia will depend on what type of arrhythmia is present, and how it affects the functioning of the heart. A small change in cardiac normal sinus pace may lead to severe arrhythmia attacks, which can reduce the ability of the heart to pump blood and causes shorting of breath, pain in chest, dizziness, tiredness, and loss of consciousness. These are the most common symptoms of an arrhythmia.

Cardiac arrhythmia is of two main types:

1. Bradycardia[1], where the heartbeat is slower than 60 beats/min
2. Tachycardia, where the heartbeat is faster than 100 beats/min

In bradycardia, typically, no symptoms are visible till a heart rate of about 50 beats/min. Low heartbeat rates are common during rest and also among certain highly trained athletes. The heart muscle of athletes has become conditioned to have a higher stroke volume and, so, requires fewer contractions to circulate the same volume of blood. Bradycardia is less likely to be congenital and older patients are most often affected. Shortness of breath, fatigue, and dizziness are common symptoms associated with bradycardia, severe cases may result in fainting. Bradycardia does not usually need treatment [58]. However, if it is due to an underlying medical condition – such as an underactive thyroid gland – treatment may be needed for that condition.

In the other group, tachycardia[2] also called tachyarrhythmia is the one that needed to be detected and treated in a more immediate manner as it is potentially more serious. In general, a resting heart rate over 100 beats/min is classified as tachycardia in regular adults. As the heart rate is controlled by electrical signals sent across heart tissues, tachycardia occurs when an abnormality in the heart produces rapid electrical signals. When the rate of the heart is too rapid, it may not effectively pump blood to the rest of the body, including the heart itself which provides less blood flow. This ends up depriving an individual's organs and tissues of oxygen. The high heart rate or tachycardia can cause symptoms such as, dizziness, shortness of breath, light headedness, rapid pulse rate, heart palpitations, chest pain, and fainting or syncope. Some people with tachycardia have no symptoms, and the condition is only discovered during a physical examination or with a heart-monitoring test such as ECG. There are many different structural abnormalities that can alter electrical signals and lead to faster heart rates. The common types of tachycardia which are the main focus of this thesis are covered in the following sub-sections.

---

[1] Bradycardia is derived from Greek, "brady" means slow and "cardia" means heart.
[2] Tachycardia is derived from Greek, "tachy-" means quick or rapid and "cardia" means heart.

### 2.2.1     Supraventricular Arrhythmia

Supraventricular arrhythmias start with fast heart rates. They begin in the areas above the ventricles, such as the upper chambers which are heart's atria, or the atrial conduction pathways. Generally, supraventricular or "atrial arrhythmias" are not as serious as ventricular arrhythmias and sometimes, they do not even require any treatment. Different types of supraventricular arrhythmias are SVT, paroxysmal supraventricular tachycardia (PSVT). AF, atrial flutter, and Wolff-Parkinson-White (WPW) syndrome.

Supraventricular arrhythmias can happen in response to a number of things, including tobacco, alcohol, or caffeine consumption, and cough and cold medicines. Supraventricular arrhythmias can cause shortness of breath, heart palpitations, chest tightness, and a very fast pulse.

#### 2.2.1.1     Supraventricular Tachycardia

Supraventricular[3] tachycardia is an overall term for any fast heart rhythm that starts from above the ventricles. SVTs are often 'paroxysmal', which means occasionally or from time to time. SVT happens when electrical signals in the heart's upper chambers fire abnormally, which interferes with electrical signals coming from the SA node, the heart's natural pacemaker. The beats in the atria then speed up the heart rate. SVT is a rapid, regular heart rate where the heartbeats anywhere from 150-250 beats/minute in the atria. This type of arrhythmia is more common in infants and young people. It is also more likely to occur in women, anxious young people, and people who are extremely tired (fatigued). People who drink a lot of coffee or alcohol or who are heavy smokers also have a greater risk. SVTs are quite common, but are rarely life-threatening [59].

#### 2.2.1.2     Atrial Fibrillation

Atrial fibrillation – or AF for short – is a fast, irregular rhythm where single muscle fibres in the heart twitch or contract. AF happens when different places in and around the atria fire off electrical impulses in an uncoordinated way [60]. These random electrical impulses result in rapid, uncoordinated, and weak contractions of the atria. The chaotic electrical signals also affect the AV node, and this usually results in an irregular rhythm of the ventricles. The origin of this chaotic activity is around the pulmonary veins [61].

---

[3] Supra means above.

AF is the most common type of arrhythmia affecting adults of any age [62]. In Europe and North America alone, about 3% of the population are affected by AF [63]. Approximately 4% of the population over 65 years old are affected by AF, and 10% of those older than 80 years old. People usually need treatment to control their AF. The type of treatment depends on several factors, including the type of AF. AF may cause blood to pool in the heart's upper chambers. The pooled blood can lead to the formation of blood clots. A stroke can occur if a blood clot travels from the heart and blocks a smaller artery[4] in the brain.

### 2.2.2    Ventricular Arrhythmia

Ventricular arrhythmias are fast, abnormal heart rhythms that start from the ventricles. Most ventricular arrhythmias are caused by underlying heart disease, and can often be life-threatening [64]. The two main types of ventricular arrhythmias are ventricular tachycardia (VT for short) and ventricular fibrillation (VF for short). VT and VF are the major causes of sudden cardiac death (up to 80%) in patients with structural heart disease [2]. Sudden cardiac death is defined as death from an unsuspected circulatory arrest, usually due to an arrhythmia occurring within an hour of the onset of symptoms [58].

#### 2.2.2.1    Ventricular Tachycardia

Ventricular tachycardia - or VT for short - is a rapid heart rate that originates with abnormality of electrical impulses in the lower chambers or ventricles of the heart. This rapid heart rate does not allow the ventricles to completely fill with blood as well as contract fully to pump enough blood to the body (Figure 2.5). Since the new electrical impulses in the lower chambers do not move through the heart muscle along the regular route, the heart muscle does not beat normally.

VT is technically defined as a series of three or more consecutive premature ventricular impulses with QRS duration > 120 milliseconds at rate between 100 and 300 beats/minute. In this state, the QRS complex is not clearly present in the ECG signal. VT may be monomorphic or polymorphic. Monomorphic VT consists of a stable (organized) single QRS morphology often with a left bundle or right bundle block pattern. Polymorphic or multiform VT exhibits an irregularly changing QRS configuration with variable cycle lengths between

---

[4] Cerebral arteries

600 and 180 milliseconds. VT may be sustained (duration > 30 seconds) and/or require termination due to hemodynamic compromise in less than 30 seconds or non-sustained (three or more beats but less than < 30 seconds) [58]. VT is often a life-threatening medical emergency and without heart rhythms treatment it will cause death within minutes [65].

### 2.2.2.2 Ventricular Fibrillation

Ventricular fibrillation - or VF for short - occurs when rapid, chaotic electrical impulses cause the ventricles to quiver ineffectively and so they do not function effectively as pumping chambers. This lack of pumping means the body and brain do not receive sufficient blood in time. VF produces a completely disorganized tachyarrhythmia usually faster than 300 beats/minute with random, asynchronous electrical activity of the ventricles [58]. There are no discrete QRS complexes.

Tachycardia brought on by VF can be very serious and can even be fatal unless treated very quickly. This is the most serious arrhythmia, causing cardiac arrest and most people who experience VF have an underlying heart disease or have experienced serious trauma [66]. In comparison with VT, instead of one misplaced beat from the ventricles, VF may have several impulses that begin at the same time from different locations, all telling the heart to beat (Figure 2.5). VF is more serious than VT and without immediate medical attention and treatment like appropriate electrical shock, it causes sudden cardiac death [67].

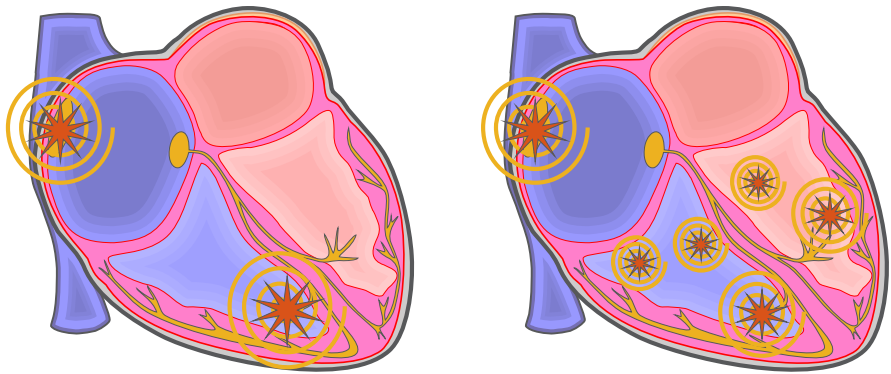

(a) Illustrating an example of VT with one misplaced electrical impulse in the left ventricle.

(b) Illustrating an example of VF with several misplaced electrical impulses in the left and right ventricle.

**Figure 2.5.** Examples of ventricular arrhythmias showing misplaced electrical impulses in lower side of the heart.

## 2.3 Cardiac Rhythm Management Approaches

Increasing incidences of arrhythmia and heart failure have fostered rapid advancements in cardiac rhythm management (CRM) devices and technologies, and associated clinical practice [67], [68]. Currently, automated external defibrillator (AEDs) as an out-of-hospital cardiac arrest treatment, and single-chamber pacemakers and/or implantable cardioverter defibrillator (ICD) as a permanent treatment, are the most efficient therapy for sudden cardiac death prevention [3]–[5].

The use of CRM devices has substantially surged over the past few decades due to increasing life expectancy and constant technological advancements that have extended their scope and abilities. With rising indications and accessibility to CRM around the world, the number of patients requiring such devices is subsequently increasing [69]. Such technologies have transformed clinical electrophysiology into one of the most rapidly expanding cardiology subspecialties. Pacemakers are the accepted standard of care for those with complex atrial and ventricular arrhythmias [69], [70]. In addition, technological innovations have resulted in the emergence of remote monitoring as an effective alternative to routine in-office patient care.

### 2.3.1 Automated External Defibrillator

An AED is portable device which includes a shock advice algorithm that checks the surface ECG and can send electrical shock if either VT or VF are detected by its algorithm [6]. These defibrillators are able to automatically detect the presence of VT and VF and perform cardioversion to cardiac arrest victims. When a cardiac arrest case is suspected, an AED operator can place electrodes on the patient's chest. The AED will detect the patient's condition, confirm the presence of a cardiac arrest, and decide if electric shock intervention is needed. The AED will also instruct the operator to carry out necessary steps for automated external defibrillation of the patient [71].

### 2.3.2 Implantable Cardioverter Defibrillator

An ICD is a battery-operated device placed under the skin in the chest that constantly monitor cardiac activities by using thin leads directly lodged into the right ventricle, and are the usual treatment for patients with substantial daily risk of severe arrhythmia [7]. By processing intra-cardiac electrical activities internally (called EGM), ICDs can decide to deliver electrical shock, if necessary, to restore normal heart rhythm. Cardioversion and defibrillation are both forms of high-energy shocks that stop dangerous arrhythmias and restore a normal heart rhythm. If the patient is conscious (awake) at the time of the shock, it is painful and usually described informally as feeling like a "kick in the chest". Most of the time, however, a shock for a dangerous arrhythmia is delivered after the patient has lost consciousness and is therefore unable to feel the shock.

Antitachycardia pacing – or ATP for short – is an alternative method of stopping ventricular arrhythmias and is available in most ICDs [7]. It involves delivering a short series (e.g., 5 to 10) of paced beats. This is not painful and may be unnoticed by the patient, although some patients may feel a brief burst of palpitations. ATP can be very effective for some slower ventricular arrhythmias (e.g., 150 to 200 beats/minute), and is often programmed as the initial therapy for these arrhythmias. High-energy shocks are often programmed as the initial therapy for very fast rhythms (e.g., more than 200 to 220 beats/minute), or as a rescue therapy if ATP treatment fails.

Nowadays, developing differential algorithms to recognize VT and VF arrhythmias from other cardiac rhythms is one of the most critical issues in ICD patients. The main goal of this process is to develop algorithms with highest possible sensitivity to avoid untreated arrhythmias and highest possible specificity to avoid inappropriate ICD shocks [14]. Nearly 80% of inappropriate shocks are caused by misclassification of supraventricular arrhythmias as a VT or VF [16], [72]. Supraventricular arrhythmias include SVT, AF, and atrial flutter, which do not need any immediate therapy.

## 2.4 Summary

This chapter has discussed some important aspects of the cardiac system and their relevance in the context of this thesis. Cardiac anatomy and electrophysiology were first discussed, including a description of the ECG waveform and HRV, which are central to this thesis. Following this, the main types of cardiac arrhythmia were presented. Finally, some current practice in the treatment of cardiac problems through external and implanted defibrillation devices were briefly considered. Such platforms can be considered a target application for the techniques developed later in this thesis.

The next chapter continues consideration of HRV analysis, through a detailed discussion of the main features that are typically used for quantitative analysis and diagnosis of arrhythmias.

# Heart Rate Variability Analysis

This chapter introduces HRV analysis principles, measurements, and features that have been used in cardiac arrhythmia prediction and detection research. HRV is a reliable reflection of the many physiological factors modulating the normal rhythm of the heart. It shows that the structure generating the signal is not only simply linear, but also involves nonlinear contributions. Over the last three decades, alterations in HRV have been found in patients with many cardiovascular conditions [36]. Since the introduction of CRM devices such as different type of ICDs, many studies have examined HRV processing, analysing effects such as the circadian rhythm in HRV signals [73] to analyse significant changes immediately prior to an episode of VTA [48], [74]–[76]. Some of the material in this chapter has been published in Parsi *et al.*, "Prediction of Sudden Cardiac Death in Implantable Cardioverter Defibrillators: A Review and Comparative Study of Heart Rate Variability Features," *IEEE Rev. Biomed. Eng.*, vol. 13, pp. 5–16, 2020 [77] and Parsi *et al.*, "Heart Rate Variability Analysis to Predict Onset of Ventricular Tachyarrhythmias in Implantable Cardioverter Defibrillators," in *41st Annual International Conference of the IEEE Engineering in Medicine and Biology Society (EMBC)*, Jul. 2019, pp. 6770–6775 [78].

## 3.1 Introduction to HRV

HRV measures variations in the heart rate and can reflect both the sympathetic and the parasympathetic components of the ANS. It is defined as the physiological variation in the duration of intervals between sinus beats and serves as a measurable indicator of cardiovascular integrity and prognosis. The instantaneous heart rate (IHR) signal is calculated from the R-R intervals recorded by e.g., an ICD using the following equation:

$$IHR \ (beats/min) = \frac{60{,}000}{RRI \ (msec)} \tag{3.1}$$

Where *RRI* (R-R intervals) is the time in milliseconds between instantaneous heartbeats as measured by any CRM devices, or as described in Chapter 2, the period between adjacent QRS complexes resulting from sinus node depolarizations, which is termed the N-N interval. The unit of IHR is beat/minutes (bpm). HRV features could be calculated from measurements of the R-R intervals (N-N intervals) or IHR signal (in this thesis, mostly IHR signals were used). By way of example, Figure 3.1 (a) shows the variations of the R-R intervals of normal (which is called CON) and abnormal (VT and VF) samples of a typical patient from the Medtronic version 1.0 dataset [79] which will be described in depth in Chapter 4. Figure 3.1(b) shows the corresponding IHR derived from these RRI traces. The significant variation in IHR for normal and abnormal conditions is evident[6].

Formal criteria for HRV and comparison of variables were developed by a joint task force between the European Society of Cardiology and the North American Society of Pacing and Electrophysiology in 1996 [45] and updated in 2015 [80]. Generally, the major features in HRV analysis can be divided in four main groups as described in the following subsections.

---

[6] VT and VF samples in these charts show cardiac activities 2 minutes before related tachyarrhythmias happening. The "CON" sample, on the other hand, was captured on a normal patient visit and check-up routine with specialist supervision.

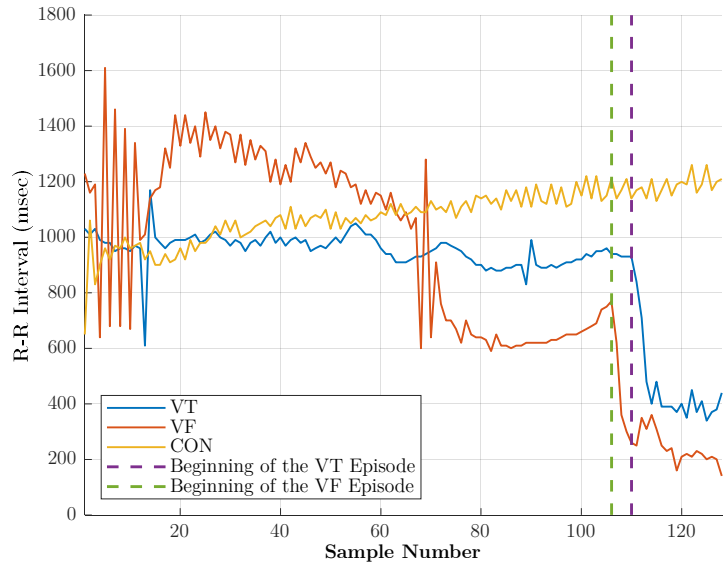

(a) R-R interval signals. VT, VF, and CON samples from Patient No. 8010 of
Medtronic version 1.0 dataset.

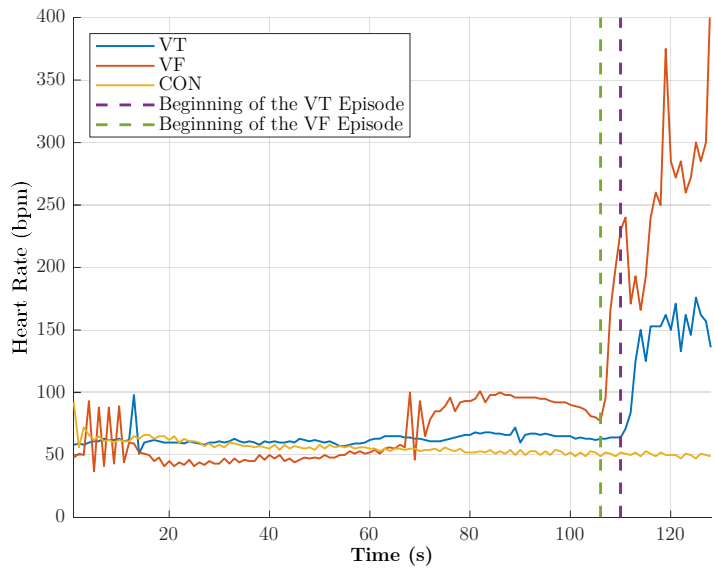

(b) IHR signals after conversion. VT, VF, and CON samples from Patient No. 8010 of
Medtronic version 1.0 dataset.

**Figure 3.1.** (a) RRI for normal, VT and VF conditions for a typical cardiac patient
[77]; (b) Corresponding IHR.

## 3.2 Time Domain HRV Analysis

Standard time domain parameters of HRV analysis are the simplest to extract and are valuable measures of variation of the R-R intervals [45]. The most commonly used time domain measures are the mean of N-N intervals or instantaneous heartbeats (MeanNN) and standard deviation (SDNN). Standard deviation is the square root of variance. Since variance is mathematically equal to total power of spectrum, SDNN reflects all the cyclic components responsible for variability in the period of recording [45]. Other time domain features include the root mean square of successive R-R intervals differences (RMSSD), number of adjacent R-R intervals differing by more than 50 milliseconds (NN50), and pNN50, the number obtained by dividing NN50 by the total number of N-N intervals [28], [81]–[84]. HRV triangular index (HRVTri), which is a total number of N-N intervals used in histogram analysis divided by the maximum of the signal histogram, is employed in HRV analysis to extract and evaluate geometrical feature [45], [84], [85]. As the value of HRVTri is dependent on the size of the bin in the histogram, the size should be specified if the HRV signal is resampled to different frequencies. Summaries of time domain HRV features have presented in Table 3.1.

**Table 3.1.** Summary of HRV features in time domain.

| Feature | Description | Unit |
|---------|-------------|------|
| MeanNN | Mean of N-N intervals | ms/bpm |
| SDNN | Standard deviation of all N-N intervals | ms/bpm |
| RMSSD | Root mean square of successive N-N interval differences | ms/bpm |
| NN50 | Number of adjacent N-N intervals differing by more than 50 ms | Count |
| pNN50 | Percentage of successive N-N intervals that differ by more than 50 ms | % |
| HRVTri | Total number of all N-N intervals divided by the height of the histogram of all N-N intervals measured on a discrete scale | --- |

ms = millisecond; bpm = beats/minute

## 3.3    Frequency Domain HRV Analysis

Although the time domain methods are computationally simple, many cannot discriminate between the sympathetic and the parasympathetic contributions in the HRV signals. Sympathetic activity is associated with the low-frequency modulation of the heart rate (0.04-0.15 Hz) while parasympathetic activity is associated with the higher frequency modulation (0.15-0.4 Hz). Therefore, frequency domain features can be used to separately estimate the sympathetic and the parasympathetic contributions in the HRV signals. This facilitates preventive intervention, such as ATP, before the onset of sudden cardiac death [85].

Many studies used average power of HRV signals across the entire frequency band in addition to the average power for specific frequency bands. The power in a given band is typically computed by integrating the power spectral density (PSD) estimate in the frequency range of interest, such as very low frequency (VLF) band (0.0-0.04 Hz), low frequency (LF) band (0.04-0.15 Hz), high frequency (HF) band (0.15-0.4 Hz) and comparing different ratio variation like LF/HF or VLF/HF [26], [84], [86]–[90]. These frequency domain features (0.04-0.4 Hz) can estimate the sympathetic and parasympathetic contributions to HRV signal, commonly reflected in changes in energy distribution prior to sudden cardiac death [85].

For example, studies have shown that there is an increase in spectral power at lower frequencies that can help predict the onset of VT and VF, and also that the spectrum power increases just prior to breakdown of normal rhythm to a ventricular arrhythmia [91]. Similarly, spectral power analysis using either Welch's method for periodogram spectrum estimates [28], [92]–[94] or fast Fourier transform, found similar changes prior to both VT and VF [31], [74], [87], [95], [96], regardless of processing time and computational complexity that each algorithm has [97], [98]. Selected HRV features in frequency domain are presented in Table 3.2.

**Table 3.2.** Summary of selected HRV features in the frequency domain.

| Feature | Description | Unit |
|---------|-------------|------|
| $PSD_{VLF}$ $PSD_{LF}$ $PSD_{HF}$ | Power spectral density (PSD) in VLF, LF, HF | $ms^2/bpm^2$ |
| $PSD_{LF/HF}$ | PSD LF/HF | --- |

ms = millisecond; bpm = beats/minute

## 3.4    Bispectrum Analysis

Poly-spectral analysis using different higher-order spectral (HOS) features uses spectral representations of higher order moments or cumulants of a signal [85] [99]. HOS, up to third-order cumulant, was employed to estimate the bispectrum from IHR data in [87]. The HOS features described in this section consider the phase of the signal in addition to magnitude from the spectral power described in the previous section [99]. Since the HRV signal is nonlinear and non-Gaussian in nature [100], [101], the bispectrum which employs up to the third order statistics can be used to reveal information not present in the spectral domain. Also, this feature has been employed to reveal the hidden information that cannot otherwise be observed by using regular power spectrum calculated from the second-order moment of the signal and to detect quadratic phase coupled harmonics arising from nonlinearities of the HRV signal[7], in the recent HRV analysis based research [84], [86], [87], [92], [102].

Bispectrum analysis may be carried out as follows. Let $x(n)$ represent a discrete, zero-mean random signal. The third order cumulant sequence $R(m,n)$ of the signal is identical to its third-moment sequence, and it may be calculated as follows:

$$R(m,n) = E\{x(k)\, x(k+m)\, x(k+n)\} \qquad (3.2)$$

Where $k$ is sample index, $m, n$ are shift values in samples and $E\{\dots\}$ denotes the expectation operation. The bispectrum $B(\omega_1,\ \omega_2)$ of $x(n)$ is defined as the two-dimensional Fourier transform of $R(m,n)$ as a follow:

$$B(\omega_1, \omega_2) = \sum_{k=-\infty}^{\infty} \sum_{k=-\infty}^{\infty} R(m,n)\, exp[-j(\omega_1 m, \omega_1 n)] \qquad (3.3)$$

with the condition: $|\omega_1|, |\omega_2| \leq \pi$ for $\omega = 2\pi f$ [103]. As the third-order moment and cumulant are identical, a bispectrum is the spectrum of these values.

The bispectrum is a symmetric function which has 12 symmetric regions however, there are three sub-bands inside the region of interest (ROI) of the bispectrum which can discriminate sympathetic and parasympathetic contents of HRV signals [92]. Figure 3.2 below indicates these sub-bands (inside the triangular area indicated with red line) as LF-LF (LL), (LF-HF) (LH), and HF-HF (HH). In previous work, features have been extracted from these three regions and the entire ROI to characterize HRV. The most commonly used features derived from the bispectrum are covered in the following sub-sections.

---

[7] Studies shows mechanisms involved in the regulation of cardiovascular system interact with each other in a nonlinear way.

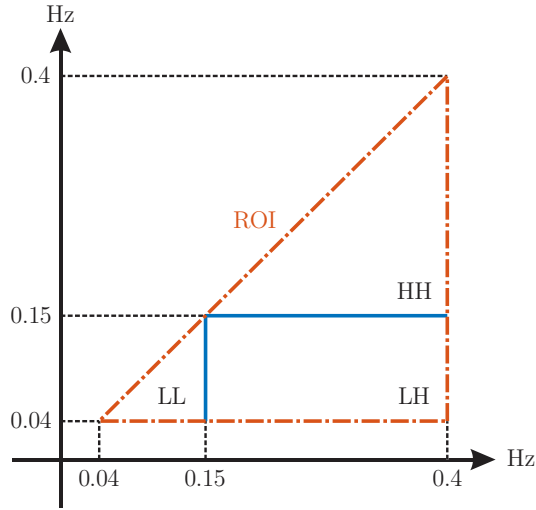

**Figure 3.2.** Segmentation of the bispectrum into sub band regions (LL, LH, and HH), and the ROI for bispectrum feature extraction, indicated by the red (with dash-dot line) triangle.

### 3.4.1   Magnitude and Power Average

Two heuristic features are the average magnitude ($M_{avg}$) and the average power ($P_{avg}$) in LL, LH, HH sub-bands, and the entire ROI. They are defined as follows:

$$M_{avg} = \frac{1}{N} \sum_{\Gamma} |B(f_1, f_2)| \qquad (3.4)$$

$$P_{avg} = \frac{1}{N} \sum_{\Gamma} |B(f_1, f_2)|^2 \qquad (3.5)$$

where $\Gamma$ represent one of four regions previously described (LL, LH, HH, and ROI in Figure 3.2) and $N$ is the number of points within $\Gamma$. These values represent fundamental bispectral properties of specific region.

### 3.4.2    Bispectrum Entropies

The mean magnitude of the bispectrum can be useful in discriminating between processes with similar power spectra but different third order statistics. However, it is sensitive to amplitude changes [104]. Entropy has been shown to be useful as a measure of "randomness" in signal analysis [105] and bispectral entropies have been successfully exploited to measure the regularity of nonlinear signals like EEG [106]. For any region $\Gamma$ from LL, LH, HH, and ROI, the normalized bispectrum entropy ($E_{nb}$) is defined as:

$$E_{nb} = \sum_m \left( \frac{|B(f_1, f_2)|}{\sum_\Gamma |B(f_1, f_2)|} \log \frac{|B(f_1, f_2)|}{\sum_\Gamma |B(f_1, f_2)|} \right) \tag{3.6}$$

and the squared normalized bispectrum entropy ($E_{snb}$) is defined as:

$$E_{snb} = \sum_m \left( \frac{|B(f_1, f_2)|^2}{\sum_\Gamma |B(f_1, f_2)|^2} \log \frac{|B(f_1, f_2)|^2}{\sum_\Gamma |B(f_1, f_2)|^2} \right) \tag{3.7}$$

Both normalized entropy measures $E_{nb}$ and $E_{snb}$ lie between 0 and 1.

### 3.4.3    Logarithmic Bispectrum Features

Linear amplitude scaling is often used to estimate the relative importance of spectral features, however, use of linear scaling can mask small, but potentially important features. Logarithmic scaling of the bispectrum has been shown to be effective in characterizing clinical EEG [107]. Two main logarithmic features have been used in HRV classification studies [84], [87], [92], [102], namely the sum of the logarithmic magnitude ($L_m$) and the sum of the logarithmic magnitudes of the diagonal elements of the bispectrum ($L_{dm}$), as an attempt to emphasize both larger and smaller components. These features are defined as:

$$L_m = \sum_\Gamma \log(|B(f_1, f_2)|) \tag{3.8}$$

$$L_{dm} = \sum_D \log(|B(f_1, f_2)|) \tag{3.9}$$

where D represents the diagonal elements of the log bispectrum of a region $\Gamma$ (one of LL, HH, and DOI; there are no diagonal elements in LH).

### 3.4.4 Weight Centre of Bispectrum

The weight centre of bispectrum (WCOB) represents the weight centre of the contour map of a bispectrum. This feature has been used in recent studies as a valuable parameter in HRV event detection methods [84], [87]. Researchers in ischemic EEG analysis have reported differences in the morphology of ischemic and normal EEG bispectral contour maps [101]. As cardiac abnormality might also show differences in the HRV bispectral contour map, WCOB has been adapted for use in previous studies by two frequency axes defined as follows:

$$WCOB_i = \frac{\sum_\Gamma i B(f_1, f_2)}{\sum_\Gamma B(f_1, f_2)} \tag{3.10}$$

$$WCOB_j = \frac{\sum_\Gamma j B(f_1, f_2)}{\sum_\Gamma B(f_1, f_2)} \tag{3.11}$$

for any $\Gamma$ in LL, LH, HH, and ROI where $i, j$ are the first and second frequency indices, respectively. Summaries of HRV bispectrum analysis features have presented in Table 3.3.

**Table 3.3.** Summary of bispectral HRV features.

| Feature | Description |
|---|---|
| $M_{avg}$ in LL, LH, HH, and ROI | Magnitude average of bispectrum in LL, LH, HH and ROI |
| $P_{avg}$ in LL, LH, HH, and ROI | Power average of bispectrum in LL, LH, HH and ROI |
| $E_{nb}$ in LL, LH, HH, and ROI | Normalized bispectrum entropy in LL, LH, HH and ROI |
| $E_{snb}$ in LL, LH, HH, and ROI | Squared normalized bispectrum entropy in LL, LH, HH and ROI |
| $L_m$ in LL, LH, HH, and ROI | sum of the logarithmic magnitude of the bispectrum in LL, LH, HH and ROI |
| $L_{dm}$ in LL, HH, and ROI | Sum of the logarithmic magnitudes of the diagonal elements of the bispectrum in LL, HH and ROI |
| $WCOB_i$ in LL, LH, HH, and ROI | WCOB$_i$ represents the weight center of the contour map of a bispectrum in LL, LH, HH and ROI for the first frequency index |
| $WCOB_j$ in LL, LH, HH, and ROI | WCOB$_j$ represents the weight center of the contour map of a bispectrum in LL, LH, HH and ROI for the second frequency index |

## 3.5    Nonlinear and Dynamic Analysis

Viewing HRV as an indirect electrical measure of autonomic heart rate modulations, influenced mainly by the permanent interplay between the two branches of the ANS and considering this as the output of a nonlinear system, can lead to better understanding of cardiac system dynamics [108]–[110]. Studies have also stressed the importance of nonlinear techniques to study HRV in issues related to both cardiovascular health and disease [111], [112].

Among them, the Poincaré plot has been demonstrated to be a simple and effective method for discriminating cardiovascular diseases based on HRV [113]. The Poincaré plot displays each R-R interval as a function of the previous R-R interval, which can indicate the degree of heart failure in a subject [114]. An ellipse can be generated to fit these data points. The width ($SD_1$) of this ellipse (minor axis), which is related to the fast beat-to-beat variability in the data, and the length ($SD_2$) of the ellipse (major axis) relate to the longer-term variability of those data [115]. These values can be directly calculated from two of the time domain features: 1) the standard deviation of *differences* between adjacent R-R intervals ($SDSD$), and 2) the mean of the standard deviation of all R-R intervals ($SDRR$):

$$SD_1 = \sqrt{\frac{1}{2}SDSD^2} \tag{3.12}$$

$$SD_2 = \sqrt{2SDRR^2 - \frac{1}{2}SDSD^2} \tag{3.13}$$

There are more features that have been extracted from Poincaré plot $SD_1/SD_2$ for detecting VTAs one hour before occurrence [94], as well as HRV density analysis (Dyx) based on the Poincaré plot [116]. Other symbolic dynamics to distinguish between different states of the autonomic interactions [96], [117] like Polvar10 (measures the incidence of 6 beat segments during which the beat-to-beat difference of R-R intervals is less than 10 ms) and Fitgra9 (finite time growth rates, dimension 9) regarding the complexity of the sinus node activity modulation system have been proposed and investigated in [47]. The results showed significant differences in the dynamic behaviour before onset of VT.

Entropy as a measure of signal complexity has also been applied. Sample entropy (SampEn) and approximate entropy (ApEn) [118] have been investigated to more strongly show HRV complexity and regularity changes in onset of ventricular arrhythmias [4]. The symbolic parameters, FWShannon, ForbiddenWords and Hjorth's parameters have also been applied [119].

FWShannon denotes the Shannon entropy of the word distribution whereas Forbidden Words stands for the number of words, which never or very seldom occur. Larger values of FWShannon reflect higher complexity whereas a high number of Forbidden Words reflect regular behaviour in the signal [49]. Conclusions can be drawn that when a ventricular arrhythmia approaches, the sympathetic tone of the patients is increased, and the complexity of their R-R intervals immediately before the onset of VT-VF events is lower than normal sinus rate and hence the values of entropy-based measures are significantly smaller for VT-VF episodes than those for normal series. Hjorth's parameters are used to capture the statistical properties of nonlinear time-series signals. Different Hjorth's parameters namely activity, mobility and complexity are used [120] which relate to the signal power, mean frequency, and change in frequency respectively [121].

Selected HRV features in nonlinear and dynamic analysis are presented in Table 3.4.

**Table 3.4.** Selected HRV features in nonlinear and dynamic analysis.

| Feature | Description |
|---|---|
| $SD_1$<br>$SD_2$<br>$SD_1/SD_2$ | Using two time domain measures, the standard deviation of differences between adjacent R-R intervals ($SDSD$) and the mean of the standard deviation of all R-R intervals ($SDRR$), $SD_1$ and $SD_2$ are calculated as presented in formula (3.12) and (3.13). |
| Sample entropy | Assume $X(t) = \{x_1, x_2 \dots, x_n\}$ represent the signal. The template vector of lenth $m$ defines such that $X_m(t) = \{x_i, x_{i+1} \dots, x_{i+m-1}\}$ and the distance function $d[X_m(i), X_m(j)]$ is to be Chebyshev distance. The Sample entropy define as: $SampEn = -\log A/B$, where<br><br>$A =$ number of temple vector pair having $d[X_{m+1}(i), X_{m+1}(j)] < r$<br>$B =$ number of temple vector pair having $d[X_m(i), X_m(j)] < r$<br>Generally, $m = 2$ and the value of $r$ would be $0.2 \times SD$ where $SD$ represent standard deviation of the input signal. |
| Rényi entropy | The Rényi entropy of order $\alpha$ where $\alpha \geq 0$ and $\alpha \neq 1$ is defined as: $\text{RényiEn}_\alpha(X(t)) = \frac{1}{1-\alpha}\log(\sum_{i=1}^{n} p_i^\alpha)$, where $X(t) = \{x_1, x_2 \dots, x_n\}$ represent the signal and corresponding probabilities $p_i = \Pr(X = x_i)$ for $i = 1, 2, \dots, n$ with the condition $\sum_{i=1}^{n} p_i = 1$. |
| Tsallis entropy | The Tsallis entropy of order $\alpha$ where $\alpha \geq 0$ and $\alpha \neq 1$ is defined as: $\text{TsallisEn}_\alpha(X(t)) = \frac{1}{\alpha-1}(1 - \sum_{i=1}^{n} p_i^\alpha)$, where $X(t) = \{x_1, x_2 \dots, x_n\}$ represent the signal and corresponding probabilities $p_i = \Pr(X = x_i)$ for $i = 1, 2, \dots, n$ with the condition $\sum_{i=1}^{n} p_i = 1$. |
| Hjorth's parameters | There are three parameters,<br><br>$Activity = var(x(t))$, $Mobility = \sqrt{var\left(\frac{dx(t)}{dt}\right)/var(x(t))}$, and<br><br>$Complexity = Mobility\left(\frac{dx(t)}{dt}\right)/Mobility(x(t))$<br><br>which represent the signal power, mean frequency, and change in frequency respectively, where $x(t)$ represents the signal and $var$ is the variance of time function. |

## 3.6     Discussion and Concluding Remarks

Originally, the aim of capturing HRV signals and investigating changes in HR were its use in the prediction of long-term survival in patients who had suffered myocardial infarction or had valvular or congestive heart disease [31]. Detecting HRV changes over a period of hours or even days requires a large volume of ECG data to be collected and analysed. This process traditionally was difficult and limiting for patients as Holter devices that record the ECG or in hospital monitoring were the only options. Also capturing dynamic changes in HRV in the period prior to arrhythmias events is harder to attain, due to the relative infrequency of such events. More recently, work has concentrated on attempts to predict the timing of onset of fatal arrhythmias like VT and VF as more CRM devices like advanced ICDs have been developed [31], [122]. Most of the cardiac arrhythmias such as ventricular arrhythmias have a circadian rhythm with increased frequency during the early morning (7–11 am) and early evening (6–7 pm) [73]. This is consistent with findings of a circadian distribution of cardiac arrests, with a low incidence at night, and peaks between 0800 - 1100 hours and 1600 - 1900 hours [123]. The evening peak may be attributable to VF, while the morning peak may be attributable to patients in non-VF and non-VT rhythms [124], [125]. The characteristics of HRV immediately prior to the onset of VTA, however, need more complex methods to be detected. Some studies report significant changes in HRV in the period immediately preceding a ventricular arrhythmia, and HR increasing prior to an episode of VT- VF is common [48], [74]–[76], [126].

Studies also show a significant change in HR, SDNN, and power in VLF, HF, and HF in patients who developed VT-VF sometimes one hour prior to the event [31], [47], and an alteration in the interaction between the sympathetic and parasympathetic nervous system prior to the onset of ventricular arrhythmias [31]. The effect on HR and HRV variables is more often heterogeneous and affected by individual patient characteristics, which make the use of advanced machine learning methods necessary to fully capture this variability.HRV as a useful tool to assess sympathetic and parasympathetic influences on disease states and is thus a useful application in assessing autonomic dysfunction in patients with AF [127], [128]. Recent study shows there is a modest correlation for a baseline HR < 60 beats/minute with an increased risk of AF. The study looked at data from a large database where subjects were followed up for an average of 20 years, allowing them to identify factors such as HRV and resting HR which might predict the chances of contracting AF years later [129]. One of the goals of this research is to combine different HRV measurements in time domain, frequency domain, bispectrum, and nonlinear analysis with the state-of-the-art machine learning algorithms for the prediction or detection of life threatening arrhythmias. As a step towards this, the next chapter in this thesis presents a detailed evaluation of a range of the features described in this chapter for this purpose. This includes a review of the relevant literature in the field, databases used in previous work, and discussion of relevant experimental results from the literature.

# Ventricular Arrhythmia Prediction using Heart Rate Variability Analysis

ICDs and other technology platforms have been widely deployed in the last four decades to reduce sudden cardiac death risk in patients with a history of life-threatening arrhythmia. By continuous monitoring of the heart rate, ICDs can use decision algorithms to distinguish normal cardiac sinus rhythm or SVT from abnormal cardiac rhythms like VT and VF and deliver appropriate therapy such as an electrical stimulus. Despite the success of ICDs, more research is still needed, particularly in decision-making algorithms. Because of low specificity in practical devices, patients with ICDs still receive inappropriate shocks which may lead to inadvertent mortality and reduction of quality of life. At the same time, higher sensitivity can lead to the use of newer tiered therapies.

This chapter presents a review of existing methods for ventricular arrhythmia prediction in ICDs based on HRV analysis, highlights current issues, and discusses some directions for future research. Specifically, the varied features which have been used for VT-VF prediction are reviewed and experimentally evaluated using a classification algorithm, and results compared with results from the literature. More than fifty different features to address heart rate changes before sudden cardiac death are considered and a general methodology for experimental evaluation is proposed, based on a variety of studies of ICD functionality. A comparative study on the prediction performance of these features, using a publicly-available database, is presented, and the results compared with those from the literature where possible. The work described in this chapter has been published in Parsi *et al.*, "Prediction of Sudden Cardiac Death in Implantable Cardioverter Defibrillators: A Review and Comparative Study of Heart Rate Variability Features," *IEEE Rev. Biomed. Eng.*, vol. 13, pp. 5–16, 2020 [77].

## 4.1     Overview

This section reviews relevant topics for HRV analysis for arrhythmia prediction and detection. It starts with a brief overview of the clinical usage of this technology, as this provides motivation for the functional requirements of lower-level signal processing algorithms; on the one hand, sensitivity is important but at the same time, specificity is also important, not least in order to avoid inappropriate therapy. Two common types of arrhythmias are VT and VF, both of which have the potential to cause sudden cardiac death. As presented in Chapter 2, VT and VF both show fast hart rate sometimes even more than 300 beats/minute with chaotic and asynchronous electrical activity in the ventricle. Without immediate and appropriate treatment such as electrical shock, VF can result in sudden cardiac death [58] and its distinctive characteristics motivate the use of machine learning to distinguish these rhythms from normal cardiac activity. ICDs are an indicated treatment for patients at high risk of sudden cardiac death due to VT or VF [4]. These battery-operated devices are implanted in the body and constantly monitor cardiac activity through thin leads lodged into the right ventricle. ICDs are typically indicated for patients with substantial daily risk of severe arrhythmia [7]. ICDs have been proven to reduce the mortality of patients with life-threatening ventricular arrhythmias [9] and have become the therapy of choice for patients at high risk of sudden cardiac death.

By processing intra-cardiac electrical activities between electrodes, called an electrogram (EGM), ICDs may deliver one of two treatments if either VT or VF is predicted: (1) direct current defibrillation (electrical shock) or (2) ATP, to restore normal heart rhythm and prevent sudden cardiac death occurring [9], [130], [131]. A critical issue for ICD design is the development of algorithms to predict VT and VF arrhythmias and distinguish VT and VF from the healthy cardiac activity. Both sensitivity and specificity are important: highly sensitive algorithms minimize the risk of untreated arrhythmias and hence sudden cardiac death, but high specificity is also important to avoid invasive interventions such as electrical shock where inappropriate or unnecessary [14].

Unnecessary and inappropriate interventions are associated with increased mortality, and more than 11.5% of 719 preventative individual ICD patients received one or more inappropriate electrical shocks based on MADIT II research in 2008, constituting 31.2% of all electrical shocks applied in therapy overall [16]. Nearly 80% of inappropriate electrical shock interventions are caused by the misclassification of SVT as a VT or VF [72]. SVT is an abnormally fast heart rhythm, however, SVT in itself does not require any immediate therapy [132].

Recent ICDs offer tiered therapy and include ATP to treat VT and prevent sudden cardiac death. ATP requires a longer duration of analysis and hence better VT prediction, but the use of ATP reduces the incidence of electrical shock, causes less patient discomfort and requires less energy [58]. Furthermore, by analysing longer cardiac signals and predicting VT-VF events before they occur, ATP can be considered the most appropriate treatment at this early stage. Should ATP be ineffective, more aggressive therapies, typically electrical shock, are administered immediately after unsuccessful ATP or if a high probability of VF is detected.

Historically, EGM signals were used to extract time or the R-R interval, which is the time between instantaneous heartbeats. The R-R interval can be converted to HRV, which is used for diagnosis. Advances in embedded device computation power, programmability, memory functions, device-related history features, therapy options and battery life have enabled EGM signal morphology to be used in ICDs [133]. SVT, VT, and VF usually activate the ventricle in a distinct manner, and as a result, different shapes are expected from ventricle electrogram morphology as well, which can be used to distinguish these cases. Algorithms based on differences in EGM morphology use either direct models of heartbeat (QRS complex) [134] or models using wavelet features [21], [22], covariance-based model [23], spatial projection of tachycardia [24] or an indirect/implicit model of heartbeat such as similarity measures which are based on information theory concepts [25]. In contrast, interval-based algorithms are less computationally intensive, utilizing information such as rate, onset, stability and entropy from the HRV signal [4], [26]–[28]. No consensus exists as to the optimal algorithms for onset of VT and VF detection, nor as to methods to decrease the inappropriate electrical shock interventions due to misclassification of SVT, VT and VF [14].

A number of aspects need to be considered when reviewing prediction algorithms used in ICDs: 1) feature extraction and classification methods, which are computationally complex; 2) accuracy (sensitivity and specificity); 3) window length required; and finally, 4) expected time remaining until sudden cardiac death, which affects the available therapeutic options for the ICD. Numerous machine learning methods have been proposed to address these concerns, however, in each case, different subsets of features derived from HRV have been used to predict sudden cardiac death. The focus in this thesis is on interval-based algorithms. Chapter 3 presented a detailed description of the individual features. Table 4.1 summarises previous research from the literature that makes use of these features in various combinations. This table includes the features used in each study, the classification method, as well as a summary of the main outcome from the study.

**Table 4.1.** Summary of previous studies conducted in prediction of sudden cardiac death using R-R interval signals from ICD devices.

| Previous Work | Used Features, Classification Methods and Outcomes |
| --- | --- |
| Mani *et al.*, 1999 [91] | **Features: time domain:** MeanNN, **frequency domain:** spectral power analysis in different frequency band from 0-0.1 to 0.9-1.0 Hz **Classification:** simple thresholding **Outcome:** best result in 0.8-0.9 Hz frequency band with time window = 100 seconds just before sudden cardiac death. |
| Lombardi *et al.*, 2000 [48] | **Features: time domain:** MeanNN, **frequency domain:** spectral power analysis in VLF, LF, HF, and total power frequency (TPF) band (0.0033-0.4 Hz), and LF/HF ratio **Outcomes:** decrease in MeanNN before VT onset and increase in LF/HF component before VT onset. |
| Meyerfeldt *et al.*, 2002 [47] | **Features: time domain:** MeanNN, SDNN, pNN50, RMSSD, **frequency domain:** spectral power analysis in VLF, LF, HF, TPF, and following ratios: VLF/TF, LH/HF, HF/P, $LF_n$ (LF power in normalised units: $Power_{LF}/(Power_{TP} - Power_{VLF}) \times 100$), **dynamic analysis:** Polvar10, Fitgra9 **Outcomes:** significant differences in the dynamic behaviour, significant MeanNN difference before onset of VT and the incidence of VT is higher during the daytime than at night. |
| Thong *et al.*, 2007 [135] | **Features: time domain:** HR and acceleration pattern **Classification:** detecting two patterns (each 50 R-R intervals) in 1.8 hour with heart rate exceed 86 bpm (700 ms) (rule-based decision) **Outcomes:** the basic acceleration pattern was found during sinus rhythm in 1.8-hour period prior to 83% of VT episodes. |
| Zhuang *et al.*, 2008 [4] | **Features: time domain:** MeanNN, SDNN, RMSSD, pNN50, **nonlinear analysis:** ApEn, SampEn **Classification:** simple thresholding **Outcomes:** obvious increase in HR of the patients, the values of two entropy measures are significantly smaller for VT-VF episodes than those for normal series. |
| Lerma *et al.*, 2008 [49] | **Features: time domain:** MeanNN, SDNN,, RMSSD **frequency domain:** spectral power analysis in LF, HF, and LF/HF, $LF_n$, and $LH_n$ ratio, **nonlinear analysis:** FWShannon, Forbidden Words, Polvar10, **along with other features:** PVCs (premature ventricular complexes in %), VTCI (VT coupling interval - CI for short - in millisecond), $RR_{n-1}$ (in millisecond), $CI/RR_{n-1}$, VT cycle length (in millisecond) **Outcome:** Polvar10 gave significant changes with respect to controlled data as early as 20 min prior to the VT onset. |

| Previous Work | Used Features, Classification Methods and Outcomes |
| --- | --- |
| Bilgin *et al.*, 2009 [26] | **Features: frequency domain:** spectral power analysis of sub-bands in LF base-bands and HF base-bands<br>**Classification:** multilayer perceptron (MLP) neural network<br>**Outcome:** most dominant frequency range in LF is [0.039-0.085] and in HF is [0.195-0.281], however, there is no result for general sudden cardiac death prediction. |
| Dong *et al.*, 2011 [27] | **Features: time domain:** MeanNN, SDNN, **frequency domain:** frequency marginal spectrum and Hilbert spectrum in LF, HF VHF, and LF/HF ratio<br>**Outcome:** the amplitude (energy) of Hilbert marginal spectrum of HRV signal preceding the onset of VT-VF event significantly increases. |
| Joo *et al.*, 2010-2012 [28], [82], [83] | **Features: time domain:** MeanNN, SDNN, RMSSD, pNN50, **frequency domain:** spectral power analysis in VLF, LF, HF, and LF/HF ratio, **nonlinear analysis:** $SD_1$, $SD_2$, $SD_1/SD_2$<br>**Classification:** artificial neural network<br>**Outcomes: in 2011:** using PCA as features selection method, **in 2012:** best prediction result for VT-VF events using artificial neural network, 10 seconds before sudden cardiac death. |
| Rozen *et al.*, 2013 [116] | Features: nonlinear analysis: Dyx parameter<br>**Classification:** Dyx value of less than 3 or greater than 6 in eight consecutive R-R intervals smaller than 600 milliseconds (rule-based decision)<br>**Outcomes:** by using multipole method, they showed that Dyx parameters dynamic became typically pathological between 800 and 5000 R-R intervals (10-60 minutes) before onset of ventricular arrhythmias and used it for sudden cardiac death prediction. |
| Wollmann *et al.*, 2015 [89] | **Features: time domain:** MeanNN, SDNN, RMSSD, **frequency domain:** spectral power analysis in LF, HF, and LF/HF ratio<br>**Classification:** regression tree analysis<br>**Outcomes:** improving the sensitivity using value of different feature as a node of the regression tree before onset of VT-VF, and early and gradual increase in HR 20 minutes before VT-VF. |
| Boon *et al.*, 2016 [84] | **Features: time domain:** MeanNN, SDNN, RMSSD, NN50, pNN50, HRVTri, **frequency domain:** spectral power analysis in LF, HF, and LF/HF ratio, **bispectrum analysis**, **nonlinear analysis:** SampEn, $SD_1$, $SD_2$, $SD_1/SD_2$<br>Classification: SVM<br>**Outcomes:** using genetic algorithm as a feature selection method to get the best result just before VT-VF events. |

## 4.2    Databases

In this subsection a number of ICD signal databases are reviewed to identify the most suitable test cases for this comparative study. A number of proprietary and publicly-available datasets have been used in the literature [4], [26], [89], [91], [116], [135], [27], [28], [47]–[49], [82]–[84], and are summarized in Table 4.2. Six of these datasets [47]–[49], [89], [116], [135] are private hospital records specifically for VT and VF prediction using mostly Medtronic ICDs (Medtronic 7220 to 7223 models), Biotronik® Phylax™ XM and Biotronik® Mycro Phylax™ single chamber ICDs, and there is little publicly-available information on these databases.

**Table 4.2.** Summary of databases of R-R interval signals from ICD devices, and their usage in the literature.

| Database (with ref) | Description |
| --- | --- |
| Medtronic version 1.0 used in: | |
| Mani *et al.*, 1999 [91]<br>Zhuang *et al.*, 2008 [4]<br>Bilgin *et al.*, 2009 [26]<br>Dong *et al.*, 2011 [27]<br>Joo *et al.*, 2010-2012 [28], [82], [83]<br>Pouyan *et al.*, 2013 [136]<br>Boon *et al.*, 2016 [84]<br>Parsi *et al.*, 2019-2021 [77], [78], [137] | R-R interval time series were obtained from 78 ICD patients (Medtronic Jewel Plus™ 7218); 126 controlled and 135 pre-VT-VF events.<br><br>(Publicly-available a on PhysioNet [79]) |
| 48 ICD patients<br>68 controlled and pre-VT events used in:<br>Lombardi *et al.*, 2000 [48] | The main dataset contains 60 patients implanted with Medtronic 7220 to 7223 devices (Minneapolis, Minnesota) from September 1, 1997 to December 1, 1998. Data were subsequently analysed during controlled period (normal rhythms), 20 minutes before VT and immediately before VT. (Private database) |
| 46 ICD patients<br>47 controlled and<br>67 pre-VT events, used in:<br>Meyerfeldt et al., 2002 [47] | The main dataset contains 63 ICD patients of Franz-Volhard-Hospital (Berlin, Germany) with Medtronic ICD models PCD 7221/7221, which are capable of storing at least 1024 beat-to-beat intervals prior to onset of VT. The study considered 131 VT and 74 controlled events and then removed 64 VT and 27 controlled events because of AF, permanent pacing, incessant VT, incomplete storage of episodes or storage artifacts. (Private database) |

| Database (with ref) | Description |
|---|---|
| 90 ICD patients 26 controlled and 208 pre-VT-VF events used in: Thong et al., 2007 [135] | A set of 208 events recorded from 90 patients in Biotronik® (Berlin, Germany) European HRV dataset which were obtained from European patients between 1997 to 2000. These R-R interval records stored by Biotronik® Phylax™ XM and Mycro Phylax™ single chamber ICDs. (Private database) |
| 13 ICD patients 42 controlled and 33 pre-VT events used in: Lerma et al., 2008 [49] | The data (68 R-R intervals) were collected from 1998 to 2003 from 13 patients at the Charité Hospital (Berlin, Germany). Each patient had at least one VT event and one controlled event obtained during a follow-up visit. (Private database) |
| 28 ICD patients 60 controlled and 64 pre-VT-VF events used in: Rozen *et al.*, 2013 [116] | Events were collected from subjects from HAWAI registry at 41 hospitals (phase I and II), who had a medical record and had experienced ventricular arrhythmia (verified and documented) during a 2-year follow-up period. (Private database) |
| 45 ICD patients 72 controlled and 68 pre-VT-VF events used in: Wollmann et al., 2015 [89] | Events were collected from subjects from HAWAI registry at 41 hospitals (phase I and II), who had a medical record and had experienced ventricular arrhythmia (verified and documented) during a 2-year follow-up period. (Private database) |

## 4.2.1    Medtronic Version 1.0

The experimental data used in this thesis for ventricular arrhythmia prediction come from a public database called "spontaneous ventricular tachyarrhythmia database (SVTD) - Medtronic version 1.0" from Medtronic, Inc. [79], available in PhysioNet signal archives [138]. This database was recorded from 78 subjects (63 males and 15 females, aged 20.7-75.3 years) in United States and Canada who were implanted between February 1995 and January 1997 by Medtronic Jewel Plus™ 7218 ICDs (Minneapolis, MN, USA).

Medtronic version 1.0 consists of the following records providing R-R interval data with 10 milliseconds resolution: 106 pre-VT records, 29 pre-VF records, and 126 controlled (CON) records (there were 135 controlled records, but 9 records were duplicated). Each sample of 135 spontaneous episodes of VT or VF included R-R intervals prior to the corresponding event. All patients experienced at least one electrical shock after VT-VF was detected by ICD. VT-VF records contain 1024 of most recently R-R interval measures (before ICD treatments), which the ICD has recorded when a tachyarrhythmia is detected by the device; these signals are recorded while patients were carrying ICDs in the course of their normal daily activities outside the hospital. The CON records were obtained from the patients at the time of interrogation of the ICD during scheduled check-ups and were the most recent R-R intervals of normal rhythm available for recording. This database has been generated by a market leader in the production of ICDs and has been used in many studies. Therefore, it is a useful choice to use for this research as it permits comparison of results obtained here with results from the literature. Nonetheless, it is noted that the amount of control data is necessarily limited, and this may have some impact on the generalizability of the results. Patient metadata including gender, age, cardiac history, and arrhythmia history are available as shown in Table 4.3.

**Table 4.3.** Patient metadata in Medtronic version 1.0 dataset.

| Dataset Characteristic | Total |
| --- | --- |
| Number of patients | 78 |
| Male/Female | 63/15 |
| Age | 20.7-75.3 |
| History of VT | 62 |
| History of heart failure | 31 |
| Antiarrhythmic drugs | 54 |
| Number of arrhythmia records | 135 |
| Number of single VT records | 58 |
| Number of single VF records | 25 |
| Number of Multi VT records | 48 |
| Number of Multi VF records | 4 |
| Number of CON records | 126 |

VT = ventricular tachycardia; VF = ventricular fibrillation; CON = controlled data

## 4.3  General Ventricular Arrhythmia Prediction Model

Many researchers have developed algorithms for prediction of ventricular arrhythmia before onset using HRV signals from different models of ICDs on the market. HRV analysis has been performed on the R-R intervals from 1 minute to 108 minutes either to improve prediction accuracy or to simplify the computational complexity of the algorithm for practical implementations on ICDs. Note that while many researchers have carried out detailed analysis of features, only a subset of these works have actually implemented computer-aided prediction systems. A summary of some previous work in this area is presented in Table 4.1, including works that have specifically used the Medtronic version 1.0 dataset, and have implemented automated prediction systems. A general methodology for ventricular arrhythmia prediction (used for experimental work in this thesis) is explained in this section including statistical analysis and performance metrics.

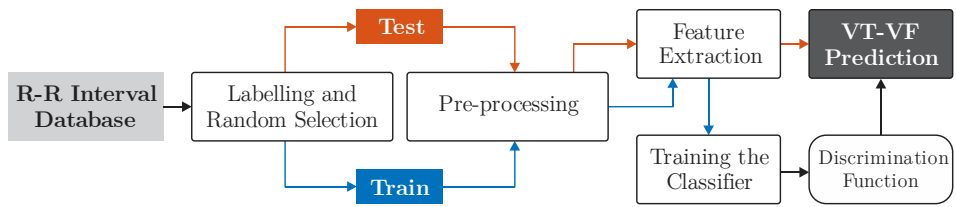

**Figure 4.1.** Overview of arrhythmia prediction methods based on cardiac signal activities.

Most studies have used a similar approach to that shown in Figure 4.1. Firstly, the database is labelled and divided into training and test datasets. In many cases, train and test cases are chosen randomly. This may mean that cardiac rhythms recorded from a single patient may appear in both the train and testing sets. In practice, a better approach is to have no overlap between training and test sets.

Pre-processing (e.g., noise reduction or resampling) is typically applied to obtain signals that are more suitable for feature extraction. Systems may also include feature selection or statistical analysis either independently of, or as part of the training phase, to get the best subset of features. Many algorithms have been proposed for sudden cardiac death prediction, including artificial neural networks [26], [28], [82], [83], [139], [140], regression tree [89], and SVM [84], [86], [87], [92], [121], [136], [141]–[143]. Here, we present prediction results from a comparison of some of the feature categories in time domain, frequency domain, bispectrum and nonlinear analysis described in Chapter 3 using the Medtronic version 1.0 dataset as a common dataset for the comparison. The system of Figure 4.1 is used as a basis for the evaluation. Many of the system components in Figure 4.1 are well known and have been extensively reviewed in previous work; however, some of the blocks used in this research that are less commonly used in previous work will be described in more detail.

### 4.3.1 Preprocessing and Noise Reduction

Data from the Medtronic version 1.0 dataset are interpolated by using cubic spline interpolation as used in [26]–[28], [83], [84], [88], [92], [94], [144], [145] and resampled at 16 Hz and converted to IHR signals based on formula (3.1). Afterwards, ectopic beats are removed using a median filter [136]. Median filter as a nonlinear filtering technique is widely used in digital signal and image processing. Given a sequence in ascending order as a $x_1, x_2, \dots, x_n$ the $n^{th}$ order median value ($m$) is defined as follows [146]:

$$m = \begin{cases} \frac{1}{2}\left(x_{n/2} + x_{n/2+1}\right), & \text{if } n \text{ is even} \\ x_{(n+1)/2}, & \text{if } n \text{ is odd} \end{cases} \tag{4.1}$$

### 4.3.2 Feature Extraction

In this chapter fifty HRV features in four different categories are considered: time domain, frequency domain, bispectrum and nonlinear analysis. The features have already been described in Chapter 3 and summarised in their individual categories in tables in that chapter; for completeness, the full set is listed again in Table 4.4.

**Table 4.4.** Summary of features considered in time domain (6 features), frequency domain (4 features), bispectrum analysis (31 features), and nonlinear analysis (9 features) [77], [78].

| Features Category (Total number of features) | Features Name |
|---|---|
| Time domain (6 features) | MeanNN, SDNN, RMSSD, NN50, pNN50, and HRVTri |
| Frequency domain (4 features) | PSD in VLF, LF, HF and LF/HF ratio |
| Bispectrum analysis (31 features) | $M_{avg}$ in LL, LH, HH, and ROI |
| | $P_{avg}$ in LL, LH, HH, and ROI |
| | $E_{nb}$ in LL, LH, HH, and ROI |
| | $E_{snb}$ in LL, LH, HH, and ROI |
| | $L_m$ in LL, LH, HH, and ROI |
| | $L_{dm}$ in LL, HH, and ROI |
| | $WCOB_i$ in LL, LH, HH, and ROI |
| | $WCOB_j$ in LL, LH, HH, and ROI |
| Nonlinear and dynamic analysis (9 features) | $SD_1$, $SD_2$, and $SD_1/SD_2$ ratio |
| | Sample entropy, Rényi entropy, and Tsallis entropy |
| | Hjorth's parameters (activity, mobility, and complexity) |

## 4.3.3 Statistical Feature Analysis

Student's *t*-test method is one of the most common methods for feature ranking applied to analyse features extracted from HRV analysis in [28], [47], [86], [89], [142], [143], [147]. This test can be used to determine whether the mean of two datasets are different or not [148]. The test gives the p-value and t-values for the features extracted for two groups of data. Statistically, a low p-value is preferred ($p < 0.05$) [149], and the higher the t-value, the better the ranking. The Student's *t*-test in this experiment was implemented using p-value = 0.05 where both tails were included. In Table 4.5 and Table 4.6, comparisons of different features derived from Medtronic version 1.0, and ranked using Student's *t*-test, are presented. Features are divided into their four main categories and statistical values have been calculated for one and five minutes before VT-VF and also for CON signals (Table 4.5 and Table 4.6). Although in all categories we see better p-values for 1-minute R-R interval lengths, statistical analysis cannot readily be used as a main feature selector for VT-VF prediction as the analysis duration is too short. Moving from 5-minute to 1-minute R-R intervals shows these features are statistically more discriminative as we get closer to a sudden cardiac death event. In many studies the time and frequency domain categories have been used over 5-minute windows to achieve better prediction rate [28], [45], [121]. In the other two categories, bispectrum analysis and nonlinear analysis, p-values are more discriminative with a 1-minute window, which suggests that prediction of VT-VF is better using shorter signal length.

**Table 4.5.** Mean, standard deviation (SD), and p-value of different feature categories using 1-minute windows of IHR signals from Medtronic version 1.0 dateset.

| Feature Category | VT-VF Dataset | | | CON Dataset | | | p-Value |
|---|---|---|---|---|---|---|---|
| | Mean | ± | SD | Mean | ± | SD | |
| Time domain | 34.3 | ± | 9.3 | 25.7 | ± | 4.9 | 0.0001 |
| Frequency domain | 1.8816 e+4 | ± | 9.3984 e+3 | 1.1314 e+4 | ± | 4.1301 e+3 | 0.0001 |
| Bispectrum analysis | 1.4872 e+13 | ± | 7.0555 e+13 | 2.2681 e+09 | ± | 10.1264 e+09 | 0.0021 |
| Nonlinear analysis | 1137.2 | ± | 545.3 | 682.3 | ± | 239.5 | 0.0258 |

**Table 4.6.** Mean, standard deviation (SD), and p-value of different feature categories using 5-minute windows of IHR signals from Medtronic version 1.0 dateset.

| Feature Category | VT-VF Dataset | | | CON Dataset | | | p-Value |
|---|---|---|---|---|---|---|---|
| | Mean | ± | SD | Mean | ± | SD | |
| Time domain | 31.9 | ± | 7.9 | 26.7 | ± | 5.1 | 0.0004 |
| Frequency domain | 1.6788 e+4 | ± | 6.8189 e+3 | 1.1968 e+4 | ± | 4.2106 e+3 | 0.0004 |
| Bispectrum analysis | 1.7853 e+13 | ± | 8.6512 e+13 | 1.2783 e+11 | ± | 5.4113 e+11 | 0.0427 |
| Nonlinear analysis | 1098.6 | ± | 415.3 | 839.9 | ± | 261.7 | 0.1154 |

### 4.3.4 Signal Classification and Performance Metrics

Results from different studies for detection of onset of VT-VF have mainly been presented in terms of sensitivity (true positive rate), specificity (true negative rate), accuracy and area under the Receiver Operating Curve (AUC). Some of these measures are defined as follows:

$$Sensitivity = \frac{TP}{TP + FN} \tag{4.2}$$

$$Specificity = \frac{TN}{TN + FP} \tag{4.3}$$

$$Accuracy = \frac{TP + TN}{TP + FP + TN + FN} \tag{4.4}$$

where TP, TN, FP and FN stand for true positive, true negative, false positive and false negative respectively.

For AUC calculations, the receiver operator characteristic (ROC) curves are first computed. For the results presented here, an SVM classifier was used for detection, with features extracted from signals of length 1 minute and 5 minutes. For each series of experiments, as well as a decision on detection (or not) the classifier can also output the confidence level ($C$) for each test, reflected in the output value of the classifier before binary decision, which lies in the range ($-1 \leq C \leq 1$). The AUC measures the two-dimensional area underneath the ROC curve to represent how well predictions are ranked, rather than their absolute values. With the value between 0 and 1, it also measures the quality of the model's predictions irrespective of what classification threshold is chosen [150].

## 4.4 Experimental Results

In the comparison described here, we measured acute changes in the HRV features extracted from signals beginning 1 or 5 minutes before a point 10 seconds prior to the onset of ventricular arrhythmias recorded from patients with an ICD, so with correct detection, the device still has time to take effective treatment action. The features have been used according to the categories in Table 4.4, i.e., time domain, frequency domain, bispectrum, and nonlinear analysis. The SVM has been implemented in MATLAB with Gaussian function as a kernel. After applying Bayesian optimization algorithm[1] on SVM in the training phase, the kernel scale has been set to 5 for 1-minute HRV analysis and 10 for 5-minute HRV analysis with the box constraint equal to 2 for both signal lengths. Broadly speaking, there is a trade-off between the signal length and performance. Longer analysis windows include more normal data which may lead to better specificities so the ICD device can detect normal (CON) events more accurately, though not with all feature categories.

Generally-speaking, adding additional categories of feature gives a performance improvement, especially in specificity, when calculated using the longer analysis window. On the whole, frequency domain features display the lowest performance when used on their own, while time domain features perform best when used on their own. A number of observations have been made which provide insight into feature selection and new feature development.

### 4.4.1    Time Domain Results

Table 4.7 and Table 4.8 show mean, standard deviation, sensitivity, and specificity of time domain features where 1 and 5-minute windows of R-R intervals have been processed. Considering time domain features individually, all achieve comparatively high specificity but at the cost of very poor sensitivity, with the exception of HRVTri. This suggests that future time domain feature development should focus on achieving higher sensitivity. In Table 4.9 the performance metrics of SVM have been presented with both 1 and 5-minute windows. The results indicate that time domain features work better with a longer analysis window.

---

[1] The Bayesian optimization algorithm attempts to minimize a scalar objective function in a bounded domain. The function can be deterministic or stochastic, meaning it can return different results when evaluated at the same input. The components of input can be continuous reals, integers, or categorical. It is usually employed to optimize expensive-to-evaluate functions [237].

**Table 4.7.** Sensitivity (SN), specificity (SP), mean, and standard deviation (SD) of time domain features using 1-minute windows of R-R interval signals from Medtronic version 1.0 dateset.

| Feature Name | SN (%) | SP (%) | VT-VF Dataset | | | CON Dataset | | |
|---|---|---|---|---|---|---|---|---|
| | | | Mean | ± | SD | Mean | ± | SD |
| MeanNN | 61.7 | 84.1 | 9.52 E+01 | ± | 2.28 E+01 | 7.39 E+01 | ± | 1.33 E+01 |
| SDNN | 64.5 | 97.2 | 1.20 E+01 | ± | 7.33 E+00 | 2.94 E+00 | ± | 2.01 E+00 |
| RMSSD | 61.7 | 86.9 | 9.62 E+01 | ± | 2.29 E+01 | 7.40 E+01 | ± | 1.33 E+01 |
| NN50 | 11.2 | 100.0 | 5.98 E-01 | ± | 1.60 E+00 | 0.00 E+00 | ± | 0.00 E+00 |
| pNN50 | 14.0 | 100.0 | 4.84 E-04 | ± | 1.25 E-03 | 0.00 E+00 | ± | 0.00 E+00 |
| HRVTri | 82.2 | 79.4 | 1.94 E+00 | ± | 1.21 E+00 | 3.87 E+00 | ± | 1.21 E+00 |

**Table 4.8.** Sensitivity (SN), specificity (SP), mean, and standard deviation (SD) of time domain features using 5-minute windows of R-R interval signals from Medtronic version 1.0 dateset.

| Feature Name | SN (%) | SP (%) | VT-VF Dataset | | | CON Dataset | | |
|---|---|---|---|---|---|---|---|---|
| | | | Mean | ± | SD | Mean | ± | SD |
| MeanNN | 54.2 | 80.4 | 8.95 E+01 | ± | 1.85 E+01 | 7.59 E+01 | ± | 1.32 E+01 |
| SDNN | 43.9 | 90.7 | 9.45 E+00 | ± | 6.18 E+00 | 5.03 E+00 | ± | 2.86 E+00 |
| RMSSD | 54.2 | 80.4 | 9.02 E+01 | ± | 1.87 E+01 | 7.61 E+01 | ± | 1.32 E+01 |
| NN50 | 8.4 | 100 | 9.53 E-01 | ± | 2.83 E+00 | 0.00 E+00 | ± | 0.00 E+00 |
| pNN50 | 5.6 | 100 | 1.67 E-04 | ± | 5.40 E-04 | 0.00 E+00 | ± | 0.00 E+00 |
| HRVTri | 84.1 | 70.1 | 1.89 E+00 | ± | 9.57 E-01 | 3.38 E+00 | ± | 1.18 E+00 |

**Table 4.9.** Sensitivity (SN), specificity (SP), accuracy (ACC), and AUC of time domain features together using 1- and 5-minute windows of R-R interval signals from Medtronic version 1.0 dateset.

| R-R interval Length | SN (%) | SP (%) | ACC (%) | AUC |
|---|---|---|---|---|
| **1-minute** | 78.6 | 78.6 | 78.6 | **0.88** |
| **5-minute** | **85.7** | **82.1** | **83.9** | 0.83 |

### 4.4.2 Frequency Domain Result

Table 4.10 and Table 4.11 show mean, standard deviation, sensitivity, and specificity of frequency domain features while 1 and 5-minute R-R intervals have been processed. Similar to time domain analysis, higher specificity than sensitivity is observed for frequency domain features. None of the four frequency domain features evaluated achieve both sensitivity and specificity higher than 50%. In Table 4.12 the performance metrics of are presented for 1 and 5-minute windows. Based on the presented results, frequency domain features display better performance with a shorter analysis window.

**Table 4.10.** Sensitivity (SN), specificity (SP), mean, and standard deviation (SD) of frequency domain features using 1-minute windows of R-R interval signals from Medtronic version 1.0 dateset.

| Feature Name | SN (%) | SP (%) | VT-VF Dataset | | | CON Dataset | | |
|---|---|---|---|---|---|---|---|---|
| | | | Mean | ± | SD | Mean | ± | SD |
| PSD$_{VLF}$ | 47.7 | 90.7 | 5.16 E+04 | ± | 2.57 E+04 | 3.10 E+04 | ± | 1.13 E+04 |
| PSD$_{LF}$ | 45.8 | 90.7 | 2.31 E+04 | ± | 1.16 E+04 | 1.39 E+04 | ± | 5.08 E+03 |
| PSD$_{HF}$ | 47.7 | 94.4 | 4.94 E+02 | ± | 2.84 E+02 | 2.77 E+02 | ± | 1.02 E+02 |
| PSD$_{LF/HF}$ | 15.0 | 95.3 | 4.83 E+01 | ± | 5.07 E+00 | 5.03 E+01 | ± | 2.27 E+00 |

**Table 4.11.** Sensitivity (SN), specificity (SP), mean, and standard deviation (SD) of frequency domain features using 5-minute windows of R-R interval signals from Medtronic version 1.0 dateset.

| Feature Name | SN (%) | SP (%) | VT-VF Dataset | | | CON Dataset | | |
|---|---|---|---|---|---|---|---|---|
| | | | Mean | ± | SD | Mean | ± | SD |
| PSD$_{VLF}$ | 49.5 | 86.0 | 4.60 E+04 | ± | 1.87 E+04 | 3.28 E+04 | ± | 1.16 E+04 |
| PSD$_{LF}$ | 49.5 | 86.0 | 2.06 E+04 | ± | 8.38 E+03 | 1.47 E+04 | ± | 5.18 E+03 |
| PSD$_{HF}$ | 46.7 | 86.0 | 4.37 E+02 | ± | 2.22 E+02 | 2.94 E+02 | ± | 1.04 E+02 |
| PSD$_{LF/HF}$ | 12.2 | 95.3 | 4.87 E+01 | ± | 4.36 E+00 | 5.02 E+01 | ± | 1.81 E+00 |

**Table 4.12.** Sensitivity (SN), specificity (SP), accuracy (ACC), and AUC of frequency domain features together using 1- and 5-minute windows of R-R interval signals from Medtronic version 1.0 dateset.

| R-R interval Length | SN (%) | SP (%) | ACC (%) | AUC |
|---|---|---|---|---|
| 1-minute | **57.1** | **67.9** | **62.5** | **0.66** |
| 5-minute | 53.6 | 64.3 | 58.9 | 0.64 |

### 4.4.3    Bispectrum Analysis Results

Table 4.13 and Table 4.14 shows mean, standard deviation, sensitivity, and specificity of bispectrum features while 1 and 5-minute R-R intervals have been processed. Higher variance is observed in the bispectral features. The first two types achieve high specificity but at the cost of effectively 0% sensitivity. In contrast, the latter six types are more balanced, achieving comparable levels of both sensitivity and specificity, suggesting that these features may be suitable. In Table 4.15 he performance metrics of SVM are presented while 1 and 5-minute R-R intervals have been used.

**Table 4.13.** Sensitivity (SN), specificity (SP), mean, and standard deviation (SD) of bispectrum analysis features using 1-minute windows of R-R interval signals from Medtronic version 1.0 dateset.

| Feature Name | SN (%) | SP (%) | VT-VF Dataset | | | CON Dataset | | |
|---|---|---|---|---|---|---|---|---|
| | | | Mean | ± | SD | Mean | ± | SD |
| $M_{avg}\text{LL}$ | 13.1 | 100.0 | 6.95 E+06 | ± | 1.37 E+07 | 6.45 E+04 | ± | 1.51 E+05 |
| $M_{avg}\text{LH}$ | 30.8 | 100.0 | 2.34 E+06 | ± | 3.14 E+06 | 1.50 E+04 | ± | 4.09 E+04 |
| $M_{avg}\text{HH}$ | 25.2 | 100.0 | 1.11 E+06 | ± | 1.60 E+06 | 5.69 E+03 | ± | 1.51 E+04 |
| $M_{avg}\text{ROI}$ | 28.0 | 100.0 | 2.49 E+06 | ± | 3.73 E+06 | 1.83 E+04 | ± | 4.20 E+04 |
| $P_{avg}\text{LL}$ | 1.9 | 100.0 | 3.62 E+14 | ± | 1.79 E+15 | 5.51 E+10 | ± | 2.48 E+11 |
| $P_{avg}\text{LH}$ | 10.3 | 100.0 | 2.09 E+13 | ± | 4.89 E+13 | 3.43 E+09 | ± | 1.59 E+10 |
| $P_{avg}\text{HH}$ | 12.2 | 100.0 | 5.18 E+12 | ± | 1.24 E+13 | 7.39 E+08 | ± | 3.47 E+09 |
| $P_{avg}\text{ROI}$ | 1.9 | 100.0 | 7.33 E+13 | ± | 3.34 E+14 | 1.10 E+10 | ± | 4.68 E+10 |
| $E_{nb}\text{LL}$ | 92.5 | 43.9 | 2.18 E+00 | ± | 1.95 E-01 | 1.85 E+00 | ± | 3.73 E-01 |
| $E_{nb}\text{LH}$ | 83.2 | 66.4 | 3.15 E+00 | ± | 2.00 E-01 | 2.79 E+00 | ± | 2.88 E-01 |
| $E_{nb}\text{HH}$ | 70.1 | 64.5 | 3.07 E+00 | ± | 2.47 E-01 | 2.79 E+00 | ± | 2.62 E-01 |
| $E_{nb}\text{ROI}$ | 84.1 | 57.9 | 3.54 E+00 | ± | 5.23 E-01 | 2.67 E+00 | ± | 7.78 E-01 |
| $E_{snb}\text{LL}$ | 84.1 | 54.2 | 1.97 E+00 | ± | 4.30 E-01 | 1.36 E+00 | ± | 6.25 E-01 |
| $E_{snb}\text{LH}$ | 75.7 | 72.9 | 2.82 E+00 | ± | 5.01 E-01 | 2.09 E+00 | ± | 5.72 E-01 |
| $E_{snb}\text{HH}$ | 69.2 | 72.9 | 2.63 E+00 | ± | 5.96 E-01 | 1.93 E+00 | ± | 5.26 E-01 |
| $E_{snb}\text{ROI}$ | 75.7 | 62.6 | 3.01 E+00 | ± | 8.83 E-01 | 1.77 E+00 | ± | 9.98 E-01 |
| $L_{m}\text{LL}$ | 76.6 | 92.5 | 1.36 E+02 | ± | 2.75 E+01 | 8.48 E+01 | ± | 2.41 E+01 |
| $L_{m}\text{LH}$ | 84.1 | 91.6 | 3.57 E+02 | ± | 7.68 E+01 | 1.79 E+02 | ± | 7.69 E+01 |
| $L_{m}\text{HH}$ | 86.9 | 88.8 | 3.33 E+02 | ± | 7.88 E+01 | 1.44 E+02 | ± | 8.10 E+01 |
| $L_{m}\text{ROI}$ | 86.0 | 91.6 | 6.85 E+02 | ± | 1.50 E+02 | 3.35 E+02 | ± | 1.49 E+02 |
| $L_{dm}\text{LL}$ | 75.7 | 91.6 | 5.45 E+01 | ± | 1.09 E+01 | 3.43 E+01 | ± | 9.63 E+00 |
| $L_{dm}\text{HH}$ | 86.0 | 86.9 | 8.36 E+01 | ± | 1.94 E+01 | 3.67 E+01 | ± | 2.05 E+01 |
| $L_{dm}\text{ROI}$ | 85.1 | 91.6 | 1.25 E+02 | ± | 2.68 E+01 | 6.38 E+01 | ± | 2.66 E+01 |
| $WCOB_{i}\text{LL}$ | 75.7 | 57.9 | 7.37 E-02 | ± | 1.15 E-02 | 6.36 E-02 | ± | 1.61 E-02 |
| $WCOB_{i}\text{LH}$ | 79.4 | 56.1 | 9.24 E-02 | ± | 1.05 E-02 | 8.30 E-02 | ± | 1.43 E-02 |
| $WCOB_{i}\text{HH}$ | 59.8 | 66.4 | 2.12 E-01 | ± | 1.59 E-02 | 2.03 E-01 | ± | 1.47 E-02 |
| $WCOB_{i}\text{ROI}$ | 69.2 | 74.8 | 1.11 E-01 | ± | 2.99 E-02 | 7.72 E-02 | ± | 2.74 E-02 |
| $WCOB_{j}\text{LL}$ | 82.2 | 52.3 | 1.11 E-01 | ± | 1.53 E-02 | 9.33 E-02 | ± | 2.48 E-02 |
| $WCOB_{j}\text{LH}$ | 75.7 | 77.6 | 2.51 E-01 | ± | 2.14 E-02 | 2.20 E-01 | ± | 1.99 E-02 |
| $WCOB_{j}\text{HH}$ | 62.6 | 66.4 | 2.82 E-01 | ± | 2.42 E-02 | 2.57 E-01 | ± | 2.11 E-02 |
| $WCOB_{j}\text{ROI}$ | 76.6 | 67.3 | 2.09 E-01 | ± | 5.24 E-02 | 1.39 E-01 | ± | 5.50 E-02 |

**Table 4.14.** Sensitivity (SN), specificity (SP), mean, and standard deviation (SD) of bispectrum analysis features using 5-minute windows of R-R interval signals from Medtronic version 1.0 dateset.

| Feature Name | SN (%) | SP (%) | VT-VF Dataset | | | CON Dataset | | |
|---|---|---|---|---|---|---|---|---|
| | | | Mean | ± | SD | Mean | ± | SD |
| $M_{avg}\text{LL}$ | 9.4 | 100.0 | 4.63 E+06 | ± | 1.11 E+07 | 4.07 E+05 | ± | 7.89 E+05 |
| $M_{avg}\text{LH}$ | 12.2 | 100.0 | 8.00 E+05 | ± | 1.81 E+06 | 2.77 E+04 | ± | 6.32 E+04 |
| $M_{avg}\text{HH}$ | 12.2 | 100.0 | 3.28 E+05 | ± | 7.66 E+05 | 7.88 E+03 | ± | 1.66 E+04 |
| $M_{avg}\text{ROI}$ | 12.2 | 100.0 | 1.24 E+06 | ± | 2.61 E+06 | 8.59 E+04 | ± | 1.59 E+05 |
| $P_{avg}\text{LL}$ | 1.9 | 100.0 | 4.61 E+14 | ± | 2.22 E+15 | 3.34 E+12 | ± | 1.42 E+13 |
| $P_{avg}\text{LH}$ | 0.9 | 100.0 | 5.96 E+12 | ± | 4.34 E+13 | 8.27 E+09 | ± | 3.10 E+10 |
| $P_{avg}\text{HH}$ | 0.9 | 100.0 | 1.21 E+12 | ± | 9.95 E+12 | 7.69 E+08 | ± | 2.86 E+09 |
| $P_{avg}\text{ROI}$ | 1.9 | 100.0 | 8.56 E+13 | ± | 4.07 E+14 | 6.10 E+11 | ± | 2.58 E+12 |
| $E_{nb}\text{LL}$ | 77.6 | 64.5 | 1.93 E+00 | ± | 4.30 E-01 | 1.46 E+00 | ± | 4.80 E-01 |
| $E_{nb}\text{LH}$ | 76.6 | 69.2 | 3.09 E+00 | ± | 2.35 E-01 | 2.78 E+00 | ± | 2.48 E-01 |
| $E_{nb}\text{HH}$ | 72.9 | 75.7 | 3.04 E+00 | ± | 2.59 E-01 | 2.65 E+00 | ± | 2.40 E-01 |
| $E_{nb}\text{ROI}$ | 79.4 | 61.7 | 3.14 E+00 | ± | 8.46 E-01 | 2.11 E+00 | ± | 8.94 E-01 |
| $E_{snb}\text{LL}$ | 63.6 | 79.4 | 1.51 E+00 | ± | 7.15 E-01 | 7.87 E-01 | ± | 6.13 E-01 |
| $E_{snb}\text{LH}$ | 69.2 | 72.9 | 2.70 E+00 | ± | 5.29 E-01 | 2.17 E+00 | ± | 4.56 E-01 |
| $E_{snb}\text{HH}$ | 57.9 | 94.4 | 2.58 E+00 | ± | 5.45 E-01 | 1.92 E+00 | ± | 3.26 E-01 |
| $E_{snb}\text{ROI}$ | 59.8 | 78.5 | 2.26 E+00 | ± | 1.22 E+00 | 1.00 E+00 | ± | 8.81 E-01 |
| $L_{m}\text{LL}$ | 76.6 | 72.0 | 1.31 E+02 | ± | 2.12 E+01 | 9.97 E+01 | ± | 2.26 E+01 |
| $L_{m}\text{LH}$ | 82.2 | 79.4 | 3.34 E+02 | ± | 6.36 E+01 | 2.12 E+02 | ± | 7.12 E+01 |
| $L_{m}\text{HH}$ | 81.3 | 91.6 | 3.06 E+02 | ± | 6.49 E+01 | 1.78 E+02 | ± | 6.72 E+01 |
| $L_{m}\text{ROI}$ | 81.3 | 89.7 | 6.40 E+02 | ± | 1.22 E+02 | 4.05 E+02 | ± | 1.31 E+02 |
| $L_{dm}\text{LL}$ | 77.6 | 67.3 | 5.30 E+01 | ± | 8.19 E+00 | 4.12 E+01 | ± | 8.97 E+00 |
| $L_{dm}\text{HH}$ | 81.3 | 90.7 | 7.69 E+01 | ± | 1.59 E+01 | 4.59 E+01 | ± | 1.68 E+01 |
| $L_{dm}\text{ROI}$ | 77.6 | 88.8 | 1.18 E+02 | ± | 2.10 E+01 | 7.89 E+01 | ± | 2.29 E+01 |
| $WCOB_{i}\text{LL}$ | 54.2 | 75.7 | 6.47 E-02 | ± | 1.42 E-02 | 5.35 E-02 | ± | 1.18 E-02 |
| $WCOB_{i}\text{LH}$ | 70.1 | 63.6 | 8.71 E-02 | ± | 1.24 E-02 | 7.96 E-02 | ± | 1.26 E-02 |
| $WCOB_{i}\text{HH}$ | 52.3 | 56.1 | 2.13 E-01 | ± | 1.66 E-02 | 2.13 E-01 | ± | 2.37 E-02 |
| $WCOB_{i}\text{ROI}$ | 56.1 | 86.9 | 9.41 E-02 | ± | 3.17 E-02 | 6.32 E-02 | ± | 1.94 E-02 |
| $WCOB_{j}\text{LL}$ | 66.4 | 74.8 | 9.57 E-02 | ± | 2.30 E-02 | 7.30 E-02 | ± | 2.07 E-02 |
| $WCOB_{j}\text{LH}$ | 63.6 | 75.7 | 2.48 E-01 | ± | 2.08 E-02 | 2.28 E-01 | ± | 1.71 E-02 |
| $WCOB_{j}\text{HH}$ | 62.6 | 59.8 | 2.81 E-01 | ± | 2.31 E-02 | 2.62 E-01 | ± | 2.24 E-02 |
| $WCOB_{j}\text{ROI}$ | 70.1 | 74.8 | 1.77 E-01 | ± | 6.23 E-02 | 1.08 E-01 | ± | 4.84 E-02 |

**Table 4.15.** Sensitivity (SN), specificity (SP), accuracy (ACC), and AUC of bispectrum analysis features together using 1- and 5-minute windows of R-R interval signals from Medtronic version 1.0 dateset.

| R-R interval Length | SN (%) | SP (%) | ACC (%) | AUC |
|---|---|---|---|---|
| 1-minute | 78.6 | **71.4** | **75** | **0.80** |
| 5-minute | **82.1** | 67.9 | **75** | 0.77 |

### 4.4.4 Nonlinear and Dynamic Analysis Results

Table 4.16 and Table 4.17 show mean, standard deviation, sensitivity, and specificity of nonlinear and dynamic analysis features while 1 and 5-minute R-R intervals have been processed. Some nonlinear features also show balanced performance, for example, both entropy-based features. These results suggest there is benefit in further feature development to improve both the sensitivity and specificity of prediction. In the Table 4.18 the performance metrics of SVM are presented while 1 and 5-minute R-R intervals have been used.

**Table 4.16.** Sensitivity (SN), specificity (SP), mean, and standard deviation (SD) of nonlinear and dynamic analysis features using 1-minute windows of R-R interval signals from Medtronic version 1.0 dateset.

| Feature Name | SN (%) | SP (%) | VT-VF Dataset | | | CON Dataset | | |
|---|---|---|---|---|---|---|---|---|
| | | | Mean | ± | SD | Mean | ± | SD |
| $SD_1$ | 70.1 | 80.4 | 2.55 E+00 | ± | 1.32 E+00 | 1.15 E+00 | ± | 1.05 E+00 |
| $SD_2$ | 44.9 | 81.3 | 7.73 E+01 | ± | 4.98 E+01 | 4.78 E+01 | ± | 3.33 E+01 |
| $SD_1/SD_2$ | 57.0 | 72.9 | 3.65 E-02 | ± | 1.49 E-02 | 2.46 E-02 | ± | 1.47 E-02 |
| SampEn | 83.2 | 18.7 | 6.86 E-02 | ± | 7.99 E-02 | 7.97 E-02 | ± | 5.11 E-02 |
| Rényi Entropy | 68.2 | 79.4 | 6.93 E+00 | ± | 2.38 E-01 | 7.18 E+00 | ± | 1.77 E-01 |
| Tsallis Entropy | 72.9 | 72.9 | 3.24 E+02 | ± | 6.29 E+01 | 3.92 E+02 | ± | 5.50 E+01 |
| Activity | 54.2 | 92.5 | 9.78 E+03 | ± | 4.79 E+03 | 5.64 E+03 | ± | 2.06 E+03 |
| Mobility | 53.3 | 88.8 | 3.39 E-02 | ± | 5.47 E-03 | 2.87 E-02 | ± | 3.33 E-03 |
| Complexity | 70.1 | 84.1 | 4.00 E+01 | ± | 7.43 E+00 | 4.96 E+01 | ± | 5.33 E+00 |

**Table 4.17.** Sensitivity (SN), specificity (SP), mean, and standard deviation (SD) of nonlinear and dynamic analysis features using 5-minute windows of R-R interval signals from Medtronic version 1.0 dateset.

| Feature Name | SN (%) | SP (%) | VT-VF Dataset | | | CON Dataset | | |
|---|---|---|---|---|---|---|---|---|
| | | | Mean | ± | SD | Mean | ± | SD |
| $SD_1$ | 29.9 | 83.2 | 1.89 E+00 | ± | 1.20 E+00 | 1.22 E+00 | ± | 8.99 E-01 |
| $SD_2$ | 4.7 | 97.2 | 7.99 E+01 | ± | 5.41 E+01 | 7.58 E+01 | ± | 4.70 E+01 |
| $SD_1/SD_2$ | 46.7 | 74.8 | 2.75 E-02 | ± | 1.47 E-02 | 1.86 E-02 | ± | 1.23 E-02 |
| SampEn | 5.6 | 100.0 | 6.49 E-02 | ± | 7.05 E-02 | 5.92 E-02 | ± | 4.72 E-02 |
| Rényi Entropy | 62.6 | 69.2 | 8.61 E+00 | ± | 2.12 E-01 | 8.76 E+00 | ± | 1.71 E-01 |
| Tsallis Entropy | 70.1 | 63.6 | 1.24 E+03 | ± | 2.16 E+02 | 1.40 E+03 | ± | 1.89 E+02 |
| Activity | 49.5 | 86.0 | 8.47 E+03 | ± | 3.44 E+03 | 5.97 E+03 | ± | 2.10 E+03 |
| Mobility | 44.9 | 87.9 | 1.58 E-02 | ± | 4.16 E-03 | 1.31 E-02 | ± | 2.33 E-03 |
| Complexity | 63.6 | 78.5 | 8.56 E+01 | ± | 2.21 E+01 | 1.08 E+02 | ± | 1.71 E+01 |

**Table 4.18.** Sensitivity (SN), specificity (SP), accuracy (ACC), and AUC of nonlinear and dynamic analysis features together using 1- and 5-minute windows of R-R interval signals from Medtronic version 1.0 dateset.

| R-R interval Length | SN (%) | SP (%) | ACC (%) | AUC |
|---|---|---|---|---|
| 1-minute | **78.6** | **64.3** | **71.4** | **0.77** |
| 5-minute | 71.4 | **64.3** | 67.9 | 0.67 |

## 4.5    Discussion and Conclusions

Modern ICDs allow for tiered therapy, permitting up to three rate detection ranges, each with a specific therapy function. By employing strategies such as burst pacing, adaptive burst pacing, and ramp pacing, many ventricular tachyarrhythmia events can be prevented from developing into sudden cardiac death through ATP [18], [122]. Detecting ventricular arrhythmias at early stages and delivering ATP as a non-aggressive inhibitor therapy can both improve patient quality of life and the lifespan of the ICD. Moreover, neither appropriate nor inappropriate ATP was associated with a significant mortality increase [16] and  ATP has been shown to be effective in terminating approximately 80% to 90% of spontaneous VT episodes [17], [50], [151], [152].

As many as fifty features have been proposed for sudden cardiac death prediction and are reviewed in this chapter. Moreover, these features are compared using a common database and classification approach. This comparison suggests that time domain features can achieve high performance alone, but the addition of more features using different approaches can increase performance, particularly with the shorter time window of one minute duration. A summary of the experimental results in different feature categories, and when all 50 features are combined, is presented in Table 4.19.

**Table 4.19.** Summary of experimental results from this chapter using Medtronic version 1.0 dataset.

| Feature Category (No. of Features) | 1-minute (R-R interval Length) | | | | 5-minute (R-R interval Length) | | | |
|---|---|---|---|---|---|---|---|---|
| | SN (%) | SP (%) | ACC (%) | AUC | SN (%) | SP (%) | ACC (%) | AUC |
| Time domain (6) | 78.6 | **78.6** | 78.6 | **0.88** | **85.7** | 82.1 | 83.9 | 0.83 |
| Frequency domain (4) | 57.1 | 67.9 | 62.5 | 0.66 | 53.6 | 64.3 | 58.9 | 0.64 |
| Bispectrum Analysis (31) | 78.6 | 71.4 | 75 | 0.80 | 82.1 | 67.9 | 75 | 0.77 |
| Nonlinear and Dynamic Analysis (9) | 78.6 | 64.3 | 71.4 | 0.77 | 71.4 | 64.3 | 67.9 | 0.67 |
| All Features (50) | **82.1** | **78.6** | **80.4** | 0.86 | **85.7** | **85.7** | **85.7** | **0.85** |

SN = sensitivity; SP = specificity; ACC = accuracy; AUC = area under curve

**Table 4.20.** Summary of experimental results from previous research, along with results from this chapter (including results obtained with Medtronic version 1.0 dataset).

| Previous Works | SN (%) | SP (%) | ACC (%) | R-R Interval Length |
|---|---|---|---|---|
| Thong *et al.*, 2007 [135] | 69 | 91 | No result reported | 108-minute |
| Rozen *et al.*, 2013 [116] | 50 | **91.6** | No result reported | 5-minute |
| Wollmann *et al.*, 2015 [89] | **94.4** | 50.6 | 70.9 | 5-minute |
| **Results using Medtronic version 1.0** | | | | |
| Mani *et al*, 1999 [91] | 76 | 76 | 76 | 10-minute |
| Bilgin *et al.*, 2009 [26] | 82.1 in LH and 70.2 in HF | 86.6 in LH and 74.6 in HF | No result reported | 10-minute |
| Joo *et al.*, 2012 [83] | 77.3 | 73.8 | 76.6 | 5-minute |
| Boon *et al.*, 2016 [84] | 77.94 | 80.88 | 79.41 | 5-minute |
| **Proposed Method** | **85.7** | **85.7** | **85.7** | 5-minute |
| | 82.1 | 78.6 | 80.4 | 1-minute |

SN = sensitivity; SP = specificity; ACC = accuracy

Table 4.20 presents a comparison of results obtained in previous works, as well as results obtained in the research presented in this chapter. Although windows of duration 4 to 5 minutes prior to ventricular arrhythmias are commonly used in previous works, very long intervals have also been examined, e.g. more than half an hour in [49], [135] with the best result being 69% sensitivity and 91% specificity. In the study presented in this chapter, the best result was achieved with 85.7% in sensitivity, specificity, and accuracy using all features considered, and with analysis window of 5-minute. In this comparative study, time domain features were found to have the highest AUC when used alone compared to the other feature categories reviewed, suggesting that these features achieve the best balance. However, addition of other feature categories was found to increase performance, suggesting that further development with appropriately chosen features is needed. Furthermore, using a lower number of features decrease the possibility of overfitting as in the current method, number of features is too high considering having only 78 patients in presented study. The next chapter extends this analysis to consider the most differentiating features in ventricular arrhythmia prediction.

# Optimizing Heart Rate Variability Features for Ventricular Arrhythmia Prediction

As noted earlier, approximately 80% of occurrences of sudden cardiac can be attributed to ventricular arrhythmias (VT-VF). The previous chapter carried out a comprehensive evaluation of HRV features for VT-VF prediction. This chapter extends this feature analysis with the goal of reducing unnecessary computational overhead by selecting only the most salient features to accurately predict a ventricular arrhythmia episode. The motivation for this is to facilitate more efficient implementation of arrhythmia prediction on resource-constrained devices such as ICDs. In order to reduce the computational overhead of the detection method, a feature selection technique is applied to reduce the number of features from an initial set. Different signal buffer lengths are examined, where 1-minute and 5-minute signal durations are processed. The features are then applied to classifiers.

To ensure there is a robust evaluation of the feature selection technique, the dataset is split into two, non-overlapping subsections. The first subset of data is used to rank and select the appropriate features and extract model tuning parameters. The second subset is then used to evaluate models using features chosen from the first section. The proposed method is evaluated using R-R interval patient data. All tests are carried out using a leave-patient-out cross-validation method, where the same patient's data cannot exist in both learning and evaluation data. This is done in order to have better generalisation in the proposed machine learning model for unseen data. Results from a selection of classification algorithms are compared. The work described in this chapter has been published in Parsi *et al.*, "Heart Rate Variability Feature Selection Method for Automated Prediction of Sudden Cardiac Death," *Biomed. Signal Process. Control*, vol. 65, p. 102310, Mar. 2021 [137].

## 5.1 Overview

Currently, there are several challenges to integrate a reliable rhythm management system in an ICD. If an ICD is configured incorrectly, it can administer inappropriate device therapy and endanger the patient and in some cases could in itself be the cause of a VT episode [58]. Misclassification of VT-VF episodes can result in false-positive detections in healthy patients and lead to improper ICD shock events [16]. Algorithms can be used to regulate ICD treatment by predicting and differentiating VT and VF arrhythmias from other cardiac rhythms in ICD patients. Such algorithms typically extract features from electrical signals measured by an ICD and use these features to make diagnostic decisions that in turn decide which treatment the ICD should apply. A key aim for any ICD detection technique is to register high sensitivity metric scores to ensure all arrhythmias are addressed as well as high specificity to reduce the number of inappropriate ICD therapies such as electrical shocks [14]. This, in practice, is a difficult balance to achieve.

In the last decade, ICD technology has advanced considerably. Now, multiple programming selection, long battery life, device history features and therapy control options are all standard features. Recently EGM signal shape analysis has been added [153]. SVT and VT-VF usually activate the ventricle in a distinct way, and as a result, different shapes and rhythms are expected from ventricle electrogram morphology as well. Many cardiac event detector algorithms have been applied to distinguish VT-VF from SVT and normal sinus rhythm. As mentioned in Chapter 4, these algorithms are sub-divided into two main categories: 1) morphological/EGM-based algorithms which use in/direct models of heartbeat signal and QRS complex [21]–[25], [134] and 2) those that rely on temporal intervals within the signal [26], [28], [84], [94], [154], [155].

Unlike morphological algorithms which have a significant computational cost, temporal classification methods use features with a lower computational overhead, such as rate, onset, stability and entropy from HRV signals [4], [26]–[28]. A review of algorithms from the literature was presented in [77] and summarized in Table 4.1. As discussed in Chapter 4 [77], each algorithm has its strengths and weaknesses, however, no consensus yet exists concerning the number of detection zones (based on HR changes), nor the optimal duration of signal used for analysis, for detection in order to decrease inappropriate shocks due to misclassification of SVT and VT-VF arrhythmias [14]. Interval based methods attempt to predict onset of an event by analysing a time window of cardiac signal that can range from 10 seconds [26], [28], [84], [91] up to as high as sixty minutes [49], [135] in advance of a VT-VF event. Ventricular arrhythmia prediction/detection at an early stage can allow ICDs to deliver ATP as a non-aggressive inhibitor therapy, improving the patient's quality of life and the ICD lifespan.

An overview of ventricular arrhythmia detection and treatment by ICD is presented in Figure 5.1. The detection and treatment algorithms typically involve a sequence of event detections. By adding an effective ventricular arrhythmia prediction method into an ICD there are opportunities to prevent unnecessary shocks [156]. Daubert *et al.* in [16] have found that the occurrence of false positives was not associated with a significant mortality increase. On the other hand, as mentioned in Chapter 4, ATP that is only applied when needed (true positive) has been effective in stopping approximately 80% to 90% of spontaneous VT episodes [17], [50], [151], [152]. However, there is a trade-off between early prediction and the buffering overhead available on a computationally limited ICD.

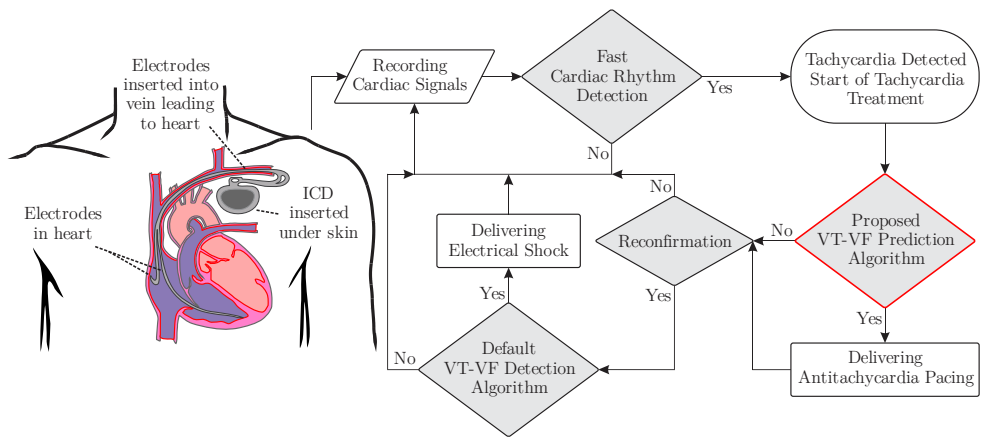

**Figure 5.1.** Flowchart describing the decision-making flow inside the ICD and where VT-VF prediction takes place in this flow. Having accurate VT-VF prediction make the ICD to deliver ATP, which can finally lead cardiac system to go to normal heartbeat. Normal heartbeat would fail tachycardia reconfirmation step which makes the ICD skip VT-VF detection and avoiding un-necessary electrical shock, while saving battery life and improve patient's life experience and quality.

An effective VT-VF prediction method can be used to complement detection and may be used to trigger ATP as a less aggressive therapy that controls heartbeats. The prediction method proposed in this section targets ventricular arrhythmias which could be prevented by nonaggressive treatment like ATP, with more invasive electrotherapy reserved only for cases that absolutely require it. For example, the PainFree I and II trials showed the effectiveness and safety of the application of ATPs (2 sequences of an 8-pulse burst pacing train at 88% of the VT cycle length) before shock delivery even on fast VT with a heart rate of 188-250 beats per minute [17]–[19]. Daubert *et al.* [16] have found that the occurrence of false positives was not associated with a significant mortality increase. On the other hand, ATP that is only applied when needed (true positive) has been effective in stopping approximately 80% to 90% of spontaneous VT episodes [17], [50], [151], [152]. However, there is a trade-off between early prediction and the buffering overhead available on a computationally limited ICD. Sensitivity (percentage of VT-VF candidate patients who are correctly identified as such) and specificity (percentage of healthy people who are identified as healthy) metrics are used to assess VT-VF interval prediction methods. A detection method normally attempts to maximize both values and achieve an appropriate balance between them depending on the clinical requirements. Boon *et al.* in [84] used a genetic algorithm to achieve 77.94% sensitivity and 80.88% specificity using a subset of features derived from an original set of 53 extracted features. Rozen *et al.* in [116] used a decision rule based on multipole analysis and reported over 94% sensitivity, however specificity was significantly lower at 51%. With 11 features, Lee *et al.* in [94] recorded 70.6% sensitivity and 76.5% specificity. The work in Chapter 4 [77] achieved 85.7% sensitivity and 85.7% specificity using 50 HRV features. While the metric results are high, the significant number of feature calculations are a significant challenge for an ICD with limited computational and storage resources.

Feature selection to search for the best subset of attributes in observed data may be an important and advantageous factor when applying machine learning methods on portable sensing devices. A minimum but representative set of features can simplify classification models, reduce the dimensionality, and enhance generalization. More recent machine learning models have reduced data and storage requirements, and improved prediction performances [84]. Many approaches, from genetic algorithm feature selection to principal component analysis, have been investigated to find a suitable HRV feature set [92], [157]. For example, Boon *et al.* in [84] used 20 features from a larger set of 53 features to predict onset of VT-VF. Lee *et al.* in [92] selected 14 most representative features for congestive heart failure recognition out of 42 features. Age and gender were among the selected fourteen features. Figure 5.2 is a radar diagram summarising the results of some previously-reported studies, along a number of different performance axes.

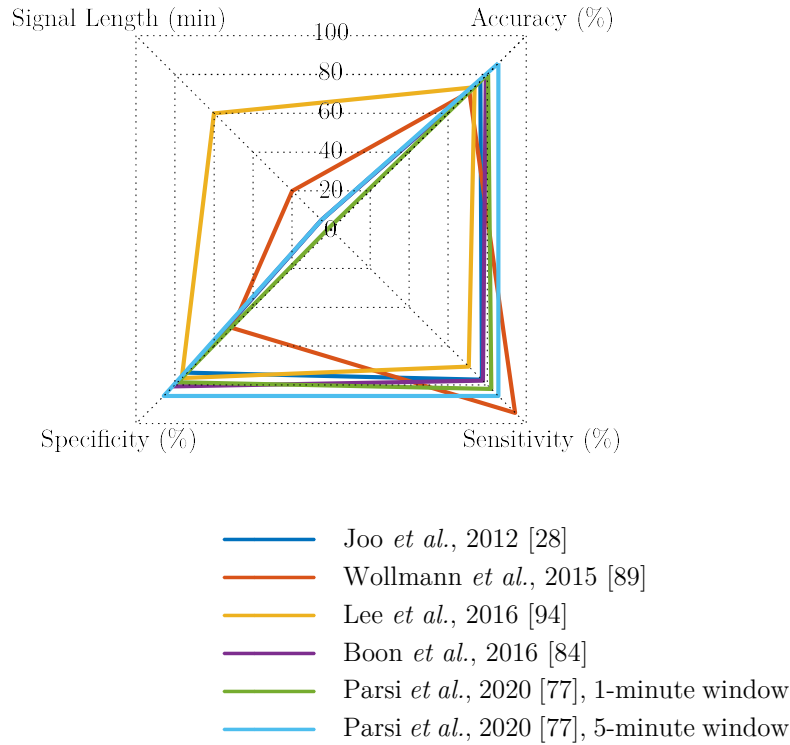

**Figure 5.2.** Best results from recent studies on VT-VF prediction.

## 5.2    HRV Feature Selection Methods

As each R-R interval signal is presented in terms of $d$ features (or measurements), it is viewed as point in $d$-dimensional space. The main goal here is to choose those features that present patterns from different classes in a well separated way [158]. The effectiveness of the representation space also is determined by how well classifiers could establish decision boundaries in the feature space which separate patterns belonging to different classes. There are many potential benefits of feature selection including facilitating data understanding, reducing the measurement and storage requirements, reducing training and utilization times, and last but not least addressing the "curse of dimensionality" to improve prediction performance [159].

Feature selection techniques can be divided into filters, wrappers, and embedded methods [160]. Wrappers utilize the machine learning algorithm of interest as a black box to score subsets of variables according to their predictive power. Filters select subsets of variables as a pre-processing step, independently of the chosen classifier or predictor. Embedded methods perform variable selection in the process of training and are usually specific to given learning machines. In both wrappers and embedded methods, a classifier plays an important role to either rank or advance the learning process respectively based on a given features subset. In the filter methods, however, the ranking and selection processes are done without classifiers. In this research a hybrid algorithm based on filter and wrapper methods has been proposed to rank and select the best subset of features, starting from the set of 50 features already described in Chapter 4.

In the first step, Student's *t*-test is used for the first selection based on p-values. The features with p-value more than 0.005 are removed from the feature set. Then using minimal redundancy-maximal relevance (mRMR) [161] or infinite latent feature selection (ILFS) [162] methods, features are ranked based on their ability to represent VT-VF or healthy classes, which can be viewed as a filter method. Finally, threshold determination which is a greedy approach[9] has been used to find the optimum feature set for each classifier. Threshold determination uses leave-one-patient-out cross-validation method on the learning database. This part is important for selecting the final subsets of features. Greedy search strategies seem to be particularly computationally advantageous and robust against overfitting. They are divided into forward selection and backward elimination. In forward selection, variables are progressively incorporated into larger and larger subsets, whereas in backward elimination one starts with the set of all variables and progressively eliminates the least promising ones [159]. Here the forward selection has been used to find the first performance maximum for each classifier, which determines the number of features for each method. Each of these is now described in more detail.

### 5.2.1 Student's *t*-Test

The Student's *t*-test can be used to determine whether the means of two sets differ significantly [148] and yields the p-value and t-values for the features extracted from each group. Statistically, a low p-value is preferred (p <0.005) [149], and the higher the t-value, the better the ranking of that feature. Previous studies have used the Student's *t*-test as a feature selection method in itself [142], [163]. In the proposed method, firstly, the Student's *t*-test is applied to remove undistinctive variables from the features. This approach has previously been used as a HRV feature-ranking method [28], [47], [86], [89],

---

[9] The name greedy comes from the fact that one never revisits former decisions to include (or exclude) variables in light of new decisions.

[142], [143], [147]. Instead, in this research this test is used as a pre-processing step to remove statistically undistinctive features. Features with p-values less than 0.005 will be deemed to pass the Student's *t*-test and considered to be a discriminative metric to be processed by one of the feature selection methods.

### 5.2.2 Minimal Redundancy-Maximal Relevance

In the second step, the minimal redundancy-maximal relevance method (mRMR for short) is implemented to rank the features according to their maximal statistical dependency on the target class [161]. In this scheme the aim is to maximize the mutual information between the feature distribution and classes and, at the same time, minimize the redundancy between features. In the first step, maximal relevance criterion (Max-Relevance) calculates the approximate value for maximal mutual information. Max-Relevance aims to derive a subset $S$ of features $x_i$ representative of a target class $c$ as follows:

$$max\ D(S,c),\ \ D = \frac{1}{|S|} \sum_{x_i \in S} I(x_i; c) \qquad (5.1)$$

where $I(x_i; c)$ is the mutual information between feature $x_i$ and class $c$. Mutual information between two random variables is defined as:

$$I(x,y) = \int \int p(x,y) \log \frac{p(x,y)}{p(x)p(y)} dx dy \qquad (5.2)$$

where $p(x,y)$ is the joint probability distribution of two random variables $x$ and $y$.

Even after this stage, there could still be a significant dependency between the selected features, where additional features could be removed without losing any class-discriminative power in classification phase. Mutually exclusive features can be extracted by additionally using the minimal redundancy (Min-Redundancy) condition:

$$min\ R(S),\ \ R = \frac{1}{|S|^2} \sum_{x_i, x_j \in S} I(x_i, x_j) \qquad (5.3)$$

By combining these two conditions, mRMR can be calculated to optimize $D$ and $R$ simultaneously as follows:

$$max\ \Phi(D,R),\ \ \Phi = D - R \qquad (5.4)$$

### 5.2.3 Infinite Latent Feature Selection

By way of comparison, the infinite latent feature selection (ILFS for short) has also been applied. IFLS is a robust probabilistic latent graph-based method which performs the feature ranking using all possible feature subsets [162]. Each feature in this method is a node in a graph and each weighted connection in this graph relates to the features' relevance. Weights are derived automatically using the probabilistic latent semantic analysis presented in [164] and then ranked using infinite feature selection method presented in [165].

Given a training set $X$ represented as a set of feature distributions $X = \{\vec{x}_1, \ldots, \vec{x}_n\}$, where each $m \times 1$ vector $\vec{x}_i$ is the distribution of the values assumed by the $i^{th}$ feature with regards to the $m$ samples, we build an undirected graph $G$, where nodes correspond to features and edges modelling relationships among pairs of nodes. Let the adjacency matrix $A$ associated with $G$ define the nature of the weighted edges: the elements $a_{ij}$ of $A$, $1 \leq i, j \leq n$, model pairwise relationships between the features. Each weight represents the likelihood that features $\vec{x}_i$ and $\vec{x}_j$ are good candidates [162]. Given the weighted graph $G$, the ILFS approach analyses subsets of features via paths connecting them. The cost of each path is given by the joint probability of all the nodes belonging to it. The method exploits the convergence property of the power series of matrices as in [165], and evaluates in an elegant fashion the relevance of each feature with respect to all the other ones taken together.

## 5.4     Proposed Feature Selection Method

The goal of this chapter is to develop a method for ventricular arrhythmia prediction in different time periods before onset of VT and VF, with consideration of the implications for implementation in CRM devices such as an ICD device, and with a particular focus on efficient feature representations. There are two phases, a first learning phase that uses a "learning" dataset and second evaluation phase that uses "evaluation" dataset. In the learning phase, the optimum feature selection method is implemented. The output comprises the best feature sets for each classification method along with optimized training parameters. Using learning dataset, each feature ranking method will derive a different order of features within the set of 50 features described in Chapter 4. The number of features ultimately used for classification is determined after feature ranking using cross-validation method implemented in the learning phase.

As noted above, in the first step, Student's $t$-test is used for an initial selection, whereby features with p-value more than 0.005 are removed from the feature set. Then using mRMR or ILFS, features are ranked based on their ability to represent VT-VF or healthy classes. Finally, threshold determination to find the best optimum features set for each classifier uses leave-one-patient-out cross-validation method on the learning database. Prediction is evaluated using a selection of classifiers.

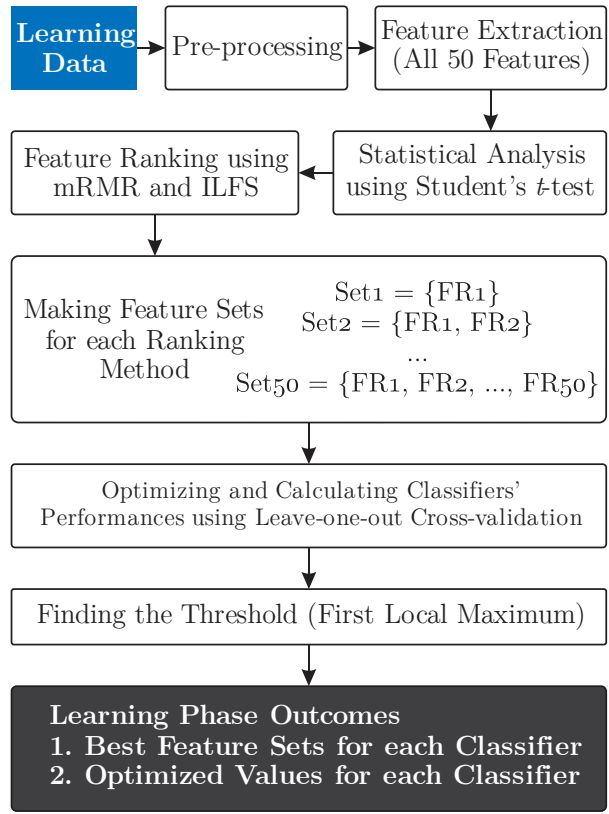

(a) Overview of proposed VT-VF prediction learning method

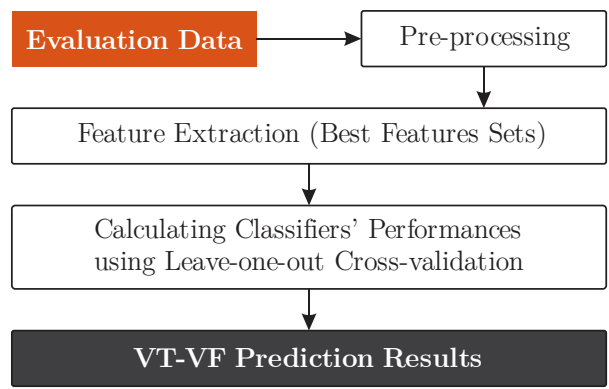

(b) Overview of proposed VT-VF prediction evaluation method

**Figure 5.3.** Overview of proposed VT-VF prediction learning (a) and evaluation (b) phases. Using the proposed method, learning the correct features, and evaluation of prediction performance are separated.

After running the learning phase which produces the best feature sets for each classifier along with optimized classifiers' parameters, in the second phase, evaluation is carried out using a second dataset called the "evaluation" dataset. Here, each proposed feature set for each classifier is tested using leave-one-out cross-validation on evaluation data. The learning and evaluation data are completely separated and randomly selected from main dataset. The block diagram of the proposed method is given in Figure 5.3. The main components are now described in more detail.

### 5.4.1 Proposed Dataset Division Algorithm

As for Chapter 4, the clinical data in this chapter are obtained from R-R interval signals within the spontaneous ventricular tachyarrhythmia database Medtronic version 1.0 from Medtronic [79]. There are 106 pre-VT records, 29 pre-VF records, and 126 CON records data points. Two-thirds of the data are used in learning phase for feature selection, while the remainder serves as evaluation data. Any individual patient's signals can only reside in either the learning or evaluation set. This exclusive allocation may reduce overfitting and represents real-life scenarios where a medical device may not have access to a test subject's previous HRV records. To ensure that patients are uniquely allocated, the following procedure is observed:

1. Select 67% of patients (52 patients) using a random seed generator.
2. Calculate the total number of VT and VF occurrences from the selected 52 patients.
3. If VT total number is not equal to 71 events (67% of total VT) return to (1) else continue.
4. If total VF number is not equal to 19 events (67% of VF events) return to (1) else continue.
5. The final selection of the 52 patients, with all their events (VT, VF and CON data) are used to make the learning set and the remaining, unselected, patients are used to make the evaluation set.

### 5.4.2 Improved Signal Pre-processing and Noise Reduction

Noise can have a negative effect on the performance of implantable cardiac devices such as pacemakers and defibrillators. Previous research shows that electromagnetic interference due to GSM cellular phones might have adverse effects under certain condition on pacemakers and implantable cardioverters during both in vivo and in vitro studies [166]–[168]. Electronic surveillance [169], household appliances [170] and muscular activity [171] are all noise sources that can adversely affect the received signals of implantable cardiac devices like ICDs.

Detecting and removing noise has been shown to reduce inappropriate shocks by over 90% [172]. In the experimental dataset used in this work, R-R interval signals obtained from an ICD device contain signal spikes which present as noise or as ectopic beats.

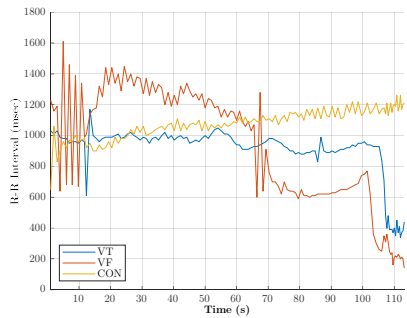

(a) R-R interval signals. VT, VF, and CON samples from Patient No. 8010.

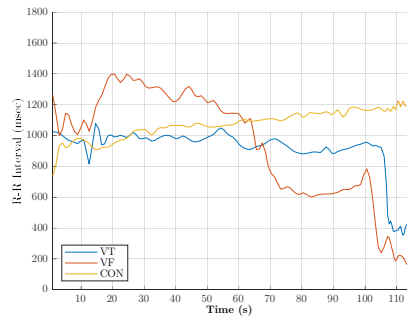

(b) After Wavelet Filter. VT, VF, and CON samples from Patient No. 8010.

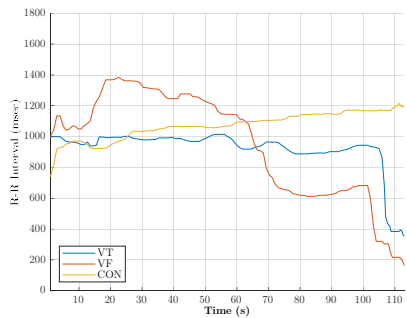

(c) After Median Filter. VT, VF, and CON samples from Patient No. 8010.

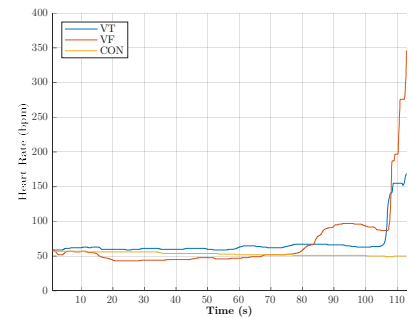

(d) IHR signals after conversion and resampling (16 Hz). VT, VF, and CON samples from Patient No. 8010.

**Figure 5.4.** Pre-processing of signals from patient No. 8010 from Medtronic version 1.0 dataset, including noise reduction and resampling. Note the removal of high-frequency complnents evident in (a) by filtering. **Figure 3.1** shows the same patient's R-R interval and IHR signals without any pre-processing.

A noise reduction scheme, based on the wavelet transform and median filter, is adapted to remove any artefact and ectopic beats , similar to [136], [173]. Using high and low pass filters, the wavelet transform can remove the higher frequencies of the background noise from the R-R interval signals. In this research, the first approximation level of wavelet transform is taken, and the sym8 wavelet is used [173]. Figure 5.4 (a) shows examples of VT, VF and CON R-R interval samples of patient No. 8010 in the dataset. Figure 5.4 (b) shows the VT, VF and CON R-R intervals after the wavelet transform. Then, a median filter is used to eliminate remaining spike beats with the resultant signal shown in Figure 5.4 (c). The IHR signal, a measure of variation in the heart rate, is then calculated using the formula (3.1) in Chapter 3 (Figure 5.4 (d)). Resampling these signals to a uniform rate of 16 Hz using shape-preserving piecewise cubic interpolation is the final step to prepare input data for the feature extraction process, as shown in Figure 5.4 (d).

### 5.4.3 Signal Classification and Cross-validation

Three different classifiers were applied to predict onset of VT-VF episodes by classifying each sample in one of two classes, either normal or VT-VF sample. The classifiers are SVM, k-nearest neighbours (k-NN) and random forest (RF) classifier. These classifiers were used in both the learning phase for feature selection, and in the evaluation phase.

#### 5.4.3.1 Support Vector Machine

Historically, SVM has shown to be an effective multi-class discriminator [174] and previous studies have highlighted its efficacy with HRV signal classification [175]. Simply stated, the SVM identifies the best separating hyperplane (the plane with maximum margins) between the two classes of the training samples within the feature space. In this way, not only is an optimal hyperplane is fitted, but also fewer training samples are used; thus, high classification accuracy is achieved with small training sets. SVM performs well when the separating hyperplane is nonlinear [158]. The main idea of SVM is to map the input data from the N-dimensional input space, through some nonlinear mapping, to the M-dimensional feature space M > N, where the data classes can be linearly separated [105]. This step in SVM happens using the help of a kernel function. The choice of kernel functions and kernel parameters depends mainly on the application. There are several kernel functions commonly used: Gaussian, radial basis functions, polynomial, and linear. A linear SVM classifier provides the optimal separating hyperplane within a feature space. In this chapter we adopt a linear SVM to reduce the computational footprint and reduce the risk of overfitting.

Each kernel function in an SVM has an associated kernel scale and box constraint. Kernel scale is a scaling parameter for the input data to be scaled with respect to the features' range before being applied to the kernel function.

So, the range of some features cannot be dominant in the kernel calculation. Box constraint, on the other hand, is the parameter that controls the maximum penalty imposed on margin-violating observations. Box constraint aids in preventing overfitting. If it increases, then the SVM classifier assigns fewer support vectors, however, increasing the box constraint can lead to longer training times. Both values, kernel scale and box constraint were defined to achieve the best classification result. To do this, the Bayesian optimization algorithm has been used. Accordingly, a linear kernel function has been used with kernel scale set to 1 and box constraint set to 1.6 [78].

### 5.4.3.2    K-Nearest Neighbours

Among the various methods of supervised statistical classification, the nearest neighbour rule is one of the most commonly used discriminators [176]. The k-NN is an extended version of this rule that has been used in many studies [140], [143], [177]–[179]. The k-NN method can be combined with different search algorithms such as genetic algorithm [180]. A new sample is classified by determining the distances to the $k$ nearest training cases. The associated class of each sample will be that one which has the greatest repetition in the $k$ nearest samples. Considering algorithm description, the k-NN is a non-parametric algorithm which means either there are no parameters or fixed number of parameters irrespective of size of data [181]. A k-NN could be the best choice for a classification study that involves little or no prior knowledge about the distribution of the data. This classification algorithm stores all training data and waits until having the test data produced, without having to create a learning model [182].

Despite its simple characteristic, surprisingly, k-NN is used extensively for data classification and even in big data analysis [183]. This is due to reasonable accuracy of k-NN classification algorithm [184]. However, choosing the optimal number of neighbours along with best distance or similarity measure is an important problem as the performance of the k-NN classifier is dependent on them. The distance function between two vectors $x$ and $y$ defines the distance between both vectors as a non-negative real number. This thesis considers the three major distance measurements that are special cases of Minkowski distance, corresponding to different values of $p$ for this power distance [184]. For two vectors $x$ and $y$, where $x = (x_1, x_2, \ldots, x_n)$ and $y = (y_1, y_2, \ldots, y_n)$ the general Minkowski distance is defined as:

$$D_{Mink}(x, y) = \sqrt[p]{\sum_{i=1}^{n} |x_i - y_i|^p} \tag{5.5}$$

where $p$ is a positive value.

When $p = 1$, the distance becomes the Manhattan distance, the sum of the absolute differences between the vectors. When $p = 2$, it becomes the well-known and widely used Euclidean distance. Finally, the Chebyshev distance, also known as maximum value distance, is a variant of Minkowski distance where $p = \infty$. In this case, the distance between two vectors is the greatest of their differences along any coordinate dimension [185]. After running the Bayesian optimization process on the learning set the best number of neighbours has been set to 5 and the Chebyshev distance has been applied in this study.

### 5.4.3.3 Random Forest

The method of random forests comprises an ensemble of decision trees, and has previously been used to distinguish various cardiac arrhythmias [186]. As individual decision trees tend towards overfitting, a random forest reduces the effect of overfitting and improves generalization by combining the results of several decision trees [187]. Each tree in a random forest has been grown using bootstrap samples of data. For every input, every tree outputs a classification value and a vote for a specific class. In classification problems, instead of averaging, the majority vote of all trees is taken to output the predicted label [188]. In this work, after applying Bayesian optimization process, each classifier has been set to 64 trees with leaf size equal to 4.

### 5.4.3.4 Cross-validation

As the supply of data for training and testing is often limited in publicly-available datasets gathered from ICDs, we wish to use as much of the available data as possible for training in order to build a good learning model for each classifier. However, if the validation set is small, it will give a relatively noisy estimate of predictive performance. One of the mostly used solution to this dilemma is to use cross-validation [105]. Cross-validation is a resampling procedure used to evaluate machine learning models on a limited dataset. This allows a proportion $(K-1)/K$ of the available data to be used for training while making use of all the data to assess performance. Importantly, each observation in the data sample is assigned to an individual group and stays in that group for the duration of the procedure. This means that each sample is given the opportunity to be used in the left-out set one time and used to train the model $K-1$ times as it is presented in Figure 5.5.

**Figure 5.5.** The technique of K-fold cross-validation has illustrated here for the case of $K = 3$. It involves taking the available data, shuffle it randomly and splitting it into $K$ groups (in the simplest case these are of equal size). For each unique group use the

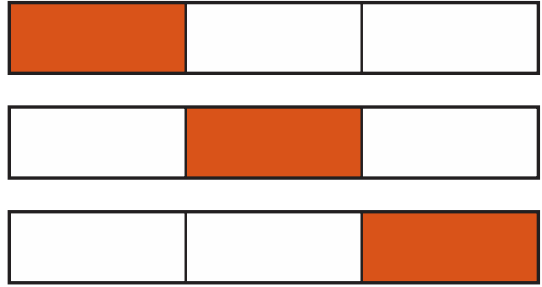

group as test dataset while the other $K - 1$ of the groups are used to fit and train a model. Retain the evaluation score and discard the model. This procedure is then repeated for all $K$ possible choices for the held-out group, indicated here by the red blocks, and the performance scores from the $K$ runs are then averaged.

When data is particularly scarce, it may be appropriate to consider the case $K = N$, where $N$ is the total number of patients in current thesis, which gives the leave-one-patient-out technique. Cross-validation is primarily used in machine learning to estimate the generalisation properties of a machine learning model on unseen data.

## 5.6    Experimental Results

In this study, features were extracted from two different time intervals of HRV signals prior to ventricular arrhythmia onset. Signals are taken from between two points ten seconds prior to VT-VF event onset. There are one-minute signals and five-minute signals. After applying student's *t*-test on all features with p-value less than 0.005% (which equals to 99.995% confidence level), ten features were removed from the 5-minute HRV signals, and three features removed when considering 1-minute HRV signals. The pre-processing and feature extraction were implemented and developed in MATLAB. For the feature analysis and selection parts, the MATLAB Statistics Toolbox along with the feature selection toolbox in [189] were used. The classifying and optimization are also implemented in MATLAB.

After Student's *t*-test filtering, the two different feature ranking methods described above (mRMR and ILFS) were applied to the Learning Dataset. The chosen classifiers evaluate different numbers of features to determine an optimal number of features for each method, balancing performance, and complexity. In this step we have used leave-one-patient-out cross-validation method on the Learning Dataset, in which there are 52 patients overall. On each round, the classifiers were trained with 51 patients and validated with one patient that has been left out. Therefore, the process repeats 52 times, and the average of performance metrics has been calculated for feature selection.

Results in this section are presented in terms of sensitivity, specificity, accuracy, and ROC. The ROC curves are calculated as well as the accuracy and area under the ROC curve (AUC) for each test condition [150]. As well as a class label, our method can also output a 'confidence' level for each test, which is the signed output of the classifier. The ROC curve captures the true positive rate (TPR or $sensitivity$) and a false positive rate (FPR or $1 - specificity$) as a function of decision threshold [150]. The curve illustrates how the classifier's confidence relates to its success and how well the model is capable of distinguishing between classes. A sample ROC curve is shown in Figure 5.6. Better performance maximizes the distance from the reference line at each point ($TPR \rightarrow 1$). AUC measures the area underneath the ROC curve to give an overall measure of performance [150]. AUC varies between 0 and 1.

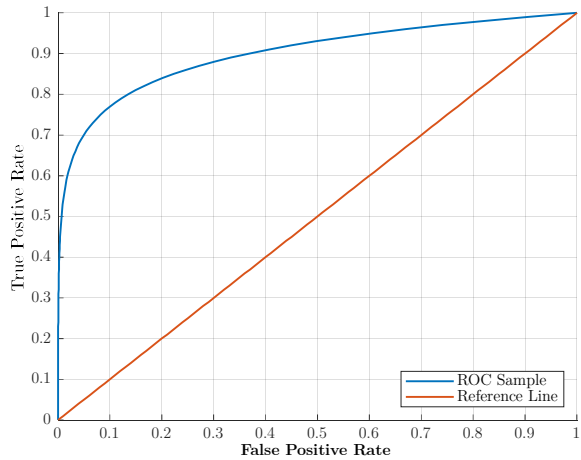

**Figure 5.6**. Sample ROC curve with Reference Line

### 5.6.1    Learning Phase Results

During the learning phase, the optimal number of features for each combination of feature selection method and classifier is determined experimentally (6 combinations in total). Classification performance is evaluated as a function of increasing numbers of features, and the optimal number of features is that number corresponding to the first local maximum in the graph of accuracy versus number of features, as shown in Figure 5.7 for 5-minute of R-R interval signals and in Figure 5.8 while using 1-minute of R-R interval signals. It is important to mention the power of individual features to distinguish or represent each class (normal or VT-VF). So, when two of them are added to the feature set one after each other, the amount of improvement is sometimes minimal (or may even result in a degradation). By running leave-one-out cross-validation on every step of adding a feature, fluctuations in the performance curve have been reduced somewhat, therefore in some cases adding features does not make any difference to the classifier performance graphs. The optimal feature sets for each classifier and feature selection method, identified in Figure 5.7 and Figure 5.8 by the vertical green dotted lines, are used in the evaluation phase described in the next sub-section.

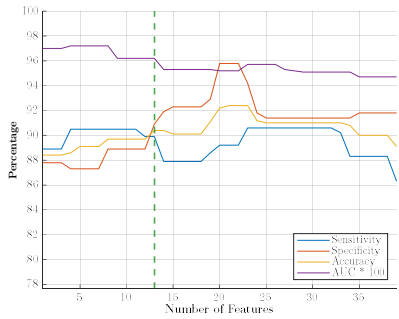

(a) Learning cross-validation results using SVM classifier with threshold: first 13 features

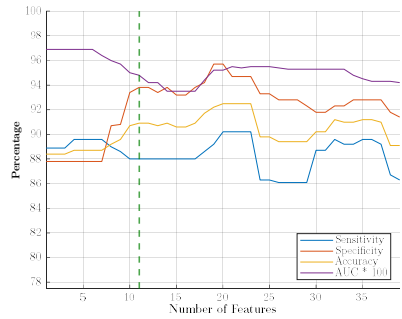

(d) Learning cross-validation resultsu sing SVM classifier with threshold: first 11 features

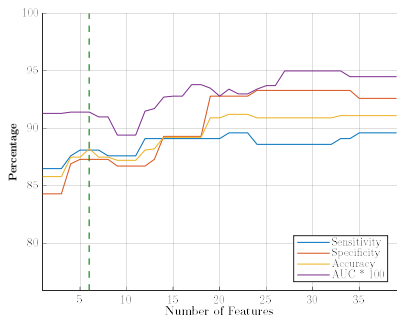

(b) Learning cross-validation results using k-NN classifier with threshold: first 6 features

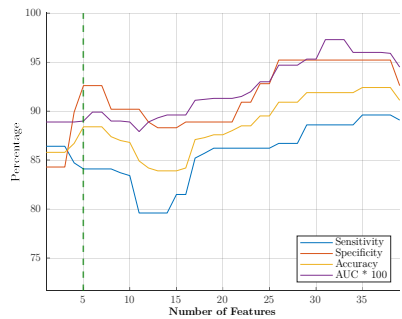

(e) Learning cross-validation results using k-NN classifier with threshold: first 5 features

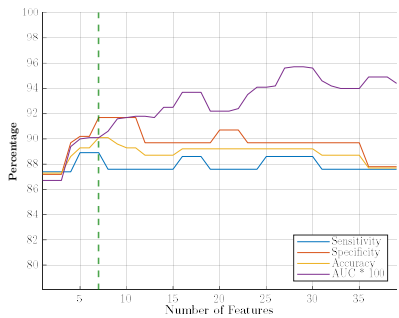

(c) Learning cross-validation results using RF classifier with threshold: first 7 features

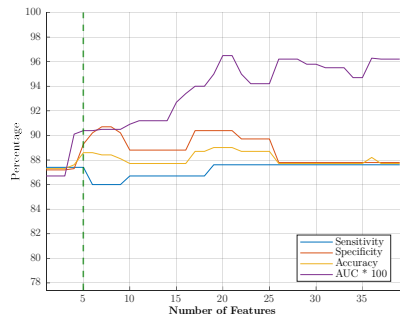

(f) Learning cross-validation results using RF classifier with threshold: first 5 features

**Figure 5.7.** Leave-one-patient-out cross-validation method results using 5-minute of R-R interval signals on Learning Dataset using mRMR ((a), (b), and (c)) and ILFS ((d), (e), and (f)) method ranking, with optimal number of features for evaluation phase indicated by vertical green dotted lines.

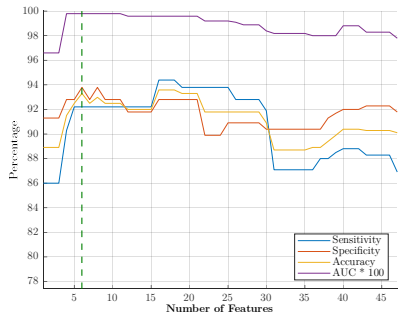

(a) Learning cross-validation results using SVM classifier with threshold: first 6 features

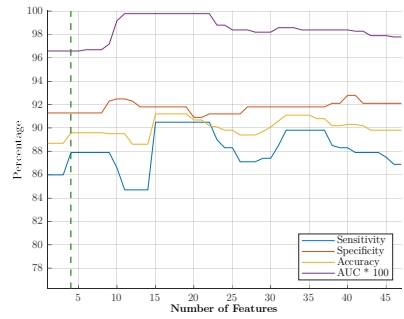

(d) Learning cross-validation results using SVM classifier with threshold: first 4 features

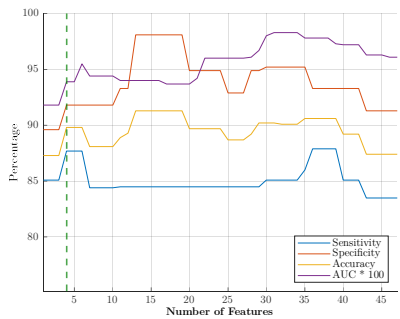

(b) Learning cross-validation results using k-NN classifier with threshold: first 4 features

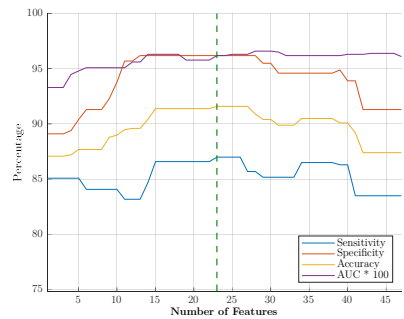

(e) Learning cross-validation results using k-NN classifier with threshold: first 23 features

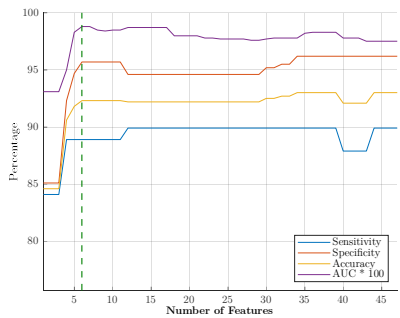

(c) Learning cross-validation results using RF classifier with threshold: first 6 features

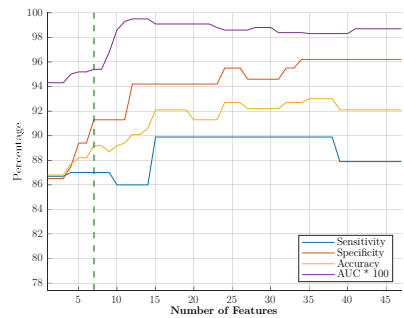

(f) Learning cross-validation results using RF classifier with threshold: first 7 features

**Figure 5.8.** Leave-one-patient-out cross-validation method results using 1-minute of R-R interval signals on Learning Dataset using mRMR ((a), (b), and (c)) and ILFS ((d), (e), and (f)) method ranking, with optimal number of features for evaluation phase indicated by vertical green dotted lines.

In Table 5.1 the Student's *t*-test, mRMR, and ILFS raking values are presented to show a detailed comparison of ranking methods of the features after the Student's *t*-test phase for 5-minute R-R interval signals. Shaded rows in Table 5.1 are the ones which are selected using each methodology and the numbers in the columns marked mRMR and ILFS represent features rankings for each method. Shaded rows are the ones which are selected using each methodology and numbers under mRMR and ILFS columns represent feature ranking after Student's *t*-test phase. Only 39 features passed the Student's *t*-test and were then ranked based on either mRMR or ILFS methods.

**Table 5.1.** The p-value, Student's *t*-test (with probability less than 0.005), and selected features based on mRMR and ILFS methods for each classifier (SVM, k-NN and RF) using 5-minute of R-R interval signals from Medtronic version 1.0 dateset.

| Features Category and Names | | p-Value | *t*-test < 0.005 | mRMR Method | | | ILFS Method | | |
|---|---|---|---|---|---|---|---|---|---|
| | | | | SVM | k-NN | RF | SVM | k-NN | RF |
| **Time domain** | MeanNN | < 0.001 | 1 | 30 | 30 | 30 | 21 | 21 | 21 |
| | SDNN | < 0.001 | 1 | 6 | 6 | 6 | 15 | 15 | 15 |
| | RMSSD | < 0.001 | 1 | 13 | 13 | 13 | 20 | 20 | 20 |
| | NN50 | 1.74E-01 | 0 | - | - | - | - | - | - |
| | pNN50 | 1.53E-01 | 0 | - | - | - | - | - | - |
| | HRVTri | 0 | 1 | 4 | 4 | 4 | 8 | 8 | 8 |
| **Frequency domain** | $PSD_{VLF}$ | < 0.001 | 1 | 11 | 11 | 11 | 28 | 28 | 28 |
| | $PSD_{LF}$ | < 0.001 | 1 | 17 | 17 | 17 | 27 | 27 | 27 |
| | $PSD_{HF}$ | < 0.001 | 1 | 7 | 7 | 7 | 30 | 30 | 30 |
| | $PSD_{LF/HF}$ | 5.00E-02 | 0 | - | - | - | - | - | - |
| **Bispectrum analysis** | $M_{avg}^{LL}$ | < 0.001 | 1 | 20 | 20 | 20 | 16 | 16 | 16 |
| | $M_{avg}^{LH}$ | < 0.001 | 1 | 27 | 27 | 27 | 13 | 13 | 13 |
| | $M_{avg}^{HH}$ | < 0.001 | 1 | 29 | 29 | 29 | 23 | 23 | 23 |
| | $M_{avg}^{ROI}$ | < 0.001 | 1 | 32 | 32 | 32 | 12 | 12 | 12 |
| | $P_{avg}^{LL}$ | 9.79E-02 | 0 | - | - | - | - | - | - |
| | $P_{avg}^{LH}$ | 1.96E-01 | 0 | - | - | - | - | - | - |
| | $P_{avg}^{HH}$ | 1.32E-01 | 0 | - | - | - | - | - | - |
| | $P_{avg}^{ROI}$ | 1.04E-01 | 0 | - | - | - | - | - | - |
| | $E_{nb}^{LL}$ | < 0.001 | 1 | 37 | 37 | 37 | 37 | 37 | 37 |
| | $E_{nb}^{LH}$ | 0 | 1 | 16 | 16 | 16 | 14 | 14 | 14 |
| | $E_{nb}^{HH}$ | 0 | 1 | 19 | 19 | 19 | 10 | 10 | 10 |
| | $E_{nb}^{ROI}$ | < 0.001 | 1 | 34 | 34 | 34 | 36 | 36 | 36 |
| | $E_{snb}^{LL}$ | < 0.001 | 1 | 12 | 12 | 12 | 26 | 26 | 26 |
| | $E_{snb}^{LH}$ | < 0.001 | 1 | 22 | 22 | 22 | 22 | 22 | 22 |
| | $E_{snb}^{HH}$ | 0 | 1 | 9 | 9 | 9 | 11 | 11 | 11 |
| | $E_{snb}^{ROI}$ | < 0.001 | 1 | 31 | 31 | 31 | 33 | 33 | 33 |
| | $L_{m}^{LL}$ | 0 | 1 | 28 | 28 | 28 | 6 | 6 | 6 |
| | $L_{m}^{LH}$ | 0 | 1 | 5 | 5 | 5 | 5 | 5 | 5 |
| | $L_{m}^{HH}$ | 0 | 1 | 10 | 10 | 10 | 3 | 3 | 3 |
| | $L_{m}^{ROI}$ | 0 | 1 | 8 | 8 | 8 | 2 | 2 | 2 |
| | $L_{dm}^{LL}$ | 0 | 1 | 18 | 18 | 18 | 7 | 7 | 7 |
| | $L_{dm}^{HH}$ | 0 | 1 | 1 | 1 | 1 | 1 | 1 | 1 |
| | $L_{dm}^{ROI}$ | 0 | 1 | 15 | 15 | 15 | 4 | 4 | 4 |
| | $WCOB_{iLL}$ | < 0.001 | 1 | 21 | 21 | 21 | 31 | 31 | 31 |
| | $WCOB_{iLH}$ | 5.31E-02 | 0 | - | - | - | - | - | - |
| | $WCOB_{iHH}$ | 7.39E-01 | 0 | - | - | - | - | - | - |
| | $WCOB_{iROI}$ | < 0.001 | 1 | 38 | 38 | 38 | 35 | 35 | 35 |
| | $WCOB_{jLL}$ | < 0.001 | 1 | 3 | 3 | 3 | 29 | 29 | 29 |
| | $WCOB_{jLH}$ | 0 | 1 | 24 | 24 | 24 | 19 | 19 | 19 |
| | $WCOB_{jHH}$ | < 0.001 | 1 | 25 | 25 | 25 | 34 | 34 | 34 |
| | $WCOB_{jROI}$ | < 0.001 | 1 | 39 | 39 | 39 | 32 | 32 | 32 |
| **Nonlinear and dynamic analysis** | $SD_1$ | 2.18E-03 | 1 | 35 | 35 | 35 | 39 | 39 | 39 |
| | $SD_2$ | 2.69E-01 | 0 | - | - | - | - | - | - |
| | $SD_1/SD_2$ | < 0.001 | 1 | 33 | 33 | 33 | 38 | 38 | 38 |
| | SampEn | 5.72E-01 | 0 | - | - | - | - | - | - |
| | Rényi Entropy | < 0.001 | 1 | 26 | 26 | 26 | 17 | 17 | 17 |
| | Tsallis Entropy | < 0.001 | 1 | 23 | 23 | 23 | 18 | 18 | 18 |
| | Activity | < 0.001 | 1 | 2 | 2 | 2 | 25 | 25 | 25 |
| | Mobility | < 0.001 | 1 | 36 | 36 | 36 | 24 | 24 | 24 |
| | Complexity | < 0.001 | 1 | 14 | 14 | 14 | 9 | 9 | 9 |
| **Threshold** | | | | 13 | 6 | 7 | 11 | 5 | 5 |

## 5.6.2    Evaluation Phase Results

After finding the optimal feature set for each classifier and feature selection method with the Learning Dataset, the Evaluation Dataset is used with the SVM, k-NN and RF models. Again, the leave-one-patient-out cross-validation method is used for the Evaluation Dataset. All performance metrics (sensitivity, specificity, accuracy, and AUC) have been calculated and presented in Table 5.2 for 5-minute and in Table 5.3 for 1-minute R-R interval signals. The predictions from 5-minute R-R interval signals show a significant improvement in terms of feature reduction, and balance between sensitivity and specificity. As shown in Table 5.2 and Table 5.3, k-NN with the mRMR selection method and SVM with the mRMR method achieved the best accuracy for 5-minute and 1-minute signal windows.

**Table 5.2.** Summaries of test results conducted in prediction of VT-VF using SVM, k-NN, and RF classifiers based on mRMR and ILFS feature ranking methods on using 5-minute of R-R interval signals from Medtronic version 1.0 dataset.

| Ranking Method | Classifier | No. of Features | SN (%) | SP (%) | ACC (%) |
|---|---|---|---|---|---|
| mRMR | SVM | 13 | 89.7 | 89.4 | 89.6 |
|  | **k-NN** | **6** | **88.8** | **94.2** | **91.5** |
|  | RF | 7 | 86.2 | 80.8 | 83.5 |
| ILFS | SVM | 11 | 85.5 | 88.5 | 87.0 |
|  | k-NN | 5 | 77.8 | 84.6 | 81.2 |
|  | RF | 5 | 79.0 | 76.9 | 78.0 |

SN = sensitivity; SP = specificity; ACC = accuracy

**Table 5.3.** Summaries of test results conducted in prediction of VT-VF using SVM, k-NN, and RF classifiers based on mRMR and ILFS feature ranking methods on using 1-minute of R-R interval signals from Medtronic version 1.0 dataset.

| Ranking Method | Classifier | No. of Features | SN (%) | SP (%) | ACC (%) |
|---|---|---|---|---|---|
| mRMR | **SVM** | **6** | **86.9** | **93.5** | **90.2** |
|  | k-NN | 4 | 84.8 | 84.6 | 84.7 |
|  | RF | 6 | 94.6 | 82.7 | 88.6 |
| ILFS | SVM | 4 | 86.9 | 82.7 | 84.8 |
|  | k-NN | 23 | 84.3 | 84.6 | 84.5 |
|  | RF | 7 | 79.8 | 83.3 | 81.6 |

SN = sensitivity; SP = specificity; ACC = accuracy

Using the k-NN there are only six features required to reach the best performance to predict VT-VF in 5-minute scenario. Two features from the time domain, SDNN and HRVTri, three features from bispectrum analysis, $L_m$ in LH, $L_{dm}$ in HH and $WCOB_j$ in LL, and one feature from nonlinear and dynamic analysis, namely activity from Hjorth's parameters, make up this group of features. By adding seven extra features with the SVM, the most balanced results (between sensitivity and specificity) on 5-minute with around 89.5% has been achieved. In the 1-minute HRV signal analysis, we obtain similar results for SVM classification method using the mRMR selection technique. With mRMR ranking, SVM used six: two features from time domain, Mean and HRVTri, three features from bispectrum analysis, $L_m$ in HH and $L_{dm}$ in LL and ROI, and one feature from nonlinear and dynamic analysis, Tsallis Entropy.

ROCs are shown in Figure 5.9. Although k-NN with mRMR ranking method has shown the best results overall, SVM performed slightly better because of better sensitivity, specificity, and accuracy in ROC and AUC metric in general. However, the k-NN uses only six features and has a lower computational cost which may render it more suitable for an embedded device such as an ICD.

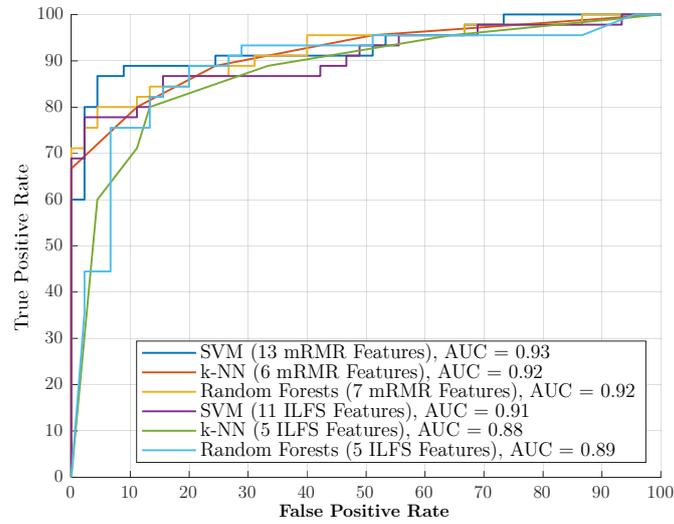

(a) ROC for each classification algorithm using first T features from mRMR and ILFS ranking methods while 5-minute of R-R interval signals have been used.

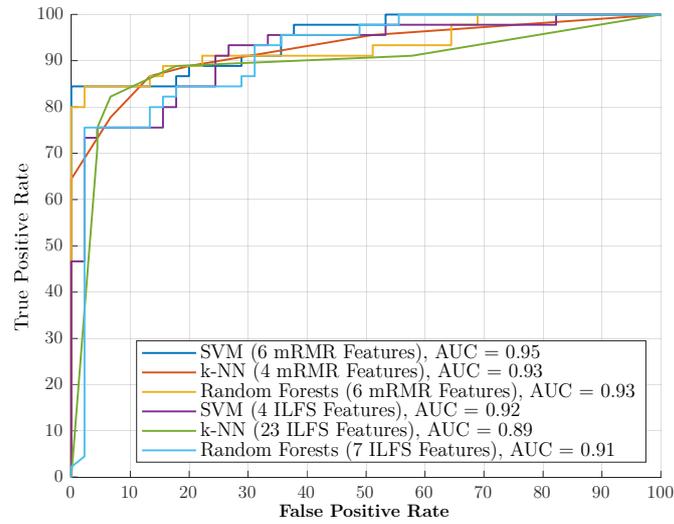

(b) ROC for each classification algorithm using first T features from mRMR and ILFS ranking methods while 5-minute of R-R interval signals have been used.

**Figure 5.9.** ROC results for both 1 and 5-minute of R-R interval signals analysis.

## 5.7    Discussion

This chapter investigated the use of feature selection techniques to improve the prediction rates of VT and VF for devices with limited computational capabilities, such as an ICD. The contributions herein are:

1. The application of feature selection techniques to HRV signals to predict VT and VF with a reduced set of features, and with performance comparable to using the full set of features.
2. The proposed method improves on the performance metrics published in the literature.
3. Good performance is achieved for a temporal VT and VF detection technique using as few as five features which is the lowest published within this field.

A summary comparison of previous works with the proposed methods is presented in Table 5.4. The longest time window, at 5-minute window, registered higher metric scores than classifications from the 1-minute data. Further research on the time-to-event duration may provide additional insights. In this study, 5-minute HRV signal analysis using k-NN with only six features achieved 88.8% in sensitivity and 94.2% in specificity. The best result however, 86.9% in sensitivity and 93.5% in specificity, has been generated using SVM with six features while processing 1-minute HRV signals. This result is a significant improvement in comparison with [84] that used 20 selected features, achieving 77.94% in sensitivity and 80.88% in specificity with the same window length on the same database. Considering that the six selected features in this study are a subset of feature sets used in previous studies [28], [77], [84], [94], coupled with the fact that simpler classifiers are used here (linear SVM and k-NN), the computational complexity of the system is decreased substantially. However, further detailed studies are required in order to quantify the numerical improvement that can be achieved using the proposed methods; this is a topic for future work.

Selected features in this study promisingly achieved a more balanced distribution of false positive and false negative results than related methods in the literature. The top ten features (ranked according to the learning phase) are listed in Table 5.5. From the table, there are some differences in mRMR ranking and type of the features in 1 and 5-minute analyses. From these features, Mean, SDNN and HRVTri are also in the final list of selected feature in [84], [92]. Table 5.5 highlights that there are 4 features in both 1 and 5-minutes analysis which are ranked in top ten features by mRMR ranking method: $L_m$ in HH, $L_{dm}$ in HH, HRVTri, SDNN. This work suggests that different frequency ranges, such as HH frequency in both 1 and 5-minute $L_m$ analysis aid the classifiers in distinguishing between sudden cardiac death events. As shown, 6% improvement has been reached while processing 5-minute signals.

**Table 5.4.** Summary of results from previous research with Medtronic version 1.0 dataset, and results presented in this chapter.

| Ranking Method | No. of Features | SN (%) | SP (%) | R-R Interval Length |
|---|---|---|---|---|
| Mani *et al.*, 1999 [91] | Frequency band | 76 | 76 | 10-minute |
| Bilgin *et al.*, 2009 [26] | Frequency band | 82.1 | 86.6 | 10-minute |
| Joo *et al.*, 2012 [28] | 11 | 77.3 | 73.8 | 5-minute |
| Wollmann *et al.*, 2015 [89] | 4-9 | 94.4 | 50.6 | 20-minute |
| Lee *et al.*, 2016 [94] | 14 | 70.6 | 76.5 | 60-minute |
| Boon *et al.*, 2016 [84] | 20 | 77.9 | 80.8 | 5-minute |
| Parsi *et al.*, 2020 [77] | 50 | 85.7 | 85.7 | 5-minute |
| **Proposed Method** | **6** | **88.8** | **94.2** | **5-minute** |
| | **6** | **86.9** | **93.5** | **1-minute** |

**Table 5.5.** Summaries of top ten features from mRMR ranking processing 1 and 5-minute of R-R interval signals. Shaded rows are the ones which are common between 1 and 5-minute analysis.

| Features Rank | 1-minute of R-R Interval Signals | 5-minute of R-R Interval Signals |
|---|---|---|
| 1 | $L_m$ in HH | $L_{dm}$ in HH |
| 2 | Tsallis Entropy | Activity |
| 3 | $L_{dm}$ in LL | $WCOB_j$ in LL |
| 4 | Mean | HRVTri |
| 5 | HRVTri | $L_m$ in LH |
| 6 | $L_{dm}$ in ROI | SDNN |
| 7 | Mobility | PSD in HF |
| 8 | SDNN | $L_m$ in ROI |
| 9 | $L_m$ in LL | $E_{snb}$ in HH |
| 10 | $L_{dm}$ in HH | $L_m$ in HH |

## 5.8    Conclusions

ICD devices are limited in their computational capability; therefore, it is important to be able to efficiently classify ventricular tachyarrhythmia events during sudden cardiac death episode. HRV analysis is a computationally efficient method to classify VT-VF events on an embedded device. Reducing the number of features required as well as the buffering length of HRV signals would aid to further decrease the overhead on the device. The study presented in this chapter shows the most effective HRV features were derived using the mRMR ranking method, combined with SVM, k-NN or RF as classifier, leading to sensitivity, specificity and accuracy greater than 86%, with accuracy as high as 91.5% and specificity as high as 94.2%. Other feature ranking method were considered [189], however, mRMR and ILFS gave the best performance. The results indicate that a shorter time window closer to the event may be possible. The study did not use signals from the same patient in learning and evaluation phases. Future work could address the optimal time window length (which may even be dynamic) informed by the clinical understanding of the heart and heart conditions.

The next chapter extends the work on feature selection to the consideration of prediction of paroxysmal atrial fibrillation (PAF) using features derived from HRV signals.

# Atrial Fibrillation Prediction using Heart Rate Variability Analysis

Paroxysmal atrial fibrillation (PAF for short) is a cardiac arrhythmia that can eventually lead to heart failure or stroke if left untreated. Early detection of PAF is therefore crucial to prevent any further complications and avoid fatalities. As for ventricular arrhythmias covered earlier, an implantable defibrillator device could be used to both detect and treat the condition. With the constraint of limited computational capability in mind, this chapter presents a set of novel features to accurately predict the presence of PAF. The proposed method in this chapter is evaluated using ECG signals from the widely used atrial fibrillation prediction database (AFPDB) from PhysioNet. 106 signals are examined from 53 pairs of ECG recordings. Each pair of signals contains one 5-minute ECG segment that ends just before the onset of a PAF event and another 5-minute ECG segment at least 45 minutes distant from the PAF event to represent a non-PAF event. The data are pre-processed to remove noise and interference. Seven novel features are extracted through the Poincaré representation of R-R interval signals, and are prioritised through the mRMR and ILFS feature ranking schemes. To determine HRV feature prediction efficacy, the presented features are applied to four standard classification techniques and their accuracies, sensitivities and specificities are estimated and compared to the state of the art from the literature. The work described in this chapter has been published in Parsi *et al.*, "Prediction of Paroxysmal Atrial Fibrillation using New Heart Rate Variability Features," *Comput. Biol. Med.*, p. 104367, Apr. 2021 [190].

## 6.1 Literature Review

Atrial Fibrillation (AF for short) is one of the most common cardiac arrhythmias affecting adults of any age [62]. AF occurs when the heart beats in a disorganized and irregular way, and if persistent and left untreated, can lead to various heart-related complications [191]. Although it is not immediately life-threatening in the same way as some other arrhythmias, AF can lead to heart failure or stroke and thromboembolic events which increase overall mortality [192]. One in four people over 50 are at risk of AF which may severely impact on the quality of their life [193]. Approximately 0.7 million (13%) of the ≈5.3 million cases of AF in the United States are undiagnosed [194]. In the United States approximately 15% of stroke patients present with AF, a figure that is projected to double by 2030 [194]. AF can be divided into three categories: paroxysmal atrial fibrillation (PAF), persistent atrial fibrillation and chronic atrial fibrillation. PAF presents as short duration episodes of AF, lasting from several minutes to days and is self-terminating. Persistent AF occurs similarly to PAF, but it cannot self-terminate without external treatment such as medication or electrical shock. Finally, chronic atrial fibrillation has the most significant effects on the body, lasting more than 7 days, and can prevent the heart rhythm returning to normal behaviour even with external treatment. AF patients often start with episodes of PAF before their condition escalates to a chronic stage. Furthermore, about 18% of PAF evolves to permanent AF (persistent or chronic) over 4 years [195].

AF can be treated by medication or electrical shock issued by an ICD [196]. Therefore, having an accurate predictor of the onset of PAF is clinically important because it increases the possibility to prevent the onset of atrial arrhythmias either electrically or using pharmacological treatments, and can enable more efficient and cost-effective screening protocols [197]. An accurate predictor would allow for a time-efficient and cost-effective screening procedure during a clinical visit which may lead to decrease the risk of strokes and thromboembolic events [198].

For many years, researchers have attempted to predict PAF using ECG signals [199]. PAF prediction work can be categorised into:

1. Premature atrial complex (PAC) detection [157], [200]–[203].
2. HRV analysis [86], [87], [102], [201], [204]–[209].

### 6.1.1 PAC Detection Methods

PAC occurs in approximately 93% of PAF events [210]. Therefore, several methods proposed in the literature use detection of PAC as a means to predict PAF. The majority of PAC detection methods in the literature analyse a 30-minute ECG signal. The number and timing of PACs in the ECG episodes have studied by Zong *et al.* [200] in 2001 showing increases in the number of PACs in episodes preceding PAF where they mostly happen toward the end of episodes. Using several criteria, Thong *et al.* in [157] developed an algorithm based on a predictor showing that an increase in activity detected by any of these three criteria is an indication of an imminent episode of PAF. The criteria they used were: first, the number of isolated PACs that are not followed by a regular R-R interval; second, presence of runs of atrial bigeminy and trigeminy; and third the length of any short run of paroxysmal atrial tachycardia. A result of 99.6% in sensitivity and 99.4%, was obtained using a convolutional neural network (CNN) by Jalali *et al.* [203]. PAC-based methods based on data obtained from the PhysioNet [138] atrial fibrillation prediction database (AFPDB for short) are summarised in Table 6.1. Furthermore, Bashar *et al.* [211] has used the AFPDB database to test their proposed algorithm focusing on specificity, where they have reached 100% in specificity while only using selected number of patients from the database. They have obtained 124 segments of PACs from 13 patients, to evaluate proposed PAC detection system using tuned support vector machine and random forest along with 9 selected features. In more recent works, it has been shown better that performance (as much as 10%) could be achieved using deep learning algorithm such as long short-term memory (LSTM) and recurrent neural network (RNN) as compared to traditional machine learning classifiers, such as support vector machines, logistic regression, etc using ECG signals [212]–[214].

From a clinical standpoint, many attempts have been made to scrutinize the role of the P-wave in the prediction of PAF. Dilvaries *et al.* [215] have shown significant difference in the maximum and minimum P-wave duration from ECGs of PAF and healthy samples in patients. With maximum duration of 110 milliseconds and minimum dispersion value of 40 milliseconds, they achieved a sensitivity of 88% and 83% and a specificity of 75% and 85%, respectively. In a similar effort, other studies also showing significant changes in amplitude, duration, and the dispersion of P-wave in ECG signals from PAF in comparison with normal ECG samples from the same patients [216], [217].

**Table 6.1.** Performance comparison between the previous works using PAC analysis on AFPDB.

| Previous Works | Method, Classifier and Signal Length | Validation Method | SN (%) | SP (%) |
|---|---|---|---|---|
| Zong *et al.*, 2001 [200] | Number and timing of PACs, 30 minutes segment | Single hold | 79 | - |
| Hickey *et al.*, 2002 [201] | PACs analysis and spectral based HRV features along with proposed classier, 30 minutes segment | 5-fold CV | 79 | 72 |
| Thong *et al.*, 2004 [157] | PAC analysis with proposed three criteria classification method, 30 minutes segment | Single hold | 89 | 91 |
| Erdenebayar *et al.*, 2019 [202] | CNN, 30 seconds segment | Single hold | 98.7 | 98.6 |
| Jalali *et al.*, 2020 [203] | PACs analysis and resampling with CNN, 30 minutes segment | 3-fold CV | 99.6 | 99.4 |
| Bashar *et al.*, 2021 [211] | PACs detection using SVM and RF along with 9 selected features | Single hold | - | 100 |

SN = sensitivity; SP = specificity; CV = cross-validation

## 6.1.2 HRV Analysis Methods

Several studies have used HRV analysis to directly detect a PAF event. This includes different HRV metrics such as time, frequency, bispectrum and nonlinear feature extracted from 5, 10, 15 and/or even 30 minutes HRV signal to predict PAF as shown in Table 6.2 Table 6.2. From all proposed HRV metrics, spectral analysis of the HRV signal is believed to be of great use in predicting PAF. This is mainly because of the increased power in LF and HF bands of HRV immediately before the PAF event [198]. Furthermore, $SD_1$ and $SD_2$ as a quantifier of Poincaré plot spread provide valuable insight into the dynamics of HRV signals, showing lower values in normal episodes compared to PAF episodes [86]. Complexity analysis of the HRV signal has also demonstrated its usefulness in estimating regularity of the signal. Studies have shown a significant decrease in the complexity of R-R intervals preceding PAF [206].

In contrast to PAC analysis, HRV methods *directly* detect a PAF episode, require less computational power and can be implemented on an implantable device like ICDs or pacemakers [102], [196], [218]. A PAF predictor could control the ICD's ATP treatment method possibly allowing it to immediately restore normal sinus rhythm once an arrhythmia is detected. Generally, the ICD device is expected to operate for more than 5 years after it is implanted in the human body [58]. To maintain the device lifespan, on-board computation must be minimised, which, in the context of a PAF prediction method means there must be a reduction in the:

1. Length of the signals analysed (reduce storage).
2. The number of features calculated.
3. The complexity of the detection method.

In order to contribute to the goal of reduced computational load of PAF detection, this chapter presents a set of novel HRV features for the detection of a PAF event. These features are extracted from a Poincaré representation of 5-minute segments of R-R interval signals extracted from patient ECG data. The features are compared to the state-of-the-art features within the literature and analysed in terms of importance using feature ranking schemes. A 10-fold cross-validation method is used to demonstrate the PAF prediction performance with different features sets. The novel features are tested by themselves, and when combined with existing features from the literature.

**Table 6.2.** Performance comparison between the previous works using PAC analysis on AFPDB.

| Previous Works | Method, Classifier and Signal Length | Validation Method | SN (%) | SP (%) |
|---|---|---|---|---|
| Lynn *et al.*, 2001 [204] | HRV based return and difference map proposed features with k-NN classifier, 30 minutes segment | Single hold | 57 | - |
| Yang *et al.*, 2001 [205] | HRV based proposed method: Footprint analysis, 10 minutes segment | Single hold | 64 | - |
| Hickey *et al.*, 2002 [201] | Spectral based HRV features along with proposed classier, 10 minutes segment | 5-fold CV | 53 | 80 |
| Chesnokov *et al.*, 2008 [206] | HRV based spectral and complexity analysis with neural network classifier, 30 minutes segment | Single hold | 72.7 | 88.2 |
| Mohebbi *et al.*, 2012 [86] | Time, frequency and nonlinear HRV features (12 metrics) with SVM classifier, 30 minutes segment | Single hold | 96.3 | 93.1 |
| Costin *et al.*, 2013 [207] | HRV features and morphological variability of QRS with proposed classifier, 30 minutes segment | Single hold | 89.3 | 89.4 |
| Boon *et al.*, 2016 [102] | Bispectrum and nonlinear HRV features (9 selected metrics) with SVM classifier, 15 minutes segment | 10-fold CV | 77.4 | 81.1 |
| Boon *et al.*, 2018 [87] | Time, frequency, bispectrum and nonlinear HRV features (7 selected metrics) with SVM classifier, 5 minutes segment | 10-fold CV | 86.8 | 88.7 |
| Ebrahimzadeh *et al.*, 2018 [209] | Time, frequency and nonlinear HRV features (12 selected metrics) using mixture of experts classifiers, 5-minute segment | 10-fold CV | 100 | 95.5 |

k-NN = k-nearest neighbour; VM = validation method; SN = sensitivity; SP = specificity; SVM = support vector machine; CV = cross validation

## 6.2 Proposed HRV Features

As covered in previous chapters, there are many classic HRV measures in time, frequency, bispectrum and nonlinear domain which can capture both the sympathetic and the parasympathetic components of the autonomic nervous system [36] and have been used for arrhythmia detection including PAF [86], [102], [208]. Here, additional features derived from the Poincaré plot representation are added. Firstly, a more detailed explanation of the Poincaré plot representation than that given in Chapter 3 is presented.

### 6.2.1 Poincaré Plot Representation

The Poincaré plot is a visual representation of R-R intervals [204], [219]–[222] and is constructed as follows. Let us denote the R-R interval signal by: $RR_1, RR_2, RR_3, \ldots, RR_t$, where $RR_i$ represent each R-R interval in millisecond and $t$ is the number of R-R intervals in the signal under analysis. The return map is a plot of $(RR_i, RR_{i+1}), i \in \{1, 2, \ldots, t-1\}$ i.e. a plot of the points: $(RR_1, RR_2), (RR_2, RR_3), \ldots, (RR_{t-1}, RR_t)$. Two standard deviations can be derived, namely $SD_1$ and $SD_2$, which if the return map is considered as an ellipse, capture the minor and major semi-axes (Figure 6.1 (a)). $SD_1$ and $SD_2$ are defined as:

$$SD_1 = \sqrt{\frac{1}{2}SDSD^2} \qquad (6.1)$$

$$SD_2 = \sqrt{2SDRR^2 - \frac{1}{2}SDSD^2} \qquad (6.2)$$

where $SDSD$ is the standard deviation ($SD$) of differences between adjacent R-R intervals, and $SDRR$ is the standard deviation of all R-R intervals in the whole signal:

$$SDSD = SD(RR_i - RR_{i+1}), i \in \{1, 2, \ldots, t-1\} \qquad (6.3)$$

$$SDRR = SD(RR_i), i \in \{1, 2, \ldots, t\} \qquad (6.4)$$

The width ($SD_1$) of this ellipse, which is related to the fast beat-to-beat variability in the HRV, and the length ($SD_2$) of the ellipse related to the longer-term variability of that data [115] are not the only features from return map that have been used before. Recent works also have proposed different angle of points or even number of clusters in the Poincaré plot as effective features in PAF prediction applications [220], [223].

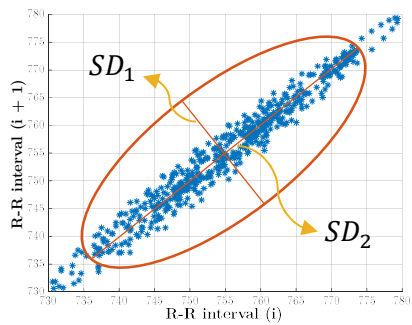

(a) Return Map - Normal Sample from Patient No.3

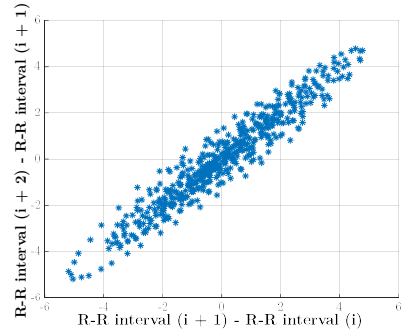

(b) Difference Map - Normal Sample from Patient No.3

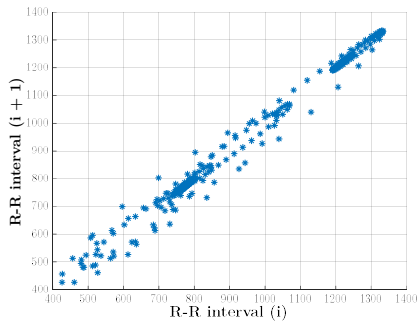

(c) Return Map - PAF Sample from Patient No.3

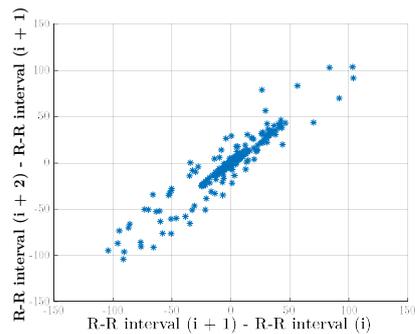

(d) Difference Map - PAF Sample from Patient No.3

**Figure 6.1.** The return map (left) and the difference map (right) of R-R intervals for a normal ECG signal (top row) and for onset of PAF event (bottom) from the AFPDB (patient No. 3). $SD_1$ and $SD_2$ represents the dispersion along minor and major axis of the fitted ellipse.

## 6.2.2    Difference Map

Another Poincaré-based representation for PAF prediction is the difference map of R-R intervals. Difference maps are constructed from the difference between consecutive R-R intervals ($RR_{n+1} - RR_n$, $RR_{n+2} - RR_{n+1}$). Figure 6.1 (b) and (d) shows the difference map plot of a healthy and onset of PAF R-R interval signals, respectively. If the return map is considered to capture the velocity of R-R intervals, the difference map is a visual representation of the rate-of-change of the velocity, i.e., the acceleration.

In both Figure 6.1 (a) and (b) which present return and difference map of normal cardiac signal respectively, changes are more gradual. In contrast, in Figure 6.1 (c) and (d), changes are more sudden and chaotic showing disordered cardiac activity which may lead to atrial fibrillation. By showing the magnitude of velocity in the return map, we can study changes in the heart rate which is a relevant factor to monitor in the event of cardiac arrhythmias such as atrial fibrillation. Furthermore, the difference map allows us to monitor the acceleration. This could help to monitor anomalies in the sinus node that control the heartbeat rate.

The first proposed feature metric is the covariance of $X$ and $Y$ axes of the difference map, calculated as follows:

$$Cov(X, Y) = \frac{1}{n} \sum_{i=1}^{n} (X_i - \mu_x)(Y_i - \mu_y) \tag{6.5}$$

where $X_i = RR_{i+1} - RR_i$ and $Y_i = RR_{i+2} - RR_{i+1}$ and $\mu$ represent the mean of each axis. Slope and angle differences between normal and PAF difference maps which show the changes in heartbeat acceleration can be represented using the change in covariance value over time.

## 6.2.3    Kernel Density Estimation

A bivariant kernel density estimation (KDE) can also be applied to difference map. KDE is a nonparametric density estimator in statistical data analysis [224]. To calculate the final bivariant KDE of difference map, first consider a one-dimensional distribution function of each axis, where $RR_{n+1} - RR_n$ is the x-axis and $RR_{n+2} - RR_{n+1}$ is the y-axis. Taking the x-axis first, let $X_i = \{x_1, x_2, \dots, x_n\} \in \mathbb{R}^d$ be a random sample from distribution with unknown univariate probability density $f(x)$. The standard kernel density estimator for $f(x)$ is calculated as follows [225]:

$$\hat{f}(x) = \frac{1}{nh} \sum_{i=1}^{n} k(\frac{x - X_i}{h}) \qquad (6.6)$$

where $n$ is the number of observations, $h$ is positive number called smoothing parameter and $k(x)$ is the kernel function satisfying the following conditions:

$$0 \leq k(x) < \infty \; for \; all \; x, and \; \int_{-\infty}^{\infty} k(x)dx = 1 \qquad (6.7)$$

where the most common kernel function is a Gaussian:

$$k(x, X_i) = (2\pi\sigma^2)^{-d/2} exp\left\{-\frac{\|x - X_i\|^2}{2\sigma^2}\right\} \qquad (6.8)$$

After calculating KDE for each dimension, the univariate form can be simply extended to bivariate or even multivariant form. In the most popular form, the standard bivariate kernel density estimator is written as follows [226]:

$$\hat{f}(x, y) = \frac{1}{nh_x h_y} \sum_{i=1}^{n} k(\frac{x - X_i}{h_x}) k(\frac{x - X_i}{h_y}) \qquad (6.9)$$

Figure 6.2 shows univariate and Figure 6.3 shows bivariate KDE from difference map for both normal R-R intervals, and R-R intervals at onset of PAF event for patient No. 3. There are some similarities between the univariate KDEs for x- and y-axis for each condition (normal and PAF), as can be seen Figure 6.2. There are also visible differences in the bivariate plots Figure 6.3 (a) and Figure 6.3 (b) as well. Six novel features are calculated from the KDEs to better highlight these differences between normal and PAF distributions.

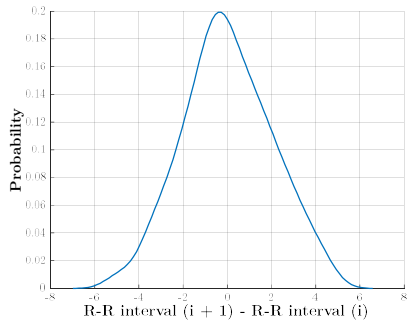

(a) X-axis Univariant KDE Plot from Difference Map - Normal Sample from Patient No.3

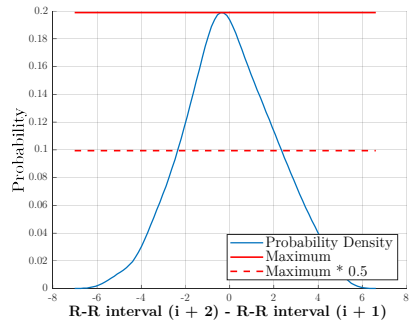

(b) X-axis Univariant KDE Plot from Difference Map - Normal Sample from Patient No.3

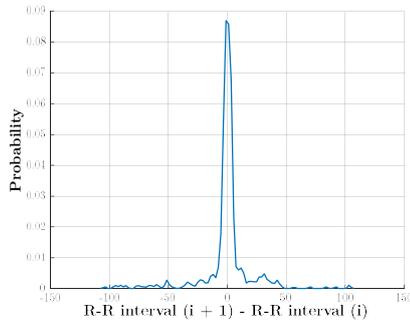

(c) Y-axis Univariant KDE Plot from Difference Map - PAF Sample from Patient No.3

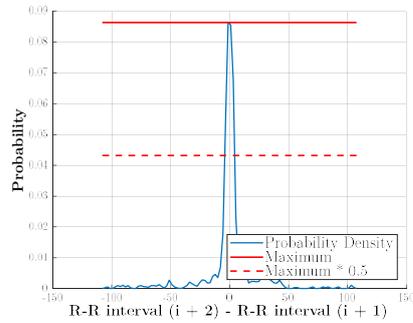

(d) X-axis Univariant KDE Plot from Difference Map - PAF Sample from Patient No.3

**Figure 6.2.** The difference map probability density of one axis of the patient No. 3. The top row shows the normal event and bottom shows the onset of PAF for this patient.

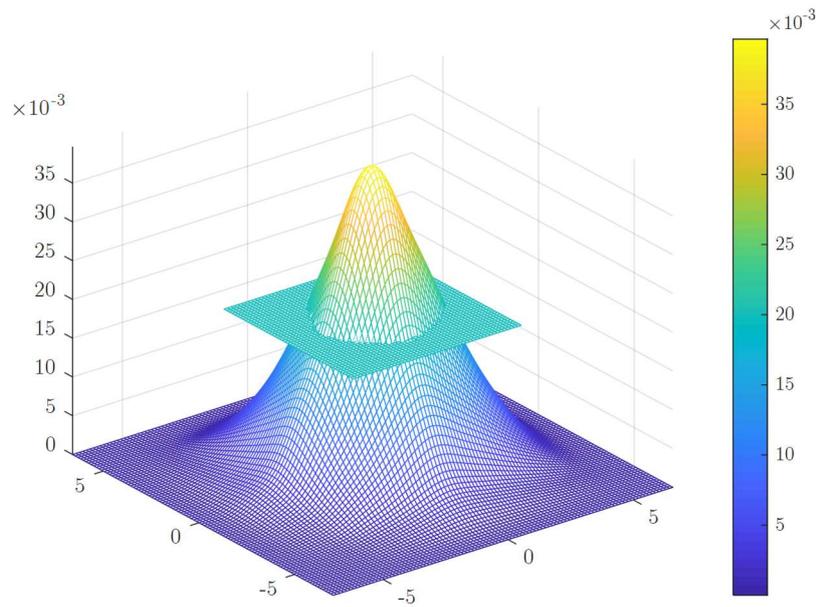

(a) XY- axis Bivariate KDE Mesh Plot – Normal Sample from Patient No. 3

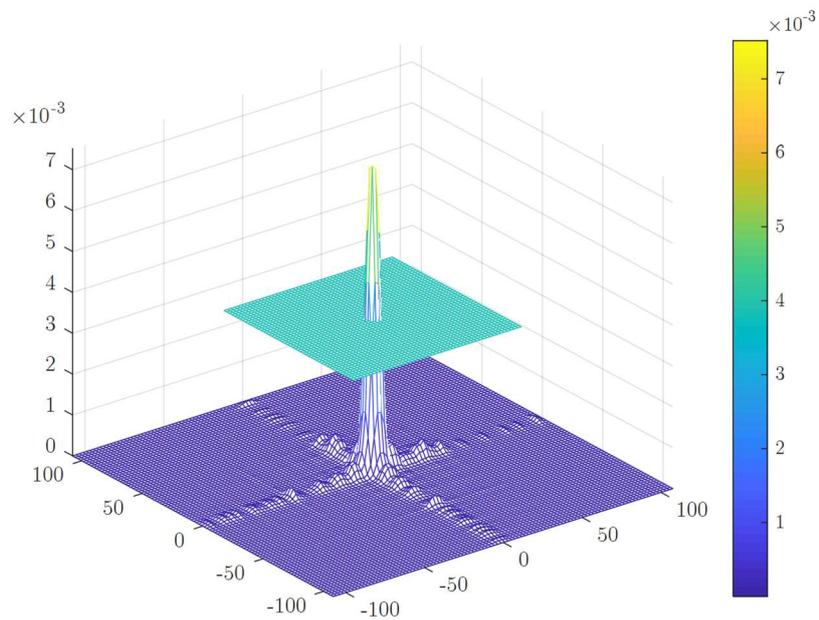

(b) XY- axis Bivariate KDE Mesh Plot – PAF Sample from Patient No. 3

**Figure 6.3.** The difference map KDE mesh plot of the patient No. 3. The top row shows the normal event and bottom shows the onset of PAF for this patient.

### 6.2.4 Univariate KDE Features

There are two features extracted from one axis of a univariate KDE. One axis is sufficient as the y-axis is only one value shift from x-axis their univariate plots are similar. The first feature is the area between the peak and half peak values of the univariate KDE, indicated by the solid red line and dashed red line respectively on Figure 6.2 (b) and Figure 6.2 (d). Then the area between is calculated as follows:

$$Area(y) = \sum_i \hat{f}(y)_i \,, where \; \hat{f}(y)_i > \max\left(\hat{f}(y)\right) * 0.5 \qquad (6.10)$$

where $\hat{f}(y)$ is the univariate KDE function of y-axis and $i$ counts all probability values, which are higher than the half of the $\hat{f}(y)$ maximum probability.

As a second feature, the energy of the peak to half peak of the KDE is calculated as follows:

$$Energy(y) = \sqrt{\sum_i \hat{f}(y)_i^2} \,, where \; \hat{f}(y)_i > \max\left(\hat{f}(y)\right) * 0.5 \qquad (6.11)$$

### 6.2.5 Bivariate KDE Features

Four features are extracted from the bivariant KDE. The first two features are the minimum and maximum spans of the bottom plan ($XY$) in the 3-D KDE plot which we call $SurfMax$ and $SurfMin$ repctivly. As an example, in Figure 6.3 plot, $SurfMax$ is about 5.5 and $SurfMin$ is -5.5. The other two features are similar to the area and energy features univariate derived from the univariate distribution. In this case, we define a half-maximum plane in the distribution, and calculate the volume and energy of the area between that plane and the peak value:

$$Volume(x, y) = \sum_i \hat{f}(x, y)_i \qquad (6.12)$$

$$Energy(x, y) = \sqrt{\sum_i \hat{f}(x, y)_i^2} \qquad (6.13)$$

where $\hat{f}(x, y)_i > \max\left(\hat{f}(x, y)\right) * 0.5$. The seven proposed features are summarised in Table 6.3.

**Table 6.3.** Proposed features extracted from difference map.

| Features Category | Features Name | Descriptions |
|---|---|---|
| **Difference Map Covariance** (1 Metric) | $Cov(X,Y)$ | The covariance of $X$ and $Y$ axes of the difference map. |
| **Univariate KDE Features** (2 Metrics) | $Area(y)$ $Energy(y)$ | The area and energy of the peak to half peak in univariate KDE. |
| **Bivariate KDE Features** (4 Metrics) | $SurfMax$ $SurfMin$ | the minimum and maximum spans of the bottom plan ($XY$) in bivariate KDE |
| | $Volume(x,y)$ $Energy(x,y)$ | The volume and energy of the area between that half-max plane and the peak value in bivariate KDE. |

## 6.3 Atrial Fibrillation Prediction Database

The ECG data that have been used for this study were taken from the PhysioNet AFPDB database [138]. This is an annotated database and consists of 3 types of record sets. The first record set, starting with the letter "n", comes from 50 subjects who did not experience atrial fibrillation at all. This set is usually used as "normal" ECG signal for tuning detectors [157], [200], [204], [207]. The second record set was taken from 25 subjects staring with letter "p". Each subject in this set has 30 minutes of ECG signal (odd number) during a period that is distant from PAF, labelled as a normal ECG signal, and 30 minutes of ECG signal (even number) immediately precedes a PAF episode. The third record set contains 100 annotated 30 minutes ECG signal recordings from 50 subjects. In this record set, there are subjects with signals that are all normal, in PAF onset, or one in each category.

Based on previous works [86], [87], [102], [201], [206], [207], [209], the selection of subjects from the database can vary. Chesnokov [206] only extracted HRV signal from 16 patients. The selection criteria was the availability of at least 60 min of signal length before PAF event to investigate the possibility of long-term prediction of PAF onset. Mohebbi *et al.*, [86] extracted 106 ECG segments immediately prior to an episode of PAF to evaluate their proposed algorithm. In common with the approach take in [102], [209], in this research 106 events from 53 patients have been processed. Each patient contributes a pair of signals which consists of one 30-minute ECG segment that ends just prior to the onset of a PAF event, and another 30-minute ECG segment at least 45 minutes distant from any PAF event, and therefore represents normal heart behaviour. With a number of classifiers, 10-fold cross-validation has been used for evaluation. Each ECG segment contains two-channel traces from Holter recordings with sampling rate of 128 Hz and 12-bit resolution. In this study we consider a 5-minute ECG segment based on proposed signals length used in previous studies [87], [209]. The 5-minute ECG segment occurring at least 45 minutes from the PAF event is assigned a class label of "Normal", while the ECG segment that immediately precedes the PAF event is given a class label of "Abnormal". This method has been chosen as the algorithm is designed for use in implantable devices where the decision is based on each patient's healthy and PAF onset signals.

## 6.4 Proposed Evaluation Method

This Section discusses the methodology used to predict PAF based on 5-minute segments of R-R interval signals derived from ECG patient data. An overview of the entire method is given in Figure 6.4. As it mentioned in the section before, the data is obtained from the PhysioNet atrial fibrillation prediction database [138], [227]. The data is first pre-processed to extract R-R information and reduce noise. Features – both existing features from the literature as well as the newly-proposed features here – are then extracted and finally the system determines whether the patient data is normal or abnormal using a range of classification techniques.

**Figure 6.4.** PAF prediction proposed algorithm block diagram. Proposed features along with other commonly used features are ranked using mRMR and ILFS based on their ability to generate the highest degree of differentiation between the two classes. With different combinations of features, prediction performance is calculated using four different classification algorithms (SVM, k-NN, RF, and MLP neural network), using cross-validation to allow fairer comparison between different results.

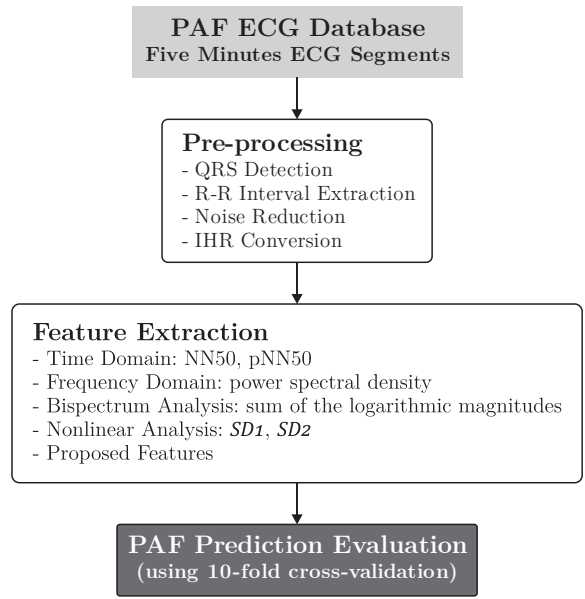

### 6.4.1    Prepossessing and Noise Reduction

HRV signals are extracted from the ECG signals using the annotation files provided with the database [138], [227]; an example is given in Figure 6.5. Although the annotation is accurate in the example in Figure 6.5, these files are unaudited and contain some errors in QRS detection, which are generally unavoidable in such data collection [138]. The presence of these errors presents a more challenging scenario for PAF prediction and will further test the robustness of the proposed algorithm.

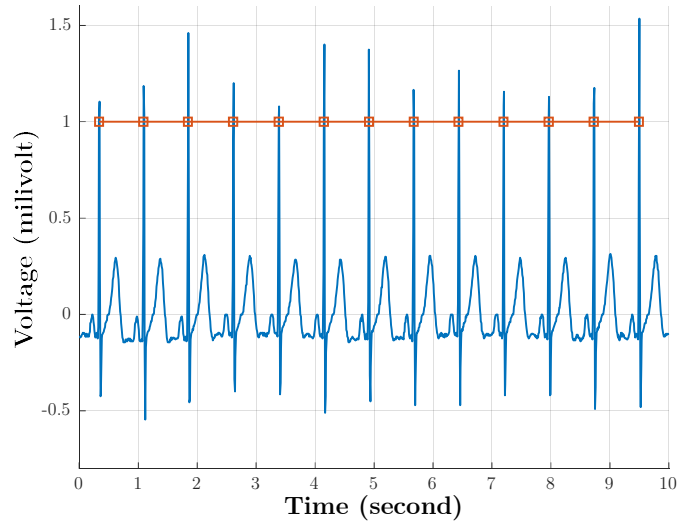

(a)  R Peak Detection on ECG Signal - Normal Sample from Patient No.3

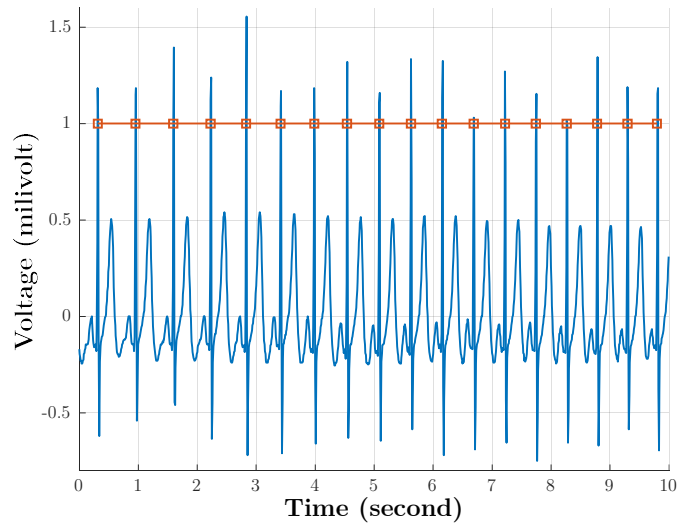

(b)  R Peak Detection on ECG Signal - PAF Sample from Patient No.3

**Figure 6.5.** R peak detection example on ECG signals from patients No. 3.

After using the annotations to find the R peaks, the R-R interval signals have been calculated and then resampled to 7 Hz by using the cubic spline interpolation. Then the *IHR* signal is calculated from the R-R interval signals using the same equation as in previous chapters:

$$IHR(beats/min) = \frac{60,000}{RRI\ (msec)} \tag{6.14}$$

where *RRI* (R-R intervals) is the time in milliseconds between instantaneous heartbeats.

A two-step noise reduction method has been implemented to remove unwanted signal noise caused by e.g. muscular activity [171]. In the first step, to deal with spikes which present as noise or as ectopic beats, the signal has been corrected by McNames's algorithm [228]. To apply the algorithm, the following statistic is first calculated on the *IHR* signals:

$$D(n) = \frac{|IHR(n) - IHR_m|}{1.483\,med\{|IHR(n) - IHR_m|\}} \tag{6.15}$$

where $med\{...\}$ is the median operator, $IHR(n)$ is instantaneous heart rate for beat $n$, $IHR_m$ is the median value of the heart rate over a given window. The filtered instantaneous heart rate, $\widehat{IHR}(n)$, is then calculated as a follow:

$$\widehat{IHR}(n) = \begin{cases} IHR(n) & D(n) < \tau \\ med\left\{IHR(n+m) : |m| < \dfrac{w_m - 1}{2}\right\} & D(n) \geq \tau \end{cases} \tag{6.16}$$

where $\tau$ is application specific threshold value. If $D(n)$ exceeds this threshold, the instantaneous heart rate is considered as abnormal heart rate otherwise it considered as a normal value $IHR(n)$. In the abnormal case the heart rate is corrected as shown in equation (16), where $w_m$ is the window length of the median filter. In this work, $\tau$ and $w_m$ are set to 4 and 11 respectively as suggested in [228]. The second step of the noise reduction scheme is based on the wavelet transform adapted to remove any artefact beats, similar to [4]. Using high and low pass filters, the wavelet transform can remove the higher frequencies of the background noise from the signals. In this research, the first approximation level of wavelet transform is taken, and the sym8 wavelet is used [173]. The pre-processed signal is now ready for feature extraction.

### 6.4.2 Feature Extraction

In addition to the seven proposed features described above, other features are also extracted from the pre-processed signals. These features have been used in previous PAF prediction studies [86], [87], [102], [207]–[209]. The features have been selected from the same four categories used in previous chapters in this thesis: time domain, frequency domain, bispectrum and nonlinear analysis, and they are summarized here in Table 6.4 [77]. The time domain features used are MeanNN, SDNN, NN50, and pNN50 [28], [45], [81]–[85], [137]. As noted in Chapter 3, time domain methods have low computational cost but cannot easily discriminate between the sympathetic (at low-frequency range (0.04-0.15 Hz)) and the parasympathetic contributions (at higher frequency range (0.15-0.4 Hz)) in the HRV signals [229]. This distinction facilitates preventive intervention at an early stage when it is most beneficial [85]. In the frequency domain category, four features based on average power of HRV signals across the very low frequency (VLF) band (0.0-0.04 Hz), low frequency (LF) band (0.04-0.15 Hz), high frequency (HF) band (0.15-0.4 Hz), and ratios like LF/HF have been computed by integrating the power spectral density (PSD) estimated using Welch's method in each frequency range of interest [26], [84], [86]–[89].

HOS features up to the third-order cumulant were used to estimate the bispectrum from HRV data in different HRV studies [84], [86], [87], [92], [102]. As described in Chapter 3, the magnitude average ($M_{avg}$), power average ($P_{avg}$), and logarithmic bispectrum features, are the most commonly used features from the bispectrum which have been extracted from the ROI; they are summarized in Table 6.4 [77], [78].

By considering a HRV signal as an indirect representation of permanent interplay between the two branches of the ANS, the signal can be decomposed using nonlinear and dynamic analysis [109], [111]. On top of Poincaré plot features, $SD_1$, $SD_2$, and $SD_1/SD_2$, entropy features as a measure of signal complexity are also used. Sample entropy [118], Rényi entropy, Tsallis entropy [121] have been extracted to highlight the complexity of HRV and aim to capture any regularity changes during the onset of a VT-VF event.

Adding these 29 features from previous studies to the 7 newly-proposed features makes an initial feature set consisting of 36 features. Feature ranking and analysis is then used in order to prioritise features; this is described in more detail in the next section.

**Table 6.4.** Existing features used for PAF prediction, along with their descriptions [77], [78].

| Features Category | Features Name | Descriptions |
|---|---|---|
| **Time Domain** (2 Metrics) | MeanNN | Mean of normal-to-normal intervals or instantaneous heartbeats |
| | SDNN | Standard deviation |
| | NN50 | Number of adjacent R-R intervals differing by more than 50 ms |
| | pNN50 | The number obtained by dividing NN50 by the total number of N-N intervals |
| **Frequency Domain** (4 Metric) | PSD | Power spectral density (PSD) in **LL**, **LH**, **HH** and **ROI** |
| **Bispectrum Analysis** (2 Metrics) | $M_{avg}$ | Magnitude average of bispectrum in LL, LH, HH and ROI: $M_{avg} = \frac{1}{N}\sum_{\Gamma}|B(f_1, f_2)|$ |
| | $P_{avg}$ | Power average of bispectrum in LL, LH, HH and ROI: $P_{avg} = \frac{1}{N}\sum_{\Gamma}|B(f_1, f_2)|^2$ |
| | $L_m$ | Sum of the logarithmic magnitude of the bispectrum in LL, LH, HH and ROI: $L_m = \sum_{\Gamma}\log(|B(f_1, f_2)|)$ |
| | $L_{dm}$ | Sum of the logarithmic magnitudes of the diagonal elements of the bispectrum in **LL**, **HH**, and **ROI**: $L_{dm} = \sum_{D}\log(|B(f_1, f_2)|)$ |
| **Nonlinear Analysis** (2 Metrics) | $SD_1$ $SD_2$ $SD_1/SD_2$ | The width ($SD_1$) and the length ($SD_2$) of the of generated ellipse in Poincaré plot (Figure 6.1) |
| | Entropy | Sample entropy, Rényi entropy, and Tsallis entropy |

$B(f_1, f_2)$ = bispectrum of HRV signal; LF = low frequency band (0.04-0.15 Hz); HF = high frequency band (0.15-0.4 Hz); LL = low frequency sub band region (LF-LF); HH = high frequency sub band region (HF-HF)

### 6.4.3 PAF Prediction

In this work four different classifiers have been applied to predict onset of PAF episodes by classifying each sample in one of two classes, either normal or abnormal, based on extracted features presented as a feature vector. A k-fold cross-validation technique is used to split the data evenly and record performance metrics for $k = 10$ folds in the 53-patient database (106 Normal and Abnormal events). 10 folds equates to having 48 patients in the training set and 5 patients in the test set. The same patient is never present in both the training and test set.

The four classifiers are: support vector machine (SVM) [86], [87], [102], k-nearest neighbours (k-NN) [77], [78], [204], random forest (RF) [230] and multilayer perceptron (MLP) [206], [209]. SVM has proven to be effective in previous studies on HRV classification [175], [222]. The HRV pre-processing, feature extraction, optimization, and classification have been implemented and developed in MATLAB R2018b, and feature analysis and selection used the MATLAB Statistics Toolbox along with the feature selection toolbox in [189]. For the SVM classifier, using Gaussian function as a kernel, we obtained the same results. In this chapter we adopt a linear SVM to reduce the computational load and reduce the risk of overfitting. The kernel scale was set to 1 and box constraint set to 1.6 [78]. Optimization of k-NN was carried out using different numbers of neighbours (1 to 50), along with different distance metrics such as Chebyshev distance, cosine distance, Minkowski distance along with Euclidean distance. For k-NN the best number of nearest neighbours was 5 and the Chebyshev distance has been applied in this study, based on optimization studies [77], [137], [204]. The method of random forests (RF for short) comprises an ensemble of decision trees, and has previously been used to distinguish various cardiac arrhythmias [137], [186], [231]. RF uses 64 trees with leaf size equal to 4 in this study [78], [230]. Finally, the MLP has 5 hidden layers with 32, 32, 16, 16 and 4 nodes on each layer, respectively. The maximum epochs are set to 4096 using mean squared error with regularization function and implemented in MATLAB R2018b.

## 6.5    Experimental Results and Discussion

In this section seven proposed HRV features are tested along with existing features from previous studies to evaluate PAF prediction using 5-minute R-R interval signals.

### 6.5.1    Results of Feature Ranking

Feature selection and ranking is used to firstly, make the model easier to interpret by removing the variables that are redundant; secondly, to reduce the size of the problem which enables algorithms to work faster; and thirdly, to guard against model overfitting. A range of methods for feature selection were evaluated in this research. An additional feature analysis method was evaluated in this particular study on PAF prediction. LASSO (least absolute shrinkage and selection operator) was first formulated by Tibshirani [232]. It uses an $L_1$ norm and tends to force individual coefficient values completely towards zero. It performs two main tasks: regularization and feature selection. LASSO puts a constraint on the sum of the absolute values of the model parameters where the sum has to be less than a fixed value which is called the upper bound. To do that LASSO applies a shrinking (regularization) process where it penalizes the coefficients of the regression variables, shrinking some of them to zero. The variables that still have a non-zero coefficient after the shrinking process are selected. As indicated in Table 6.5, twenty-two features have been selected with LASSO where all the newly proposed features are included. LASSO feature analysis and selection helps to deal with multicollinearity and redundant predictors by quickly identifying the key variables [233]. This set of features provides a useful baseline for further prioritization.

To further analyse the original set of features, two other feature selection and ranking methods are used. The minimal redundancy-maximal Relevance (mRMR) and the infinite latent feature selection (ILFS), previously applied in Chapter 5, are also used here. Table 6.5 summarises the results of ranking of the top 14 features using both approaches. These 14 features selected by mRMR and ILFS are a subset of the features identified by LASSO. It was found in experimental evaluation that incrementally adding more features does not improve the cross-validation result. Based on the results of feature selection and ranking, at least five out of the seven proposed features are included in the top 10 rankings using both ranking methods. Both methods include one of the proposed features at the highest rank.

**Table 6.5.** The Lasso, mRMR, and ILFS features ranking and selection processing on 5 minutes signals from AFPDB database. Shaded rows are the ones which are selected using each methodology and numbers under mRMR and ILFS columns represent feature ranking.

| Features Category | Features Name | Lasso | mRMR | ILFS |
|---|---|---|---|---|
| **Time Domain** (4 Metrics) | MeanNN | 0 | 0 | 0 |
| | SDNN | 1 | 0 | 0 |
| | NN50 | 1 | 4 | 14 |
| | pNN50 | 1 | 2 | 2 |
| **Frequency Domain** (4 Metric) | PSD in VLF | 0 | 0 | 0 |
| | PSD in LF | 1 | 0 | 0 |
| | PSD in HF | 0 | 0 | 0 |
| | PSD in LF/HF | 1 | 1 | 3 |
| **Bispectrum Analysis** (15 Metrics) | $M_{avg}$ in LL | 0 | 0 | 0 |
| | $M_{avg}$ in LH | 1 | 0 | 0 |
| | $M_{avg}$ in HH | 0 | 0 | 0 |
| | $M_{avg}$ in ROI | 0 | 0 | 0 |
| | $P_{avg}$ in LL | 0 | 0 | 0 |
| | $P_{avg}$ in LH | 0 | 0 | 0 |
| | $P_{avg}$ in HH | 0 | 0 | 0 |
| | $P_{avg}$ in ROI | 0 | 0 | 0 |
| | $L_m$ in LL | 1 | 0 | 0 |
| | $L_m$ in LH | 1 | 0 | 0 |
| | $L_m$ in HH | 1 | 0 | 0 |
| | $L_m$ in ROI | 1 | 0 | 0 |
| | $L_{dm}$ in LL | 1 | 12 | 4 |
| | $L_{dm}$ in HH | 1 | 14 | 11 |
| | $L_{dm}$ in ROI | 1 | 0 | 0 |
| **Nonlinear Analysis** (6 Metrics) | $SD_1$ | 1 | 7 | 5 |
| | $SD_2$ | 1 | 13 | 12 |
| | $SD_1/SD_2$ | 0 | 0 | 0 |
| | SampEn | 0 | 0 | 0 |
| | Rényi Entropy | 0 | 0 | 0 |
| | Tsallis Entropy | 0 | 0 | 0 |
| **Difference Map Covariance** (1 Metric) | $Cov(X,Y)$ | 1 | 11 | 6 |
| **Univariate KDE Features** (2 Metrics) | $Area(y)$ | 1 | 6 | 8 |
| | $Energy(y)$ | 1 | 3 | 1 |
| **Difference Map Covariance** (4 Metrics) | $SurfMin$ | 1 | 9 | 6 |
| | $SurfMax$ | 1 | 8 | 7 |
| | $Volume(x,y)$ | 1 | 10 | 10 |
| | $Energy(x,y)$ | 1 | 5 | 13 |

## 6.5.2    PAF Prediction: Results of Feature Comparison

Results from different studies for PAF prediction have mainly been presented in terms of sensitivity (true positive rate), specificity (true negative rate), and accuracy:

$$Sensitivity = \frac{TP}{TP + FN} \tag{6.17}$$

$$Specificity = \frac{TN}{TN + FP} \tag{6.18}$$

$$Accuracy = \frac{TP + TN}{TP + FP + TN + FN} \tag{6.19}$$

where TP, TN, FP and FN stand for true positive, true negative, false positive and false negative respectively. The results for different feature sets are presented in Table 6.6. The proposed features by themselves yield higher specificity and accuracy results than the classic feature set, though the classical features result in better sensitivity. By combining the proposed and state-of-the-art feature sets, a linear SVM achieves 98.8% in sensitivity and 96.7% in specificity. The combination of proposed and classical features combines the best performance features of each set of features, i.e., better overall sensitivity in the classical features mixed with high specificity provided in the proposed features.

**Table 6.6.** Summaries of experimental results using different feature sets on AFPDB using 10-fold cross-validation method.

| Features Category | Classifier | SN (%) | SP (%) | ACC (%) |
|---|---|---|---|---|
| **Classic Features** | SVM | 97.8 | 88.9 | 93.3 |
| | MLP | 95.6 | 91.1 | 93.3 |
| | **RF** | **98.8** | 87.8 | 93.3 |
| | k-NN | 91.1 | 82.2 | 86.7 |
| **Proposed Features** | **SVM** | **96.7** | **96.7** | **96.7** |
| | MLP | 90 | 96.7 | 93.3 |
| | RF | 93.3 | 96.7 | 95 |
| | k-NN | 90 | 83.3 | 86.7 |
| **Combine** | **SVM** | **98.8** | **96.7** | **97.7** |
| | MLP | 96.7 | 97.8 | 97.2 |
| | RF | 97.8 | 91.1 | 95 |
| | k-NN | 87.8 | 95.6 | 91.7 |

SN = sensitivity; SP = specificity; ACC = accuracy

### 6.5.3    Comparison with Previous Work

The highest result using HRV analysis on the AFPDB, 100% in sensitivity and 95.5% in specificity, was obtained in [209]. Although the sensitivity reported in that study cannot be improved upon, better specificity has been achieved in the present study while both methods ended with the same overall accuracy (97.7%). However, the study in [209] is based on mixture of experts (ME) classifier that is a much more complex approach in comparison with the linear SVM that has been used on this study. Other classifiers also have been used in [209] as shown in Table 6.7. Based on the reported results in [209], the results from the SVM-based method in this thesis show over 3% improvement, k-NN-based over 2% improvement, and finally MLP-based over 6% improvement all on accuracy. These results suggest that the proposed features have potential for application in implantable device, where computational resources are limited. In particular, the proposed features improve upon specificity compared to previous research.

**Table 6.7.** Comparison of experimental results in the leading literature using different classifiers on 5-minue of AFPDB ECG signals using 10-fold cross-validation method.

| Features Category | Classifier | SN (%) | SP (%) | ACC (%) |
|---|---|---|---|---|
| Boon *et al.*, 2018 [87] using time, frequency, bispectrum and nonlinear HRV features (7 selected metrics) | SVM | 86.8 | 88.7 | 87.7 |
| Ebrahimzadeh *et al.*, 2018 [209] using time, frequency and nonlinear HRV features (12 selected metrics) | SVM | 96.3 | 93.1 | 94.6 |
| | k-NN | 92.3 | 86.7 | 89.3 |
| | MLP | 92.6 | 89.7 | 91.1 |
| | ME | 100 | 95.5 | 98.2 |
| **Proposed Method** using time, frequency, bispectrum and nonlinear HRV features (7 metrics) along with 7 proposed metrics | SVM | 98.8 | 96.7 | 97.7 |
| | k-NN | 87.8 | 95.6 | 91.7 |
| | MLP | 96.7 | 97.8 | 97.2 |
| | RF | 97.8 | 91.1 | 95 |

SN = sensitivity; SP = specificity; ACC = accuracy

## 6.6    Conclusion

Predicting the onset of paroxysmal atrial fibrillation (PAF) can dramatically improve quality of life in cardiac patients and decrease the risk of mortality. Accurate prediction remains a significant challenge, considering the noise and interference present in recorded ECG measurements, as well as the limitations of wearable and implantable devices increasingly used for cardiac rhythm management. In this study, seven novel HRV features were proposed to address the PAF prediction problem. The features were shown to represent the problem space well compared to a selection of popular state-of-the-art features from the literature, using two feature ranking methods. Four well-known classification algorithms were used to compare the classic and proposed feature sets for PAF prediction, using data from the PhysioNet AFPDB. Test and training data were split in a 10-fold cross-validation manner and both sets remained distinct. Using a linear SVM kernel (comparable in complexity between this study and previous research), a combination of the proposed feature set and existing features provides an improvement of over 3% over the leading published results in literature (97.7% vs 94.6%).

This chapter summarises the research objectives and the results of this thesis. The motivation and main findings of this thesis are summarised in Section 7.1 including the main conclusions of this thesis. Related future work to further develop and extend the findings of this thesis are presented in Section 7.3.

## 7.1    Summary and Conclusions

The use of implantable and wearable technology in healthcare and its application potential in cardiology has been a topic of great interest in recent years. In cardiology, HRV plays an important role in this regard. as wearable or implanted devices can use HRV parameters as a means to detect or predict arrhythmias in order to support patient management, including the application of therapeutic interventions where needed. Although there are generally accepted standards for the calculation of HRV features and some agreement concerning the clinical application of these metrics [36], [45], the field of HRV analysis contains many studies reaching a variety of conclusions and there remains an open question on the clinical value of different HRV features. Early detection of major cardiac arrhythmias such as VT and VF could directly help to improve patients' mortality rate and stop sudden cardiac death whereas in AF, prediction and detection could reduce misclassification by implanted devices to improve patients' quality of life. This thesis has conducted an investigation of HRV-based features for the detection of cardiac arrhythmias, in the particular context of resource-constrained implantable devices.

In Chapter 2 of this thesis, the fundamentals of human cardiac activity and its implications for ECG and HRV signals was discussed. It began with the anatomy of the human heart and the most commonly used cardiac signals, ECG and HRV, with their definitions and specifications. Then, cardiac arrhythmias with their definitions and biomarkers were described. Finally, the chapter ended with different cardiac rhythm management systems AED and ICD which are considered to be the main application of prediction algorithms proposed in this thesis.

Chapter 3 presented an overview of HRV analysis, describing the standard and classic metrics for HRV-based detection and prediction algorithms. These measurements and features have been used to develop the automated cardiac arrhythmias classification algorithms for more than three decades. HRV features in this chapter were categorised into time domain, frequency domain, bispectrum and nonlinear analysis features. HRV as a reliable reflection of the many physiological factors shows not only linear structures in the signals but also nonlinear contributions. Since the introduction of CRM devices such as different type of ICDs, many studies show the positive effects of using HRV analysis to reflect significant changes immediately prior to an episode of ventricular arrhythmias.

Chapter 4 presented a review of existing methods based on HRV features for ventricular arrhythmia prediction in ICDs. It highlighted the decision-making algorithms in ICDs to deliver either ATP or electrical shock as a treatment for the patients. Because of low specificity, patients with ICDs still receive inappropriate shocks which may lead to inadvertent mortality and reduction of quality of life. It mainly happens because of misclassification of normal sinus rhythm and SVT from ventricular arrhythmias. This points to the need for more effective features and prediction methods. This chapter also conducted a detailed analysis of features that have been used for VT-VF prediction and then conducts a comparative study of these features for classification of onset in a VT-VF detection task. The results were compared with results from the literature. The study suggests that time domain features can achieve high performance alone, but the addition of more features in other categories can increase performance. It showed by using fifty features together a result of 85.7% in sensitivity, specificity, and accuracy results was achieved, which is 6% higher than previous studies to date [77]. However, further development of feature selection methods was proposed to address the performance of HRV-based sudden cardiac death prediction for use with ICDs with limited resources.

Chapter 5 addressed this challenge by proposing a suitable feature selection method to overcome high computational overhead of previously proposed methods. In this chapter signal buffer lengths where examined, where 1-minute and 5-minute signal durations were processed from the Medtronic version 1.0 dataset. To ensure a robust evaluation of the feature selection technique, the dataset was split into two, non-overlapping subsections. The first subset of data, which is called learning data, was used to rank and select the appropriate features using mRMR and ILFS feature ranking methods. The second subset which is called evaluation data was then used to evaluate models which calculate features based on the chosen features from the first section. All tests were carried out using a leave-patient-out cross-validation method, where the same patient's data cannot exist within the algorithm learning and evaluation data. Results from the SVM, k-NN, and RF classification algorithms were compared. The results showed that the most effective HRV futures were derived using mRMR ranking method, combined with SVM, k-NN or RF as classifier, leading to greater than 86% sensitivity, specificity and accuracy, with accuracy as high as 91.5% and specificity as high as 94.2%. Other feature ranking methods were also considered [189], however, mRMR and ILFS gave the best performance. The results indicate that a shorter time window closer to the event can give good performance, with the added benefit of reduced storage and data management requirements.

Finally, Chapter 6 extended the study to consider the most common cardiac arrhythmia, paroxysmal atrial fibrillation (PAF), which can eventually lead to ventricular arrhythmia and cases heart failure or stroke if left untreated. This chapter proposed a set of seven novel features based on Poincaré analysis to accurately predict the presence of PAF. 106 ECG signals from 53 patients were used in this chapter from the widely used AFPDB in PhysioNet. Each pair of signals contains one 5-minute ECG segment that ends just before the onset of a PAF event and another 5-minute ECG segment at least 45 minutes distant from the PAF event to represent a non-PAF event. The features are shown to represent the problem space well and accurately predict the onset of PAF, using mRMR and ILFS feature ranking methods and four different classifiers. The accuracies of over 97.7%, almost 4% greater than the classical state-of-the-art feature set, were obtained using a linear SVM kernel, along with the proposed feature set.

## 7.2    7.2. Summary of Contributions

The contributions of this thesis are summarized again as follows:

1. A comprehensive comparative study on the performance of HRV features in time domain, frequency domain, bispectrum, and nonlinear analysis in VT-VF prediction. As many as fifty HRV features and their performance in VT-VF prediction system were analysed and a prediction method based on mixture of these feature achieve a better result in compare with the literature. This work was published in **A. Parsi**, D. O'Loughlin, M. Glavin, and E. Jones, "Prediction of Sudden Cardiac Death in Implantable Cardioverter Defibrillators: A Review and Comparative Study of Heart Rate Variability Features," *IEEE Rev. Biomed. Eng.*, vol. 13, pp. 5–16, 2020, and in **A. Parsi**, D. Byrne, M. Glavin, and E. Jones, "Heart Rate Variability Analysis to Predict Onset of Ventricular Tachyarrhythmias in Implantable Cardioverter Defibrillators," *41st Annual International Conference of the IEEE Engineering in Medicine and Biology Society* (EMBC), Jul. 2019, pp. 6770–6775.

2. The application of feature selection techniques to HRV signals to predict VT and VF with a reduced set of features, with performance comparable to using an exhaustive set of features. This work proposed low-complexity pre-processing using a mix of wavelet transform and median filter. The features proposed in this study achieved a more balanced distribution of false positive and false negative results than related methods in the literature. using as few as six features which is the lowest published within this field. This work was published in **A. Parsi**, D. Byrne, M. Glavin, and E. Jones, "Heart Rate Variability Feature Selection Method for Automated Prediction of Sudden Cardiac Death," *Biomed. Signal Process. Control*, vol. 65, p. 102310, Mar. 2021.

3. Seven novel HRV features for onset of PAF detection were proposed. The proposed method in this work provides an improvement of 3% in accuracy over the leading published results in literature using the same classifier. This work was published in **A. Parsi**, M. Glavin, E. Jones, and D. Byrne, "Prediction of Paroxysmal Atrial Fibrillation using New Heart Rate Variability Features," *Comput. Biol. Med.*, p. 104367, Apr. 2021.

## 7.3      Future Work

Several aspects of the current study can be improved in future work.

The study of cardiac arrhythmia prediction should address the optimal time window to be used, informed by the clinical understanding of the heart. The time period where it is feasible to predict the occurrence of a cardiac event is still an open question. Moreover, most work to date has focused on VT-VF *detection* rather than *prediction*. This points to the need for further research on the temporal aspects of cardiac behaviour and how this may be exploited in cardiac event prediction.

ICDs can send their data wirelessly to a central server or could-based infrastructure, making it easier to gather a large database to have different patients with whole different type of arrhythmias. As an example, most patients that have experienced VT-VF should also have experienced a PAF event, but at the moment there is no database covering samples including PAF, VT, VF, and normal signal form a same patient. By developing a database like this, it would be possible to test and validate each propose method and to be able to use more recently-proposed approaches such as deep learning where massive amounts of data are required. The performance of the proposed classifiers should be further investigated using additional data collected in different environments.

Also experiments using artificially added perturbations to the input signal in a controlled way would further test the robustness of the proposed method.

The testing and validation a life-critical medical device requires a clinical trial. Furthermore, the idea of using computer models of the physiology to help plan and execute a trial is very attractive, particularly with a much larger database where different kind of cardiac arrhythmias are included. It may also be possible to extend feature ranking to include both HRV feature ranking as presented in this thesis but also applies additional clinical information input from specialist and cardiologist experience.

Some future works can be done to extend current work in a matter of HRV feature ranking. Finding a good balance between sensitivity and specificity is one of the major goals in prediction and detection algorithms and remains a topic of much research interest. The optimal value of parameter for some feature extraction techniques can be investigated on other databases to improve prediction performance of HRV features. In this study, more mathematical method like feature ranking method based on mutual information used, however using artificial neural network as a feature ranking method could be an interesting area here, subject to data sets of sufficient size being available to render this a viable approach.

In the context of PAF Prediction, investigating the KDE resolution could be the key to implement proposed method on ICDs as computational complexity of KDE decreases dramatically by lowering the resolution or using estimation methods for less complex calculation of KDE [234], [235]. Furthermore, as the maps are almost Gaussian in nature, future works will investigate using Gaussian approximations instead of KDE [236] to classify between Normal and PAF events on verity of atrial fibrillation databases.